\documentclass{article} 
\usepackage{iclr2027_conference,times}
\iclrfinalcopy 

\usepackage{amsmath,amsfonts,bm}

\def\eqref#1{equation~\ref{#1}}

\def\1{\bm{1}}

\DeclareMathAlphabet{\mathsfit}{\encodingdefault}{\sfdefault}{m}{sl}
\SetMathAlphabet{\mathsfit}{bold}{\encodingdefault}{\sfdefault}{bx}{n}

\usepackage{hyperref}
\usepackage{xcolor}
\usepackage{graphicx}
\usepackage{subcaption}
\usepackage{array}
\usepackage{tabularx}
\usepackage{multirow}
\usepackage{xspace}
\usepackage{booktabs}

\usepackage{url}
\usepackage{cleveref}
\usepackage{natbib}
\usepackage{wrapfig}

\usepackage{algorithm}
\usepackage{algpseudocode}

\usepackage{amsfonts}
\usepackage{amsmath}
\usepackage{amssymb}
\usepackage{amsthm}

\usepackage{microtype}
\usepackage{enumitem}

\definecolor{darkblue}{rgb}{0, 0, 0.5}
\hypersetup{colorlinks=true, citecolor=darkblue, linkcolor=darkblue, urlcolor=darkblue}

\newcommand{\approach}{\emph{DCD}\xspace}
\newcommand{\approachC}{\approach-Cluster\xspace}
\newcommand{\krandom}{\textit{K}-\textsc{Random}}
\newcommand{\krep}{\textit{K}-\textsc{Representative}}

\title{Data-driven Circuit Discovery \\ for Interpretability of Language Models}

\author{
Daking Rai \\
George Mason University \\
\texttt{drai2@gmu.edu} \\
\And
Mor Geva\thanks{Co-advising roles.} \\
Tel Aviv University \\
\texttt{morgeva@tauex.tau.ac.il} \\
\And
Ziyu Yao\footnotemark[1] \\
George Mason University \\
\texttt{ziyuyao@gmu.edu}
}

\begin{document}
\maketitle
\lhead{Preprint. Under review.}

\begin{abstract}
{Circuit study aims to explain how language models (LMs) implement a task by localizing and interpreting a \emph{circuit}: a computational subgraph responsible for a specific model's behavior. Existing methods for circuit localization are largely \emph{hypothesis-driven}: they define a task through a dataset, assume that all examples in the task are implemented by the same mechanism, and apply circuit discovery to obtain a single circuit. This workflow does not account for \emph{mechanistic diversity} within a task, where different examples may rely on different mechanisms. We first show that when a dataset mixes tasks processed through distinct mechanisms, current circuit discovery can combine these mechanisms into a single circuit while retaining faithfulness comparable to circuits discovered on each task separately. Across four tasks and three models, we further provide evidence for such mechanistic diversity within individual tasks. To account for this diversity, we propose \emph{Data-driven Circuit Discovery}
(\approach{}), which modifies the circuit-discovery workflow by adding a
grouping step before the discovery. The grouping step aims to partition
the dataset into subsets of examples that rely on similar mechanisms.
We instantiate this step with clustering over per-example
attribution patterns (\approachC{}). Across four tasks and three models,
\approachC{} decomposes the dataset into multiple subsets whose circuits exhibit distinct faithfulness patterns, providing useful hypotheses about mechanistic differences among examples.}\footnote{
Code is available at
\url{https://github.com/Ziyu-Yao-NLP-Lab/data-driven-circuit-discovery}.
}
\end{abstract}

\section{Introduction}
Mechanistic interpretability (MI) aims to reverse-engineer the internal mechanisms of LMs into human-understandable explanations~\citep{olah2020zoom, bereska2024mechanistic, rai2024practical, ferrando2024primer}. A prominent paradigm within MI is circuit discovery, which seeks to explain how LMs implement a human-defined task by localizing and interpreting a circuit: a subgraph of the model's computation graph responsible for implementing the task~\citep{elhage2021mathematical, wang2023interpretability, hanna2023does, marks2025sparse}.

\begin{figure*}[t]
    \centering
    \includegraphics[width=0.9\textwidth]{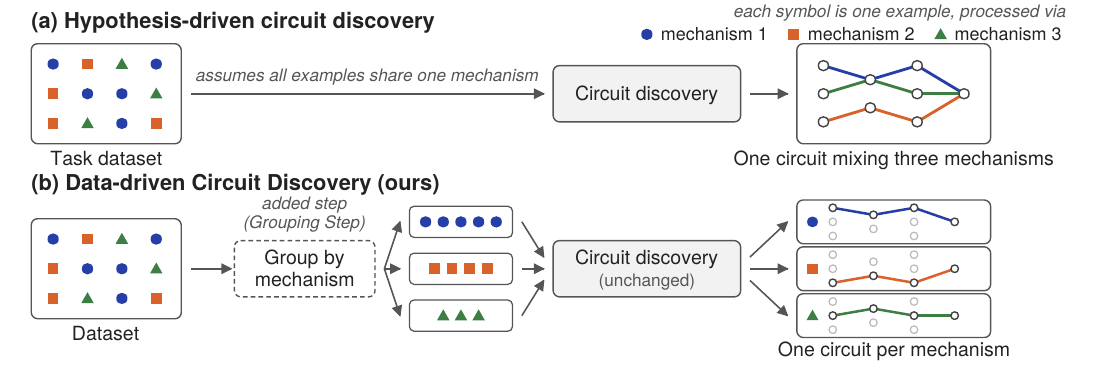}
    \caption{\textbf{Comparison of circuit discovery workflows.} 
    \textbf{(a) Hypothesis-driven:} researchers construct a dataset for the task and localize a single circuit, hypothesizing that the single mechanism implements the task. \textbf{(b) Data-driven Circuit Discovery (\approach)} adds a new grouping step to the workflow: \approach groups examples in a dataset with similar computation and then localizes one circuit for each group, yielding multiple mechanism-specific circuits.
    }
    \vspace{-1em}
    \label{fig:approach_comparison}
\end{figure*}

Existing circuit studies~\citep{elhage2021mathematical, wang2023interpretability, hanna2023does, hanna2024have} share a common workflow (Figure~\ref{fig:approach_comparison}a): define a task, construct a representative dataset, discover a single circuit, and interpret it as the task mechanism. By design, this workflow returns one circuit per task (or dataset) and therefore implicitly assumes that all examples of a task are processed by the model the same way. We refer to such a workflow as \emph{hypothesis-driven} circuit discovery. This assumption does not consider the possibility of \emph{mechanistic diversity within a task}: examples that fall under the same task label may be processed by genuinely different mechanisms. In fact, recent work has shown that LMs indeed employ multiple distinct mechanisms for a single task, including tasks like arithmetic reasoning~\citep{nikankin2025arithmetic}, factual recall~\citep{chughtai2024summing}, entity binding~\citep{gur2025mixing}, syntactic code generation~\citep{rai2025failure}, and indirect object identification~\citep[IOI;][]{franco2026finding}. 

To understand this limitation, we first apply existing circuit discovery methods to datasets mixing two tasks the model handles through distinct mechanisms. The methods return a single circuit that combines both mechanisms while retaining faithfulness comparable to circuits discovered separately for each task. 
Thus, high faithfulness does not imply a single mechanism: a faithful circuit can combine distinct mechanisms. We argue that mechanistic diversity can also arise within a single task, as models may process different examples of the same task using different mechanisms, as also evidenced by prior work~\citep{nikankin2025arithmetic, rai2025failure, chughtai2024summing, gur2025mixing}. We investigate this across four widely studied tasks -- Indirect Object Identification~\citep[IOI;][]{wang2023interpretability}, entity binding~\citep{prakash2024finetuning, gur2025mixing}, arithmetic addition~\citep{nikankin2025arithmetic, mamidanna2025all}, and sequence completion~\citep{elhage2021mathematical} — by constructing variants that preserve the task semantics while varying syntax, complexity, or domain. We find that circuits discovered from different variants of the same task can have low cross-variant faithfulness and limited edge overlap, providing evidence for mechanistic diversity within a task.


In light of these results, we propose \emph{Data-driven Circuit Discovery} (\approach), which modifies the \emph{circuit discovery workflow} to account for mechanistic diversity within a task. \approach{} adds a grouping step after dataset construction that aims to partition examples according to the mechanisms they rely on (Figure~\ref{fig:approach_comparison}b). Circuit discovery is then applied independently to each group using any existing method unchanged. If the grouping separates examples that rely on different mechanisms, their circuits can be discovered separately rather than combined into a single circuit. Specifically, \approach{} introduces no new attribution or circuit-discovery technique; the key change is to separate examples before their attribution scores are aggregated into a single circuit. {As one way to identify such groups, we represent each example by its edge-attribution scores~\citep{hanna2024have}, which provide a signal of which model components contribute to processing that example. This gives an instantiation of \approach, referred to as \approachC. We do not treat the resulting clusters as definitive mechanisms; we use them as hypotheses about mechanistic differences and test whether their circuits exhibit distinct faithfulness patterns. Our experiments suggest that \approachC{} can generate useful hypotheses about mechanistic differences across examples without access to task or variant labels for individual examples. We note that \approach is agnostic to this specific instantiation, and future work can investigate alternative approaches.}

We highlight our three key contributions:
\begin{itemize}[noitemsep, topsep=0pt, partopsep=0pt, parsep=0pt]
  \item We show that existing circuit-discovery workflows do not account for mechanistic diversity: when a dataset contains examples processed through different mechanisms, they can combines these mechanisms into a single circuit.
  \item To account for this mechanistic diversity, we introduce Data-driven Circuit Discovery (\approach{}), which adds a grouping step to the existing workflow that aims to partition the dataset into subsets of examples processed through similar mechanisms (Figure~\ref{fig:approach_comparison}).
  \item {We instantiate the grouping step with clustering (\approachC{}), using per-example edge-attribution scores, and show that it can generate useful hypotheses about mechanistic differences without access to task or variant labels for individual examples.}
\end{itemize}

\section{Background and notations}\label{sec:background}
In this section, we introduce the notions of circuits and tasks, and describe the hypothesis-driven approach to circuit discovery.

\paragraph{Circuit and task} 
In circuit studies, we understand how LMs implement a specific task by localizing the circuit and interpreting it as human-understandable algorithms. Formally, we represent the LM as a directed computation graph $G=(N,E)$, where each node $n \in N$ corresponds to a distinct model component (e.g., the layer 5 MLP, or attention head 10 at layer 6) and each edge $(n_1,n_2)\in E$ represents a direct causal pathway from $n_1$ to $n_2$. A circuit $C = (N_C, E_C)$ is a subgraph of $G$ where $N_C \subseteq N$ and $E_C \subseteq E$, representing a sparse subset of model components and their connections. A task ($\tau$) in circuit studies is a human-defined concept (e.g., arithmetic addition); they are not formally specified but are operationalized through a dataset $D_{\tau} = \{(x_i, y_i)\}_{i=1}^{|D_{\tau}|}$ consisting of input-output pairs representative of the task. A circuit $C$ is considered to provide a \textit{faithful} mechanistic explanation of $\tau$ if the model’s behavior on $D_{\tau}$ is preserved even when all nodes and edges outside $C$ are ablated, e.g., through zero or mean activation~\citep{zhang2023towards}.

\paragraph{Hypothesis-driven circuit discovery}
To find a circuit for a given task $\tau$, existing circuit discovery methods~\citep{conmy2023acdc, syed2024attribution, hanna2024have} follow a common workflow: prepare a dataset $D_{\tau}$ to represent $\tau$, then run a procedure that returns a single circuit $C$ implementing $\tau$. 
Specifically, given a dataset $D_{\tau} = \{(x_i, y_i)\}$, these methods compute a per-example attribution score $s(e; x_i) \in \mathbb{R}$ for each edge $e \in E$, and then take average across all dataset examples to obtain $s_{D_\tau}(e)$.
Once we have $s_{D_\tau}(e)$ for all edges, circuit $C$ is generally constructed by selecting the smallest set of top-ranked edges such that a target level of faithfulness (generally $\geq 80\%$) on the held-out test set. Several algorithms have been proposed for circuit discovery, and they primarily differ in how they compute the attribution score $s(e; x_i)$. For instance, edge activation patching \citep[{E-Act};][]{conmy2023acdc} measures the effect on the model output logit when an edge's activation is ablated~\citep{zhang2023towards}. Edge Attribution Patching \citep[\textsc{EAP};][]{syed2024attribution} approximates this effect via gradient-based attribution, offering improved computational efficiency. \textsc{EAP-IG}~\citep{hanna2024have} extends \textsc{EAP} by using integrated gradients for a more reliable approximation. 
{We refer to this paradigm as \textit{hypothesis-driven circuit discovery} because it rests on an unstated hypothesis: that the task $\tau$ is implemented by a single mechanism. The workflow is built without testing this hypothesis: averaging attribution scores across all examples of $D_\tau$ treats every example as an instance of the same computation, and the procedure returns one circuit. There is no step that asks whether the examples are processed the same way, and no way to return more than one circuit if they are not. Mechanistic diversity within a task, if present, is therefore invisible to the workflow by construction. In Section~\ref{sec:mix-experiment}, we will carefully examine this limitation.



\section{Experimental setup}
\label{sec:experiment_setup}
\paragraph{Tasks}
\label{subsec:tasks}
We conduct our study across four tasks commonly used in prior circuit discovery work: \emph{(1) Indirect Object Identification (IOI)} consists of input sentences like \emph{``When Mary and John went to the store, John gave an apple to''}, which contain a subject (\emph{John}) and an indirect object (\emph{Mary}); the model's task is to predict the indirect object~\citep{wang2023interpretability, conmy2023acdc, hanna2024have, mueller2025mib}.
\emph{(2) Entity Binding (EB)} consists of input sentences like \emph{``The key is in box D, the rose is in box C. Box D contains the''}, which consists a list of (entity, container) pairs and query the entity associated with a specific container; the model's task is to predict the correct entity (\emph{key})~\citep{prakash2024finetuning, gur2025mixing}.
\emph{(3) Arithmetic Addition (Arith)} involves expressions such as \emph{``10 + 50 =''}; the model's task is to predict the result (e.g., \emph{60}). Following prior works~\citep{stolfo2023mechanistic, nikankin2025arithmetic, mueller2025mib}, we restrict both operands and results to single tokens. \emph{(4) Sequence Completion (SC)} consists of input sequences like $[A][B] \ldots [A]$, in which a base sequence is repeated with the final repetition truncated; the model predicts the next token ($[B]$) by matching the current context against earlier occurrences of the same sub-sequence. Sequence completion is widely used for studying induction circuits~\citep{elhage2021mathematical, olsson2022context}. We provide dataset details and representative examples in Appendix~\ref{app:dataset}.


\paragraph{Models}
{We conduct our experiments on three models spanning a range of sizes, families, and capability levels: GPT-2 Small (124M parameters)~\citep{radford2019language}, Qwen2.5-7B-Instruct~\citep{qwen2024qwen2}, and Llama-3.1-8B-Instruct~\citep{dubey2024llama}. Circuit study is only meaningful for behaviors the model reliably exhibits; we follow prior work~\citep{mueller2025mib, hanna2024have} in restricting each model to only tasks where it exhibits reasonable accuracy. Per-task accuracies for all models are reported in Appendix~\ref{app:dataset-accuracy}.}

\paragraph{Circuit faithfulness}
\label{subsec:faithfulness}
We measure how well a discovered circuit $C$ recovers the full model's behavior for given task using the faithfulness score: $f(C,G;m)=\frac{m(C)-m(\varnothing)}{m(G)-m(\varnothing)}$, where $G$ is the full model, 
$\varnothing$ is the model with all circuit components ablated by running a counterfactual prompt,
and $m(\cdot)$ is the logit difference between the correct and counterfactual labels, following prior works~\citep{hanna2024have, mueller2025mib}.
A score of $1$ indicates that $C$ matches the full model's behavior, and $0$ indicates it matches the empty-circuit baseline. The score is not bounded to $[0, 1]$: it can exceed $1$ when $m(C) > m(G)$ (i.e., circuit outperforms the full model) and fall below $0$ when $m(C) < m(\varnothing)$ (i.e., circuit is not more performant than the baseline).


\section{{Hypothesis-Driven Circuit Discovery Mixes Mechanisms into a Single Circuit}}
\label{sec:mix-experiment} 
\begin{figure}[t!]
    \centering
    \includegraphics[width=\textwidth]{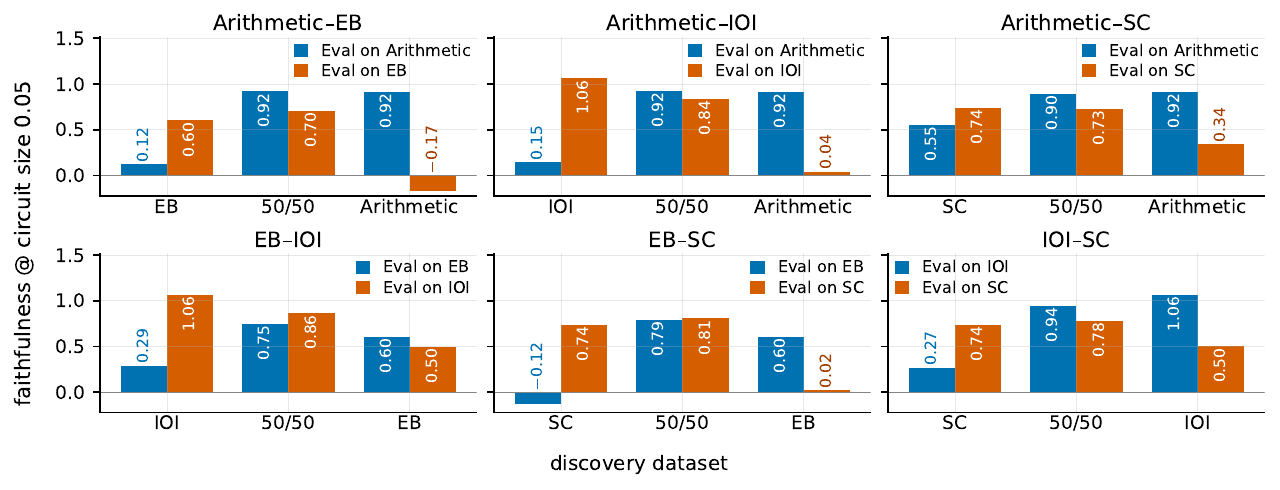}
    \caption{{A circuit discovered on a 50/50 mixture of two tasks is faithful to both; a circuit discovered on either task alone is not. Each panel is a task pair (A, B). The x-axis is the discovery dataset (pure A, 50/50 mixture, pure B); bars show faithfulness at circuit size 0.05 (i.e., top 5\% of model edges) on held-out test sets of A (blue) and B (orange). Results are for EAP-IG on Llama-3.1-8B-Instruct; other models and attribution methods are in Appendix~\ref{app:mixing}. EB: Entity Binding; SC: Sequence Completion.}
    }
    \label{fig:exp2}
\end{figure}
\subsection{{A dataset mixing two tasks can return one circuit with high faithfulness on both}}
\label{subsec: cross-task-mixing}
\paragraph{Experiment}
{We first investigate what happens when a circuit discovery dataset contains examples that are processed by different mechanisms. 
To construct such a dataset, we utilize the four tasks introduced in Section~\ref{subsec:tasks}. We first confirm that these tasks employ different mechanisms. To this end, we discover one circuit on the training set of each task and then measure its faithfulness on the test set of the other task.
Each circuit performs poorly when evaluated on the other task compared to its own (Figure~\ref{fig:exp2}), suggesting that they use different mechanisms. Now that we have four tasks that use different mechanisms, we construct a dataset that deliberately mixes two tasks with equal proportions.\footnote{We also mix datasets at other proportions and report their results in Appendix~\ref{app:mixing-sweep}.}  
For each mixture, we likewise discover a single circuit and then measure its faithfulness on the test set of each task.
This lets us examine what a single circuit discovered from a mechanistically heterogeneous dataset captures: whether it primarily reflects one task or incorporates components from both mechanisms. We show results of Llama-3.1-8B-Instruct based on EAP-IG \citep{hanna2024have} in Figure~\ref{fig:exp2}. We additionally measure structural similarity between circuits using Jaccard similarity over their edge sets, $J(C^i, C^j) = |E_{C^i} \cap E_{C^j}| / |E_{C^i} \cup E_{C^j}|$.} Dataset construction details and examples are provided in Appendix~\ref{app:dataset}.

\paragraph{Circuit discovery can combine distinct mechanisms into a single circuit}
{
As shown in Figure~\ref{fig:exp2}, the circuits discovered on the mixed datasets retain faithfulness on both tasks comparable to circuits discovered on each task separately. 
For example, for Arithmetic--EB, the mixed circuit achieves $0.92$ faithfulness on arithmetic, identical to the arithmetic-only circuit ($0.92$), and $0.70$ on EB, compared with $0.60$ for the EB-only circuit. In contrast, the task-specific circuits transfer poorly across tasks: the arithmetic circuit achieves $-0.17$ on EB, while the EB circuit achieves only $0.12$ on arithmetic. Thus, although the two task-specific circuits do not explain each other's behavior, a single circuit discovered from their mixture can remain highly faithful to both. 
We observe similar findings on GPT-2 and Qwen2.5-7B-Instruct, and when using EAP or E-Act (Appendix~\ref{app:mixing}). Together, these results show that existing circuit discovery can combine distinct mechanisms into a single circuit, and faithfulness may not reveal the mixture.}

\paragraph{A single circuit can contain components of multiple mechanisms}
{The faithfulness results show that a single circuit from the mixed dataset can be faithful to both tasks, but not whether it captures a shared mechanism or combines distinct task-specific mechanisms. We therefore examine the circuit's edge composition. For each mixture, we partition the discovered circuit's top-$5\%$ edges according to whether they occur in the circuits discovered on task $A$ alone, task $B$ alone, both, or neither. Across the task pairs, models, and attribution methods we test, $88$--$99\%$ of the mixed circuit's edges belong to at least one of the two task-specific circuits, with both represented (Appendix~\ref{app:mixing-composition}). For example, in the 50/50 Arithmetic--EB mixture, $28\%$ of the edges occur in both task-specific circuits, $15\%$ only in the arithmetic circuit, $55\%$ only in the EB circuit, and just $2\%$ in neither. This provides evidence that the discovered circuit does not correspond to a new, single mechanism shared across the full dataset, but instead combines components of the distinct mechanisms used by different subsets of examples.
}

\subsection{A Single Task Can Also Contain Multiple Mechanisms}
\label{subsec: single-task-mixing}

{Section~\ref{subsec: cross-task-mixing} deliberately mixes two tasks to show that existing circuit discovery can combine distinct mechanisms into a single circuit. In practice, however, circuit studies typically construct a dataset for a single task and interpret the resulting circuit as the mechanism for that task. We therefore ask whether examples that fall under the \emph{same task definition} necessarily rely on the same mechanism.}

\paragraph{Experiment.}
For each task $\tau$, we construct several dataset variants that preserve the input--output relation defining the task while varying surface features along one of three axes: complexity (e.g., 2-person vs.\ 3-person IOI), syntax (e.g., active vs.\ passive), and domain (e.g., names vs.\ single-letter labels). Across the four tasks, we construct 32 variants in total: 7 for IOI, 17 for entity binding, 5 for arithmetic, and 3 for sequence completion (Table~\ref{tab:variant-inventory}). Each variant contains a discovery set of 600 prompts and a held-out test set of 200 prompts.
For two variants $i$ and $j$ from the same task and axis, we independently discover circuits $C_i$ and $C_j$ using EAP-IG~\citep{hanna2024have}. We then ask whether the circuit discovered on variant $i$ remains faithful when evaluated on variant $j$: $\Delta_{i\rightarrow j}=f(C_j,D_j^\tau)-f(C_i,D_j^\tau)$. We count $i\rightarrow j$ as a \emph{transfer failure} only when this shortfall exceeds the larger of the two variants' noise floors, estimated by bootstrapping their in-distribution test sets (Appendix~\ref{app:faithfulness-floor}). For each unordered pair $\{i,j\}$, we evaluate 3 possible outcomes: both directions transfer, only one direction transfers, or neither direction transfers. We apply this test to a total of 136 variant pairs spanning 4 tasks and 3 models; 
Appendix~\ref{app:variant-pairs} details how these pairs are constructed.

\begin{figure}[t!]
    \centering
    \includegraphics[width=\textwidth]{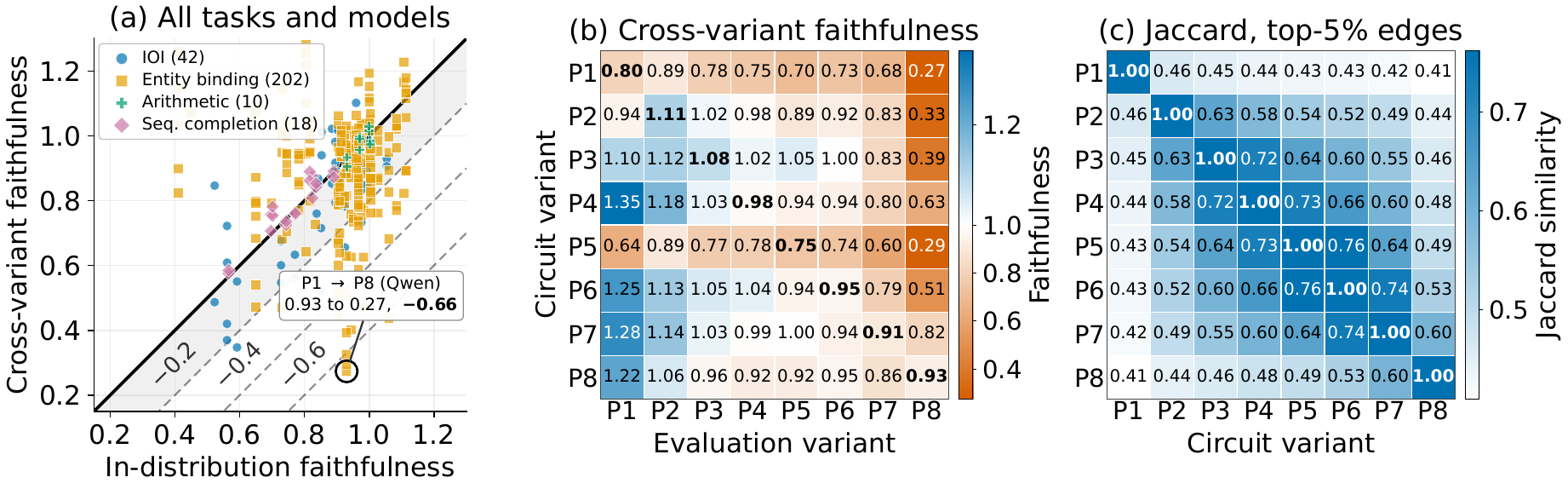}
    \caption{\textbf{(a)} Cross-variant vs in-distribution faithfulness at circuit size $0.05$ for every ordered variant pair, colored by task, for all models (e.g., 42 IOI pairs for 3 models $\times$ 7 pairs each $\times$ 2 way transfers in Table~\ref{tab:transfer}). Each point is a transfer $i \rightarrow j$ between two variants $i$ and $j$ of the same task, plotting $f_{ij}$ against $f_{jj}$ on $j$'s test set.
    \textbf{(b)} Cross-variant faithfulness for the EB position variants $P_1$--$P_8$ on Qwen2.5-7B; rows are circuits, columns are the variants they are evaluated on. \textbf{(c)} Pairwise Jaccard similarity of the same circuits' edges.}
    \label{fig:exp1}
\end{figure}

\begin{table}[t]
\centering
\footnotesize
\setlength{\tabcolsep}{6pt}
\renewcommand{\arraystretch}{1.0}
\caption{Transfer outcomes (both\,/\,one\,/\,neither direction transfers) for the 136 within-axis variant pairs.}
\label{tab:transfer}
\begin{tabular}{lcccc}
\toprule
& GPT-2 S & Llama-3.1 & Qwen2.5 & Total \\
\midrule
IOI    & 0/6/1 & 1/4/2   & 2/4/1   & 3/14/4 \\
EB     & --    & 19/31/1 & 9/39/2  & 28/70/3 \\
Arith. & --    & 3/2/0   & --      & 3/2/0 \\
SC     & 3/0/0 & 3/0/0   & 3/0/0   & 9/0/0 \\
\midrule
Total  & 3/6/1 & 26/37/3 & 14/43/3 & 43/86/7 \\
\bottomrule
\end{tabular}
\end{table}


\paragraph{Circuits discovered on one variant often fail on another variant of the same task.}
Table~\ref{tab:transfer} reports the transfer outcome for all 136 pairs, and Figure~\ref{fig:exp1}a plots cross-variant against in-distribution faithfulness for every ordered pair.
Only 43/136 pairs transfer in both directions, i.e., each circuit remains faithful on the other variant's test set. The largest group, 86/136 pairs, transfers only in one direction. A one-way transfer suggests either that the mechanism for $i$ subsumes the mechanism for $j$, or that the two are distinct mechanisms with partial overlap. The asymmetry can be substantial: for the EB position variants on Qwen2.5-7B, the $P_1$ (i.e., querying the 1st entity in the prompt) circuit drops from $0.93$ faithfulness on $P_1$ to $0.27$ on $P_8$ (i.e., querying the 8th entity), a loss of 0.66 faithfulness (Figure~\ref{fig:exp1}a). 
Finally, 7/136 pairs fail to transfer in either direction (Appendix, Table~\ref{tab:s42-twoway}). This is the strongest evidence that variants of the same task can rely on different mechanisms, as neither circuit explains the model's behavior on the other variant. These pairs arise on IOI in all three models and on EB in Qwen2.5-7B. 
Arithmetic and SC show no such pairs among the variants we construct, so not every surface variation induces a new mechanism. Taken together, a circuit discovered on one variant is faithful to all other variants of its task in fewer than half of the pairs we test, illustrating the mechanistic diversity within a single task.

\paragraph{Circuits of different variants also differ structurally.} {
Across the 136 variant pairs, circuits that fail to transfer in either direction have the lowest median top-$5\%$ edge overlap: Jaccard $0.45$, compared with $0.55$ for one-way pairs and $0.54$ for pairs that transfer in both directions. On Qwen2.5-7B's entity-binding position variants, the edge overlap falls steadily as positions get further apart, from 0.60–0.76 for neighbouring positions to 0.41 for $P1$ and $P8$, and transfer weakens with distance in the same way. Together, the behavioral and structural results provide evidence that different subsets of a single task can rely on different mechanisms, which a single-circuit workflow will combine.

}


\section{\approach: Data-driven Circuit Discovery} 
\label{exp3: our-approach}
Our results in Sections~\ref{sec:mix-experiment} show that hypothesis-driven workflow can easily mix distinct mechanisms when they return a single circuit, which hides the mechanistic diversity of a model and complicates its interpretation.
In light of this, we propose a \emph{Data-driven Circuit Discovery} (\approach), a circuit discovery framework that aims to discover multiple circuits with distinct mechanisms from a single dataset, where each circuit explains a subset of the dataset that relies on a similar mechanism, not the full task.
\subsection{Approach} 
\label{subsec: approach}

\begin{algorithm}[t]
\caption{\textit{Data-driven Circuit Discovery (\approach)}}
\label{alg:dcd}
\small
\begin{algorithmic}[1]
\Require model $G$; data $D$; method $\mathcal{F}$; grouping $\Pi$
\Ensure circuits $\{C_k\}$; partition $\{D_k\}$
\State $\{D_k\}_{k=1}^{K} \gets \Pi(D,G)$ \Comment{new step}
\For{$k=1,\dots,K$}
  \State $C_k \gets \mathcal{F}(D_k)$ \Comment{$\mathcal{F}$ unchanged}
\EndFor
\State \Return $\{C_k\},\{D_k\}$
\end{algorithmic}
\end{algorithm}


\paragraph{The framework}
We present \approach{} workflow in Algorithm~\ref{alg:dcd}. \approach{} modifies the standard circuit-discovery workflow by adding a \emph{grouping step} before circuit discovery; everything else remains unchanged. Given a dataset $D=\{x_1,\dots,x_n\}$ and an LM with computation
graph $G=(N,E)$, a grouping method $\Pi$ partitions the examples into
$\{D_k\}_{k=1}^{K}=\Pi(D,G)$, with the goal of grouping examples that rely on
the same mechanism. An existing circuit-discovery method $\mathcal{F}$ is then
applied independently to each group, producing
$C_k=\mathcal{F}(D_k)$ (Algorithm~\ref{alg:dcd}). Thus, rather than applying
$\mathcal{F}$ once to the full dataset and returning a single circuit, DCD
applies the same method to each group and returns $\{C_k\}_{k=1}^{K}$.
Importantly, $\mathcal{F}$ itself is unchanged: DCD only introduces $\Pi$ to
separate examples before their attribution scores are aggregated. If the groups capture different mechanisms, circuit discovery can recover them separately rather than merge them into one circuit.

\noindent\textbf{Instantiating the grouping step}\hspace{1em}%
\emph{\approach{} does not prescribe how the grouping step should be performed.}
In practice, identifying mechanistically meaningful groups may require iterative circuit discovery and interpretation rather than a single automated method. Here, we investigate an instantiation,
\approachC{}, which uses clustering techniques to propose hypotheses for candidate groupings from per-example attribution vectors (Algorithm~\ref{alg:cluster}). These vectors indicate which edges contribute to the model's behavior on each example, so similarity between them provides a natural signal that two examples may rely on similar computation. Specifically, for each example $x_i$, we obtain an attribution vector $\mathbf{s}_i \in \mathbb{R}^{|E|}$ and binarize it to the smallest set of top-magnitude edges containing a fraction $\gamma = 0.99$ of its attribution mass. Because these vectors are high-dimensional (${\sim}10^4$ edges for GPT-2 to ${\sim}10^6$ for Llama-3.1-8B), we reduce them to $\tilde{\mathbf{s}}_i \in \mathbb{R}^{20}$ with PCA, cluster with $k$-means, and select the number of groups with different strategies like the gap statistic~\citep{tibshirani2001estimating} or, when a reference partition is available, by fixing K to the number of known subgroups. The attribution signal, binarization, reduction, clustering algorithm, and criterion for $K^*$ all belong to this instantiation rather than to \textit{DCD}; we report hyperparameter selections and clustering variants in Appendix~\ref{app:clustering}. 



\subsection{Can attribution clustering inform mechanistic diversity of dataset?}
\label{subsec:evaluation}

\paragraph{Experiment}
We investigate whether \approachC{} provides useful signals for mechanistic boundaries within a dataset, treating each cluster as a hypothesis that its examples rely on a similar mechanism. First, we construct \emph{cross-task} and \emph{within-task} mixtures matching Section~\ref{sec:mix-experiment}, each with $n=1{,}000$ examples split 60/20/20 into discovery, validation, and test sets (Table~\ref{tab:mixture-sizes}). For cross-task, \approachC{} finds $K$ using the gap statistic over the dev set, and for the within-task mixtures we fix $K$ to the number of variants on the axis, since the gap statistic over-splits these smaller mixtures; details on dataset construction and setup in Appendix~\ref{app:clustering}. Specifically, after \approachC{} partitions the dataset into multiple groups, we analyze each group through (1) its composition by task or task variant, i.e., how many examples come from each task or variant, and (2) the per-example faithfulness of its circuit across the held-out test set.Composition reveals what distinguishes the cluster, while circuit faithfulness tests whether that distinction corresponds to a consistent difference in which examples the circuit explains. We compare these patterns with the transfer relationships from Section~\ref{sec:mix-experiment}, without treating transfer as ground truth, and with \krandom{}, which shows what structure can arise from splitting the data alone. Together, these analyses test whether the candidate groupings proposed by \approachC{} correspond to meaningful mechanistic differences.\footnote{We additionally verify that the $K$ cluster circuits collectively cover the held-out examples using best-of-$K$ faithfulness; details are in Appendix~\ref{app:dcd-best-of-k}.}

\begin{figure}[t!]
    \centering
    \includegraphics[width=0.95\linewidth]{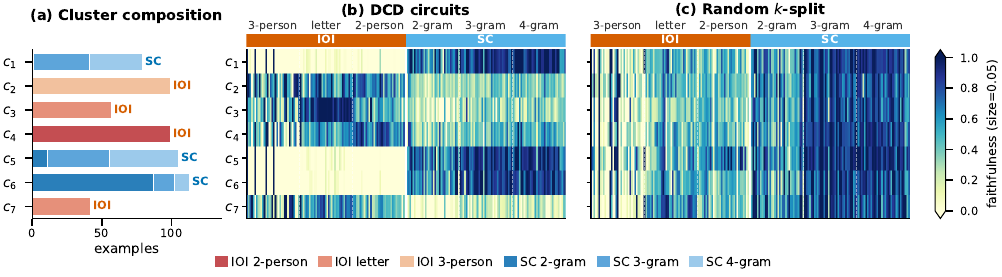}
    \caption{Cross-task circuit discovery on GPT-2. Each row across (c1-c7) corresponds to one circuit. (a) Each \approach{} cluster consists of examples from only a single task; (b) per-example faithfulness shows clear block structure -- each \approach{} circuit achieves high faithfulness on a subset of examples. (c) \textsc{K-Random}'s circuits show diffuse, near-uniform faithfulness with no circuit aligning with a specific subset of examples.}
\label{fig:dcd-coherence-gpt2}
\end{figure}

\paragraph{Across tasks, DCD separates examples of different tasks into different clusters}
{
Across all three cross-task mixtures, \approachC{} partitions each dataset into clusters containing examples from only one task (Appendix~\ref{app:composition-crosstask}). For instance, the GPT-2 cross-task dataset contains IOI and SC examples spanning three variants of each task (Figure~\ref{fig:dcd-coherence-gpt2}a). \approachC{} partitions this dataset into seven clusters, each containing examples from only one task: $c_1$, $c_5$, and $c_6$ contain only SC examples, while $c_2$, $c_3$, $c_4$, and $c_7$ contain only IOI examples. The IOI clusters further align partly with variants---$c_2$ with 3-person, $c_4$ with 2-person, and $c_3$ and $c_7$ with letter examples---whereas the SC clusters mix $n$-gram lengths. In addition, \approachC{} circuits also show a corresponding faithfulness pattern (Figure~\ref{fig:dcd-coherence-gpt2}b): SC circuits are highly faithful on SC test sets and near zero on almost all IOI examples, while IOI circuits show the reverse, revealing a clear task-level boundary. Within IOI, the correspondence is weaker: $c_3$ is most faithful on letter examples, but $c_7$ is not. These results are consistent with Section~\ref{subsec: cross-task-mixing} and contrast with \krandom{}, which mixes both tasks within each group and yields circuits that remain faithful across both, with no comparable structure (Figure~\ref{fig:dcd-coherence-gpt2}c).}

\paragraph{Within a task, DCD recovers differences identified by cross-variant transfer.} 
{We evaluate \approachC{} across all within-task settings corresponding to the 136 variant pairs from Section~\ref{subsec: single-task-mixing}. We first focus on the 7 pairs (involving 13 distinct variants) for which neither variant-specific circuit transfers to the other, since these provide our strongest evidence that the two variants rely on different mechanisms. We begin by investigating the cluster composition, and found \approachC{} keeps the two variants separate different clusters across all seven cases. We found 100\% of the IOI examples and 87--99\% of the EB examples fall into clusters dominated by one variant, compared with 50\% under random assignment (Appendix~\ref{app:cluster-composition}). In addition, 11 of the 13 total variants form their own cluster: most of their examples fall into a cluster dominated by that variant. All IOI variants, $P_1$ and $P_8$ on Llama and $P_8$ on Qwen2.5, form their own clusters. The only exceptions are the two middle-position variants on Qwen2.5, $P_3$ and $P_4$, whose examples are spread across several clusters.
We next ask whether this compositional structure is reflected in the faithfulness of the corresponding cluster circuits. Four of the seven exhibit a clear two-block pattern, where different circuits cover different variants, while one shows a weaker asymmetric pattern (Appendix~\ref{app:twoway-clusters}). \krandom{} shows a boundary for only two of the seven pairs. For the remaining pairs with one-way or two-way transfer, however, \approachC{} is not consistently more structured than \krandom{}. Figure~\ref{fig:dcd-within-task} illustrates representative faithfulness patterns across the three transfer outcomes. Overall, \approachC{} is most informative in the clearest cases of within-task mechanistic differences, while subtler relationships are harder to distinguish from structure induced by random partitioning.}

\begin{figure}[t!]
    \centering
    \includegraphics[width=0.95\linewidth]{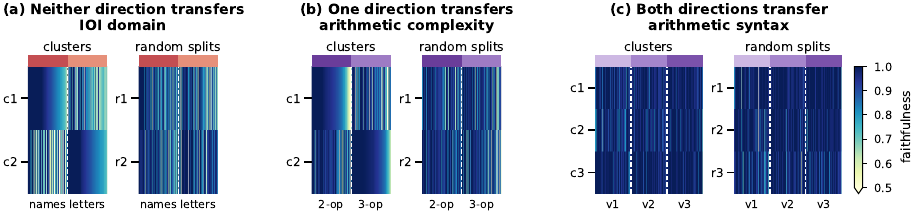}
    \caption{One example of each transfer outcome from Section~\ref{subsec: single-task-mixing} for Llama-3.1-8B at circuit size $0.05$. Rows correspond to circuits from \approachC{} clusters (left) or matched random partitions (right; \krandom), and columns to test variants. \approachC yields more structured circuit separation: (a) when neither direction transfers, each cluster circuit covers a different variant; (b) in one-way transfer, the 3-operand cluster circuit (c2) covers both variants, while the 2-operand circuit (c1) covers only its own; and (c) in two-way transfer, all circuits cover all variants.} 
    
\label{fig:dcd-within-task}
\end{figure}

\section{Related work}
\label{sec:related-work}


\noindent\textbf{Circuit study} {aims to understand how a specific model implements a task by localizing and interpreting a circuit. Earlier works relied on manual techniques such as activation and path patching to identify circuits, including induction heads~\citep{elhage2021mathematical}, IOI~\citep{wang2023interpretability}, and greater-than~\citep{hanna2023does}.
Subsequent work automated and scaled the localization process~\citep{conmy2023acdc, ferrando2024information, syed2024attribution, hanna2024have}; some were introduced in Section~\ref{sec:background}.
These approaches typically summarize a task with a single circuit, leaving open the possibility of \emph{mechanistic diversity within a task}. Recent work has identified related forms of variation: \citet{li2026much} study circuit consistency across examples and specificity across tasks, while \citet{chen2026all} show that multiple low-overlap circuits can support the same task. Our work focuses on a complementary question: whether examples within the same task rely on different mechanisms. We show that aggregating attribution across such examples can mix these mechanisms into a single circuit and thus propose \approach for data-driven discovery of multiple circuits.
}

\paragraph{Multi-mechanism findings in language models} {Recent work has found that LMs implement individual tasks via multiple distinct mechanisms. \citet{nikankin2025arithmetic} show that LMs solve arithmetic via a ``bag of heuristics'' rather than a single unified algorithm; \citet{chughtai2024summing} find that factual recall combines additive contributions from multiple independent mechanisms; and \citet{gur2025mixing} revisit entity binding and finds three interacting mechanisms where prior circuit work had identified a single circuit. Similar findings were also found beyond the text modality: \citet{nikankin2025same} investigate question-answering circuits in vision-language models, where they find that circuits across modalities are largely disjoint, but they implement relatively similar functionalities; and \citet{rai2025failure} find multiple competing mechanisms for balanced parentheses, a code completion task. Closely related to our study, \citet{franco2026finding} also discover multiple circuits for IOI and advocate for prompt-specific and representative circuits.

\section{Discussions and limitations}
\label{sec:limitations}
\paragraph{From task-specific to mechanism-specific circuits.}
Circuit studies typically organize the study around human-defined tasks and seek a single circuit for them. Our results suggest, however, that examples within the same task can rely on different mechanisms, which existing circuit-discovery workflows may combine into a single circuit. This can complicate interpretation: a task-level circuit may combine multiple mechanisms, requiring several underlying algorithms to be disentangled from a single circuit and making the role of components shared across them harder to characterize. Mechanisms used by only a small subset of examples may also be only partially recovered when attribution scores are aggregated across the full dataset, further complicating the interpretation of the circuit. We therefore argue that circuit discovery should aim for \emph{mechanism-specific} rather than task-specific circuits, where the scope of a circuit is determined by the examples that rely on the same underlying computation rather than fixed in advance by a task label. However, it also makes defining that scope a separate interpretive problem. One must characterize what each cluster represents, either by inspecting its examples or by interpreting the circuit itself. Additionally, open questions remain about how to effectively assign new examples to clusters at test time, and how to handle clusters that resist human-legible description. We leave these problems for future work.

\paragraph{What does \approachC{} assume?} We instantiate \approach with \approachC{}, which relies on existing circuit discovery methods, such as EAP-IG~\citep{hanna2024have}, to characterize per-example mechanism. It therefore inherits their assumptions, especially that attribution scores faithfully reflect the model's mechanism and provide a meaningful basis for clustering examples with similar mechanisms. \approachC{} also depends on clustering choices, including the selection criteria for the number of clusters $K^*$, dimensionality reduction method, distance metric, and clustering algorithm. We select these hyperparameters using grid search with the silhouette score, gap statistic; for within-task mixtures we fix $K^*$ to the number of variants, but whether these choices recover mechanistically meaningful structure remains an important question for future work.

\section{Conclusions}
{In this study, we provide evidence for mechanistic diversity within a single task and show that standard circuit-discovery workflows do not explicitly account for this diversity. To address this limitation, we introduce Data-driven Circuit Discovery (\approach), which adds a grouping step before circuit discovery to group examples that may rely on similar mechanisms and separate those that may not. The change is to the workflow, not the underlying discovery method: by discovering circuits within groups rather than averaging attribution scores across the full dataset, \approach{} can expose mechanistic diversity that would otherwise be combined into a single circuit. More broadly, our paper argues for discovering mechanism-specific circuits, whose boundaries come from the model's own computation, rather than task- or dataset-specific circuits, whose boundaries we fix in advance.}

\subsection*{Acknowledgments}
This project was sponsored by the National Science Foundation (Award Number 2311468/2423813). The project was also supported by the Alon Scholarship and by GPU resources provided by the Office of Research Computing at George Mason University (URL: \url{https://orc.gmu.edu}) and funded in part by grants from the National Science Foundation (Award Number 2018631).

\subsection*{AI use statement}
In this work, we used generative AI tools for writing code to implement experiments and for polishing the writing. We have not used generative AI tools for coming up with research ideas, designing experiments, or writing the scientific narrative of the paper; the remaining required-disclosure tasks are not applicable to this work. Additionally, we used generative AI tools to reorganize and condense parts of the appendix and to check consistency in numbers and cross-references between the main text and appendix. We have reviewed all AI-assisted work: LLM-generated code was verified and tested for correctness by the authors, all experimental results were produced by running the code on our own infrastructure, and all AI-assisted edits to the manuscript were reviewed against the underlying experiments and results before being accepted. We take responsibility for the final content of this work, including text, claims, or artifacts produced with the aid of generative AI.

\section*{Reproducibility Statement}
We provide the details needed to reproduce our experiments throughout the main paper and appendix. Section~\ref{sec:experiment_setup} describes the tasks, models, circuit-discovery methods, and faithfulness metric used in our study. Dataset construction and all task variants are documented in Appendix~\ref{app:dataset}, including the corresponding counterfactuals and evaluation setup. Appendix~\ref{app:mixing} provides the full setup for the task-mixing experiments, while Appendix~\ref{app:clustering} describes the \approachC{} pipeline, hyperparameter choices, clustering variants, and selection of $K$. Additional implementation details, transfer criteria, noise-floor estimation, and complete experimental results are also provided in the appendix. We will release the code used to construct the datasets, run circuit discovery, perform DCD-Cluster, and reproduce all figures and tables.

\bibliography{main}

@inproceedings{mamidanna2025all,
  title={All for one: Llms solve mental math at the last token with information transferred from other tokens},
  author={Mamidanna, Siddarth and Rai, Daking and Yao, Ziyu and Zhou, Yilun},
  booktitle={Proceedings of the 2025 Conference on Empirical Methods in Natural Language Processing},
  pages={30735--30748},
  year={2025}
}

@article{olah2020zoom,
  author = {Olah, Chris and Cammarata, Nick and Schubert, Ludwig and Goh, Gabriel and Petrov, Michael and Carter, Shan},
  title = {Zoom In: An Introduction to Circuits},
  journal = {Distill},
  year = {2020},
  note = {https://distill.pub/2020/circuits/zoom-in},
  doi = {10.23915/distill.00024.001}
}

@article{rai2024practical,
  title={A practical review of mechanistic interpretability for transformer-based language models},
  author={Rai, Daking and Zhou, Yilun and Feng, Shi and Saparov, Abulhair and Yao, Ziyu},
  journal={arXiv preprint arXiv:2407.02646},
  year={2024}
}

@inproceedings{
wang2023interpretability,
title={Interpretability in the Wild: a Circuit for Indirect Object Identification in {GPT}-2 Small},
author={Kevin Ro Wang and Alexandre Variengien and Arthur Conmy and Buck Shlegeris and Jacob Steinhardt},
booktitle={The Eleventh International Conference on Learning Representations },
year={2023},
url={https://openreview.net/forum?id=NpsVSN6o4ul}
}

@article{chughtai2024summing,
  title={Summing up the facts: Additive mechanisms behind factual recall in llms},
  author={Chughtai, Bilal and Cooney, Alan and Nanda, Neel},
  journal={arXiv preprint arXiv:2402.07321},
  year={2024}
}

@article{bereska2024mechanistic,
title={Mechanistic Interpretability for {AI} Safety - A Review},
author={Leonard Bereska and Stratis Gavves},
journal={Transactions on Machine Learning Research},
issn={2835-8856},
year={2024},
url={https://openreview.net/forum?id=ePUVetPKu6},
note={Survey Certification, Expert Certification}
}

@article{zhang2023towards,
  title={Towards best practices of activation patching in language models: Metrics and methods},
  author={Zhang, Fred and Nanda, Neel},
  journal={arXiv preprint arXiv:2309.16042},
  year={2023}
}

@inproceedings{syed2024attribution,
  title={Attribution patching outperforms automated circuit discovery},
  author={Syed, Aaquib and Rager, Can and Conmy, Arthur},
  booktitle={Proceedings of the 7th BlackboxNLP Workshop: Analyzing and Interpreting Neural Networks for NLP},
  pages={407--416},
  year={2024}
}

@inproceedings{
hanna2024have,
title={Have Faith in Faithfulness: Going Beyond Circuit Overlap When Finding Model Mechanisms},
author={Michael Hanna and Sandro Pezzelle and Yonatan Belinkov},
booktitle={First Conference on Language Modeling},
year={2024},
url={https://openreview.net/forum?id=TZ0CCGDcuT}
}

@inproceedings{conmy2023acdc,
 author = {Conmy, Arthur and Mavor-Parker, Augustine and Lynch, Aengus and Heimersheim, Stefan and Garriga-Alonso, Adri\`{a}},
 booktitle = {Advances in Neural Information Processing Systems},
 editor = {A. Oh and T. Naumann and A. Globerson and K. Saenko and M. Hardt and S. Levine},
 pages = {16318--16352},
 publisher = {Curran Associates, Inc.},
 title = {Towards Automated Circuit Discovery for Mechanistic Interpretability},
 url = {https://proceedings.neurips.cc/paper_files/paper/2023/file/34e1dbe95d34d7ebaf99b9bcaeb5b2be-Paper-Conference.pdf},
 volume = {36},
 year = {2023}
}

@article{olsson2022context,
   title={In-context Learning and Induction Heads},
   author={Olsson, Catherine and Elhage, Nelson and Nanda, Neel and Joseph, Nicholas and DasSarma, Nova and Henighan, Tom and Mann, Ben and Askell, Amanda and Bai, Yuntao and Chen, Anna and Conerly, Tom and Drain, Dawn and Ganguli, Deep and Hatfield-Dodds, Zac and Hernandez, Danny and Johnston, Scott and Jones, Andy and Kernion, Jackson and Lovitt, Liane and Ndousse, Kamal and Amodei, Dario and Brown, Tom and Clark, Jack and Kaplan, Jared and McCandlish, Sam and Olah, Chris},
   year={2022},
   journal={Transformer Circuits Thread},
   note={https://transformer-circuits.pub/2022/in-context-learning-and-induction-heads/index.html}
}

@article{qwen2024qwen2,
  title={Qwen2. 5 technical report},
  author={Qwen, A Yang and Yang, Baosong and Zhang, B and Hui, B and Zheng, B and Yu, B and Li, Chengpeng and Liu, D and Huang, F and Wei, H and others},
  journal={arXiv preprint},
  year={2024}
}

@article{dubey2024llama,
  title={The llama 3 herd of models},
  author={Dubey, Abhimanyu and Jauhri, Abhinav and Pandey, Abhinav and Kadian, Abhishek and Al-Dahle, Ahmad and Letman, Aiesha and Mathur, Akhil and Schelten, Alan and Yang, Amy and Fan, Angela and others},
  journal={arXiv e-prints},
  pages={arXiv--2407},
  year={2024}
}

@article{radford2019language,
  title={Language models are unsupervised multitask learners},
  author={Radford, Alec and Wu, Jeffrey and Child, Rewon and Luan, David and Amodei, Dario and Sutskever, Ilya and others},
  journal={OpenAI blog},
  volume={1},
  number={8},
  pages={9},
  year={2019}
}

@inproceedings{ferrando2024information,
  title={Information flow routes: Automatically interpreting language models at scale},
  author={Ferrando, Javier and Voita, Elena},
  booktitle={Proceedings of the 2024 Conference on Empirical Methods in Natural Language Processing},
  pages={17432--17445},
  year={2024}
}

@inproceedings{
chen2026all,
title={All Circuits Lead to Rome: Rethinking Functional Anisotropy in Circuit and Sheaf Discovery for {LLM}s},
author={Xi Chen and Mingyu Jin and Jingcheng Niu and Yutong Yin and Jinman Zhao and Bangwei Guo and Dimitris N. Metaxas and Zhaoran Wang and Yutao Yue and Gerald Penn},
booktitle={Forty-third International Conference on Machine Learning},
year={2026},
url={https://openreview.net/forum?id=3uC9teMlUt}
}

@article{li2026much,
  title={How Much Do Circuits Tell Us? Measuring the Consistency and Specificity of Language Model Circuits},
  author={Li, Michael and Subramani, Nishant},
  journal={arXiv preprint arXiv:2605.08348},
  year={2026}
}

@inproceedings{
rai2025failure,
title={Failure by Interference: Language Models Make Balanced Parentheses Errors  When Faulty Mechanisms Overshadow Sound Ones},
author={Daking Rai and Samuel Miller and Kevin Moran and Ziyu Yao},
booktitle={The Thirty-ninth Annual Conference on Neural Information Processing Systems},
year={2025},
url={https://openreview.net/forum?id=1t4hR9JCcS}
}

@inproceedings{nikankin2025same,
  title={Same Task, Different Circuits: Disentangling Modality-Specific Mechanisms in VLMs},
  author={Nikankin, Yaniv and Arad, Dana and Gandelsman, Yossi and Belinkov, Yonatan},
  booktitle={Mechanistic Interpretability Workshop at NeurIPS 2025},
  year={2025}
}

@article{tibshirani2001estimating,
  title={Estimating the number of clusters in a data set via the gap statistic},
  author={Tibshirani, Robert and Walther, Guenther and Hastie, Trevor},
  journal={Journal of the royal statistical society: series b (statistical methodology)},
  volume={63},
  number={2},
  pages={411--423},
  year={2001},
  publisher={Wiley Online Library}
}

@article{franco2026finding,
  title={Finding Highly Interpretable Prompt-Specific Circuits in Language Models},
  author={Franco, Gabriel and Tassis, Lucas M and Rohr, Azalea and Crovella, Mark},
  journal={arXiv preprint arXiv:2602.13483},
  year={2026}
}

@article{hanna2023does,
  title={How does GPT-2 compute greater-than?: Interpreting mathematical abilities in a pre-trained language model},
  author={Hanna, Michael and Liu, Ollie and Variengien, Alexandre},
  journal={Advances in Neural Information Processing Systems},
  volume={36},
  pages={76033--76060},
  year={2023}
}

@inproceedings{
nikankin2025arithmetic,
title={Arithmetic Without Algorithms: Language Models Solve Math with a Bag of Heuristics},
author={Yaniv Nikankin and Anja Reusch and Aaron Mueller and Yonatan Belinkov},
booktitle={The Thirteenth International Conference on Learning Representations},
year={2025},
url={https://openreview.net/forum?id=O9YTt26r2P}
}

@article{elhage2021mathematical,
   title={A Mathematical Framework for Transformer Circuits},
   author={Elhage, Nelson and Nanda, Neel and Olsson, Catherine and Henighan, Tom and Joseph, Nicholas and Mann, Ben and Askell, Amanda and Bai, Yuntao and Chen, Anna and Conerly, Tom and DasSarma, Nova and Drain, Dawn and Ganguli, Deep and Hatfield-Dodds, Zac and Hernandez, Danny and Jones, Andy and Kernion, Jackson and Lovitt, Liane and Ndousse, Kamal and Amodei, Dario and Brown, Tom and Clark, Jack and Kaplan, Jared and McCandlish, Sam and Olah, Chris},
   year={2021},
   journal={Transformer Circuits Thread},
   note={https://transformer-circuits.pub/2021/framework/index.html}
}

@inproceedings{
feng2024how,
title={How do Language Models Bind Entities in Context?},
author={Jiahai Feng and Jacob Steinhardt},
booktitle={The Twelfth International Conference on Learning Representations},
year={2024},
url={https://openreview.net/forum?id=zb3b6oKO77}
}

@inproceedings{
prakash2024finetuning,
title={Fine-Tuning Enhances Existing Mechanisms: A Case Study on Entity Tracking},
author={Nikhil Prakash and Tamar Rott Shaham and Tal Haklay and Yonatan Belinkov and David Bau},
booktitle={The Twelfth International Conference on Learning Representations},
year={2024},
url={https://openreview.net/forum?id=8sKcAWOf2D}
}

@inproceedings{
marks2025sparse,
title={Sparse Feature Circuits: Discovering and Editing Interpretable Causal Graphs in Language Models},
author={Samuel Marks and Can Rager and Eric J Michaud and Yonatan Belinkov and David Bau and Aaron Mueller},
booktitle={The Thirteenth International Conference on Learning Representations},
year={2025},
url={https://openreview.net/forum?id=I4e82CIDxv}
}

@article{ferrando2024primer,
  title={A primer on the inner workings of transformer-based language models},
  author={Ferrando, Javier and Sarti, Gabriele and Bisazza, Arianna and Costa-Juss{\`a}, Marta R},
  journal={arXiv preprint arXiv:2405.00208},
  year={2024}
}

@article{davies2023discovering,
  title={Discovering Variable Binding Circuitry with Desiderata},
  author={Davies, Xander and Nadeau, Max and Prakash, Nikhil and Shaham, Tamar Rott and Bau, David},
  journal={CoRR},
  year={2023}
}

@article{gur2025mixing,
  title={Mixing Mechanisms: How Language Models Retrieve Bound Entities In-Context},
  author={Gur-Arieh, Yoav and Geva, Mor and Geiger, Atticus},
  journal={arXiv preprint arXiv:2510.06182},
  year={2025}
}

@inproceedings{stolfo2023mechanistic,
  title={A Mechanistic Interpretation of Arithmetic Reasoning in Language Models using Causal Mediation Analysis},
  author={Stolfo, Alessandro and Belinkov, Yonatan and Sachan, Mrinmaya},
  booktitle={Proceedings of the 2023 Conference on Empirical Methods in Natural Language Processing},
  pages={7035--7052},
  year={2023}
}

@inproceedings{
mueller2025mib,
title={{MIB}: A Mechanistic Interpretability Benchmark},
author={Aaron Mueller and Atticus Geiger and Sarah Wiegreffe and Dana Arad and Iv{\'a}n Arcuschin and Adam Belfki and Yik Siu Chan and Jaden Fried Fiotto-Kaufman and Tal Haklay and Michael Hanna and Jing Huang and Rohan Gupta and Yaniv Nikankin and Hadas Orgad and Nikhil Prakash and Anja Reusch and Aruna Sankaranarayanan and Shun Shao and Alessandro Stolfo and Martin Tutek and Amir Zur and David Bau and Yonatan Belinkov},
booktitle={Forty-second International Conference on Machine Learning},
year={2025},
url={https://openreview.net/forum?id=sSrOwve6vb}
}
\bibliographystyle{iclr2027_conference}

\appendix
\section{Dataset construction}
\label{app:dataset}

In this section, we provide details on the dataset construction across all four tasks,
representative example prompts (\cref{app:dataset-examples}), and the accuracy of each model
on each variant (\cref{app:dataset-accuracy}). In our circuit study, we construct multiple
datasets for each task that vary along one of three axes --- \emph{complexity},
\emph{syntax}, or \emph{domain} --- while holding all other properties fixed. Two tasks
vary along a single axis in more than one way, and we treat these as \emph{sub-axes}:
IOI syntax separates word order (\textsc{abba} vs.\ \textsc{baba}) from voice (active vs.\
passive), and entity-binding syntax separates the queried position from the delimiter.
Variants are compared within a sub-axis and not across, since two variants drawn from
different sub-axes differ in more than one respect. \Cref{tab:variant-inventory} lists the $32$ variants and the axis each one varies.

\begin{table}[H]
\centering\small
\caption{The $32$ dataset variants, grouped by the axis along which each differs from
its task's base variant. Variants that anchor more than one axis (2-person IOI, 2-comma
entity binding, 2-operand arithmetic) are listed once, under complexity. The last column
points to the table of example prompts for that task, with its clean and counterfactual
inputs.}

\label{tab:variant-inventory}
\begin{tabular}{llp{0.42\textwidth}rc}
\toprule
Task & Axis & Variants & $n$ & Examples \\
\midrule
\multirow{3}{*}{IOI}
 & Complexity & 2-person, 3-person, filler & 3 & \multirow{3}{*}{Table~\ref{tab:ioi-distributions}} \\
 & Syntax     & \textsc{abba}, \textsc{baba} (word order); passive (voice) & 3 & \\
 & Domain     & letter-label & 1 & \\
\midrule
\multirow{4}{*}{Entity binding}
 & Complexity & 2-, 3-, 4-, 5-, 6-, 7-, 8-comma & 7 & \multirow{4}{*}{Table~\ref{tab:entity-binding-distributions}} \\
 & Syntax     & queried position $P_1$--$P_8$ & 8 & \\
 & Syntax     & 2-period (delimiter) & 1 & \\
 & Domain     & 2-colour-box & 1 & \\
\midrule
\multirow{2}{*}{Arithmetic}
 & Complexity & 2-operand, 3-operand & 2 & \multirow{2}{*}{Table~\ref{tab:arithmetic-distributions}} \\
 & Syntax     & verbal v1 (\emph{addition of}), v2 (\emph{sum of}), v3 (\emph{how much is}) & 3 & \\
\midrule
Sequence completion & Complexity & 2-gram, 3-gram, 4-gram symbolic & 3 & Table~\ref{tab:sequence-distributions} \\
\midrule
\multicolumn{3}{l}{\textbf{Total}} & \textbf{32} & \\
\bottomrule
\end{tabular}
\end{table}

\subsection{Model accuracy by task and prompt type}
\label{app:dataset-accuracy}

Table~\ref{tab:accuracy} reports raw per-model accuracy for all task and prompt-type combinations used in our experiments.

\begin{table}[htbp]
\centering
\caption{Per-model accuracy (\%) across all task and prompt-type variants (raw accuracy over 500 examples per condition).}
\label{tab:accuracy}
\small
\begin{tabular}{llccc}
\toprule
\textbf{Task} & \textbf{Prompt Type} & \textbf{GPT-2} & \textbf{Qwen-7B-Inst.} & \textbf{Llama-8B-Inst.} \\
\midrule
\multirow{6}{*}{IOI}
 & ABBA (IO-First)          & 98.8  & 99.4  & 100.0 \\
 & BABA (Subject-First)          & 99.6  & 99.2  & 100.0 \\
 & Filler        & 68.4  & 99.8  & 100.0 \\
 & Letter        & 99.8  & 100.0 & 100.0 \\
 & Passive       & 94.8  & 97.8  & 100.0 \\
 & 3-Person      & 60.8  & 98.2  & 100.0 \\
\midrule
\multirow{10}{*}{Entity-Binding}
 & 2-Comma       & 50.4  & 67.8  & 96.2  \\
 & 3-Comma       & 37.4  & 79.6  & 95.0  \\
 & 5-Comma       & 22.6  & 98.2  & 95.0  \\
 & P1 & 43.4 & 96.2  & 99.8  \\
 & P4 & 10.6 & 98.4  & 98.6  \\
 & P7 & 11.2 & 99.2  & 96.8  \\
 & 2-Period      & 50.0  & 64.0  & 92.8  \\
 & 8-Period      & 14.0  & 98.0  & 98.2  \\
 & 2-Color-Box   & 54.6  & 57.0  & 89.2  \\
 & 8-Color-Box   & 17.2  & 98.8  & 98.0  \\
\midrule
\multirow{5}{*}{Arithmetic}
 & 2-Operand     & 0.0   & 28.6  & 97.2  \\
 & 3-Operand     & 0.0   & 34.9  & 96.8  \\
 & Verbal v1     & 0.0   & 19.6  & 99.8  \\
 & Verbal v2     & 0.0   & 85.0  & 99.8  \\
 & Verbal v3     & 0.0   & 17.2  & 99.6  \\
\midrule
\multirow{3}{*}{Seq.\ Completion}
 & 2-Gram        & 100.0 & 65.8  & 100.0 \\
 & 3-Gram        & 86.4  & 60.8  & 94.0  \\
 & 4-Gram        & 65.0  & 52.4  & 80.8  \\
\bottomrule
\end{tabular}
\end{table}

\subsection{Task Axes}
Each task consists of three axes: Complexity (Appendix~\ref{app:dataset-complexity}), Domain (Appendix~\ref{app:dataset-domain}), and Syntax (Appendix~\ref{app:dataset-syntax}).
\subsubsection{Complexity}
\label{app:dataset-complexity}

We scale the number of candidate entities the model must disambiguate among, while holding the task structure and surface form fixed.

\begin{itemize}
    \item \textbf{IOI.} The 2-person variant contains two names (one indirect object, one subject). The 3-person variant contains three names (one indirect object and two subjects who perform the action together), increasing the number of candidates the model must track. Prompts are drawn from a fixed template; only the number of names varies. In addition, we also construct \emph{filler} dataset, where we simply add the filler sentence that consists of an indirect object to increase the difficulty. 
    \item \textbf{Entity binding.} We construct datasets with $n \in \{2, 3, 4, 5, 6, 7, 8\}$ entity-attribute pairs. Each prompt lists $n$ (entity, box) pairs and queries the entity associated with one specific box. Entities are sampled without replacement from a fixed vocabulary; box labels are sampled from a fixed label set.
    \item \textbf{Arithmetic.} The 2-operand variant uses expressions of the form \texttt{a + b =}, and the 3-operand variant uses \texttt{a + b + c =}. Operands are sampled from a fixed numeric range; the final answer is held constant across variants where possible (e.g., \texttt{10 + 50 = 60} vs.\ \texttt{10 + 30 + 20 = 60}) to isolate the effect of operand count.
    \item \textbf{Sequence completion.} We construct 2-gram, 3-gram, and 4-gram variants. For $k$-gram sequences, a base sequence of length $k$ is repeated multiple times, and the model must predict the next token given a truncated final repetition. Larger $k$ requires the model to attend further back to identify the repeating pattern.
\end{itemize}

\subsubsection{Syntax}
\label{app:dataset-syntax}

We vary the surface structure of the prompt --- word order or target position --- while holding task semantics and entity identities fixed.

\begin{itemize}
    \item \textbf{IOI.} We construct two pairs of datasets, treated as two sub-axes within syntax.
    \begin{itemize}
        \item \emph{Active vs.\ passive.} The active variant uses constructions of the form ``\textit{John gave a drink to Mary},'' while the passive variant uses ``\textit{John was given a drink by Mary}.'' The semantic roles (giver, recipient) are preserved; only the syntactic realization differs.
        \item \emph{IO-first (ABBA) vs.\ subject-first (BABA).} In the IO-first variant, the indirect object is mentioned before the subject in the opening clause (e.g., ``\textit{When Mary and John went to the store...}''); in the subject-first variant, the subject appears first (``\textit{When John and Mary went to the store...}''). The task, names, and action remain fixed; only the order of first mention changes.
    \end{itemize}
    \item \textbf{Entity binding (positional).} Using a fixed template of $n=8$ entity-attribute pairs, we construct 8 datasets that differ only in the position of the target pair. In the $k$-th dataset ($k \in \{1, \ldots, 8\}$), the target entity-attribute pair always appears in position $k$ of the list, while the remaining seven positions are filled with distractor pairs sampled from the same vocabulary. This isolates whether the discovered circuit captures a position-invariant binding mechanism or relies on the target's location within the prompt.
    \item \textbf{Arithmetic (phrasing).} We construct multiple datasets that pose the same arithmetic operation using different natural-language phrasings while keeping the operands and answer identical. The phrasings are: ``\textit{What is the addition of 20 and 40? Answer:}'', ``\textit{Question: What is the sum of 20 and 40? Answer:}'', ``\textit{Question: How much is 20 plus 40? Answer:}'', and ``\textit{Q: What is the result of 10 plus 50? A:}''. Each phrasing represents a distinct surface realization of the underlying addition task.
\end{itemize}

\subsubsection{Domain}
\label{app:dataset-domain}

We vary the lexical domain of the entities in the prompt while holding task structure and difficulty fixed.

\begin{itemize}
    \item \textbf{IOI.} The \emph{person-name} variant uses real first names (e.g., ``\textit{Mary},'' ``\textit{John}'') drawn from a fixed name vocabulary. The \emph{letter-label} variant replaces names with placeholder identifiers of the form ``\textit{Person X},'' ``\textit{Person Y}''. All other aspects of the prompt --- sentence structure, action verb, and object --- are held constant.
    \item \textbf{Entity binding.} The \emph{letter-label} variant identifies boxes with single-letter labels (e.g., ``\textit{box D},'' ``\textit{box C}''). The \emph{color-adjective} variant identifies boxes with color descriptors (e.g., ``\textit{purple box},'' ``\textit{orange box}''). The entities to be bound (medicine, rose, etc.) and the prompt template are held constant; only the box-identification scheme changes.
    \item \textbf{Arithmetic.} The \emph{numerical} variant uses purely symbolic notation (e.g., \texttt{10 + 50 =}). The \emph{verbal} variant uses the natural-language phrasings described in Section~\ref{app:dataset-syntax} (Arithmetic). Operands and answers are held constant across variants.
\end{itemize}

\subsection{Dataset Examples} 
\label{app:dataset-examples}

\subsubsection{Indirect Object Identification}
Table~\ref{tab:ioi-distributions} shows representative prompts of the IOI task for each (axis, variant) combination, with the correct and counterfactual labels used for circuit discovery. In IOI, each input consists of a subject and an indirect object (e.g., \emph{``When Mary and John went to the store, John gave an apple to''}), and the model is expected to predict the indirect object (\emph{Mary}) as the next token. IOI has been extensively studied in prior circuit discovery work~\citep{wang2023interpretability, conmy2023acdc, hanna2024have, mueller2025mib}.

\begin{table}[h!]
\centering
\caption{Input examples for Indirect Object Identification (IOI) across the three dataset-construction axes, showing example prompts with their correct and counterfactual labels.}
\label{tab:ioi-distributions}
\vspace{0.5em}
\begin{tabular}{p{2.5cm} p{7cm} c c}
\toprule
\textbf{Distribution} & \textbf{Prompt} & \textbf{Correct} & \textbf{Counter.} \\
\hline
\multicolumn{4}{c}{\textbf{Complexity}} \\
\hline
2-person
& \textit{When Mary and John went to the store, John gave a drink to}
& Mary
& John \\
\midrule
3-person 
& \textit{When Mary, John, and Sam went to the store, John and Mary gave a drink to}
& Sam
& John, Mary \\
\midrule
Filler text
& \textit{When Mary and John went to the store, Mary was talking on the phone. John gave a drink to}
& Mary
& John \\
\hline
\multicolumn{4}{c}{\textbf{Syntax}} \\
\hline
Active
& \textit{When Mary and John went to the store, John gave a drink to}
& Mary
& John \\
\midrule
Passive
& \textit{When Mary and John went to the store, John was given drink by}
& Mary
& John \\
\midrule
ABBA (IO First)
& \textit{When Mary and John went to the store, John gave a drink to}
& Mary
& John \\
\midrule
BABA (Subject First)
& \textit{When John and Mary went to the store, John gave a drink to}
& Mary
& John \\
\hline
\multicolumn{4}{c}{\textbf{Domain}} \\
\hline
2-person
& \textit{When Mary and John went to the store, John gave a drink to}
& Mary
& John \\
\midrule
letter
& \textit{When Person X and Person Y went to the store, Person X gave a drink to Person}
& Y
& X \\
\bottomrule
\end{tabular}
\end{table}

\subsubsection{Entity Binding}
Table~\ref{tab:entity-binding-distributions} shows representative prompts of the entity binding task for each (axis, variant) combination, with the correct and counterfactual labels used for circuit discovery. In entity binding, each input introduces a list of (entity, container) pairs and queries the entity associated with one specific container (e.g., \emph{``The medicine is in box D, the rose is in box C. Box D contains the''}), and the model is expected to predict the bound entity (\emph{medicine}) as the next token. Entity binding has been studied in prior circuit and mechanistic interpretability work~\citep{davies2023discovering, feng2024how, prakash2024finetuning, mueller2025mib, gur2025mixing}.
\begin{table}[htbp]
\centering
\caption{Input examples for Entity Binding across the three dataset-construction axes.}
\label{tab:entity-binding-distributions}
\vspace{0.5em}
\small
\setlength{\tabcolsep}{4pt}
\renewcommand{\arraystretch}{1.15}
\begin{tabularx}{\linewidth}{
  >{\raggedright\arraybackslash}p{1.8cm}
  >{\raggedright\arraybackslash}X
  >{\raggedright\arraybackslash}p{1.8cm}
  >{\raggedright\arraybackslash}X
}
\toprule
\textbf{Distribution} & \textbf{Prompt} & \textbf{Correct} & \textbf{Counter.} \\
\midrule

\multicolumn{4}{c}{\textbf{Complexity}} \\
\midrule
2-Entity
& \textit{The medicine is in box D, the rose is in box C. Box D contains the}
& medicine
& rose\\

\midrule
3-Entity
& \textit{The medicine is in box D, the rose is in box C, the apple is in box R. Box C contains the}
& rose
& medicine, apple\\

\midrule
4-Entity
& \textit{The medicine is in box D, the rose is in box C, the apple is in box R, the plate is in box A. Box C contains the}
& rose
& medicine, apple, plate \\

\midrule
\multicolumn{4}{c}{\textbf{Syntax}} \\
\midrule
P$1$
& \textit{The fan is in box E, the computer is in box H, the tie is in box F, the bomb is in box Z, the coat is in box R, the plant is in box W, the pot is in box L, the egg is in box V. Box E contains the}
& fan
& computer, tie, bomb, coat, plant, pot, egg\\

\midrule
P$2$
& \textit{The fan is in box E, the computer is in box H, the tie is in box F, the bomb is in box Z, the coat is in box R, the plant is in box W, the pot is in box L, the egg is in box V. Box H contains the}
& computer
& fan, tie, bomb, coat, plant, pot, egg\\

\midrule
P$3$
& \textit{The fan is in box E, the computer is in box H, the tie is in box F, the bomb is in box Z, the coat is in box R, the plant is in box W, the pot is in box L, the egg is in box V. Box F contains the}
& tie
& computer, fan, bomb, coat, plant, pot, egg\\

\midrule
\multicolumn{4}{c}{\textbf{Domain}} \\
\midrule
Alphabet Box
& \textit{The medicine is in box D, the rose is in box C. Box D contains the}
& medicine
& rose\\

\midrule
Color Box
& \textit{The medicine is in purple box, the rose is in in orange box. Purple box contains the}
& medicine
& rose\\

\bottomrule
\end{tabularx}
\end{table}

\subsubsection{Arithmetic}
Table~\ref{tab:arithmetic-distributions} shows representative prompts of the arithmetic task for each (axis, variant) combination, with the correct and counterfactual labels used for circuit discovery. In arithmetic, each input poses a multi-operand addition problem in either symbolic or natural-language form (e.g., \emph{``10 + 50 =''}), and the model is expected to predict the numerical answer (\emph{60}) as the next token. Following prior work~\citep{stolfo2023mechanistic, nikankin2025arithmetic}, we restrict operands and answers to single-token values to avoid confounding the discovered circuit with multi-token decoding behavior. We conduct arithmetic experiments only on Llama-3.1-8B-Instruct, since GPT-2 and Qwen2.5-7B tokenizers do not represent most two- and three-digit numbers as single tokens, whereas Llama does. The task has been studied in prior mechanistic interpretability work~\citep{stolfo2023mechanistic, nikankin2025arithmetic, mamidanna2025all, mueller2025mib}.

\begin{table}[htbp]
\centering
\caption{Input examples for Arithmetic across the dataset-construction axes.}
\label{tab:arithmetic-distributions}
\vspace{0.5em}
\begin{tabular}{p{2.0cm} p{7cm} c c}
\toprule
\textbf{Distribution} & \textbf{Prompt} & \textbf{Correct} & \textbf{Counter.} \\
\hline
\multicolumn{4}{c}{\textbf{Complexity}} \\
\hline
Two-operand
& \textit{10 + 50 =}
& 60
& $[0, 999] \setminus \{60\}$\\
\midrule
Three-operand
& \textit{10 + 30 + 20 =}
& 60
& $[0, 999] \setminus \{60\}$ \\
\hline
\multicolumn{4}{c}{\textbf{Syntax (Phrasing)}} \\
\hline
Verbal phrasing 1
& \textit{What is the addition of 20 and 40? Answer:}
& 60
& $[0, 999] \setminus \{60\}$ \\
\midrule
Verbal phrasing 2
& \textit{Question: What is the sum of 20 and 40? Answer:}
& 60
& $[0, 999] \setminus \{60\}$ \\
\midrule
Verbal phrasing 3
& \textit{Question: How much is 20 plus 40? Answer:}
& 60
& $[0, 999] \setminus \{60\}$ \\
\hline
\multicolumn{4}{c}{\textbf{Domain}} \\
\hline
Verbal phrasing 1
& \textit{What is the addition of 20 and 40? Answer:}
& 60
& $[0, 999] \setminus \{60\}$ \\
\hline
Two-operand (Numerical)
& \textit{10 + 50 =}
& 60
& $[0, 999] \setminus \{60\}$ \\

\bottomrule
\end{tabular}
\end{table}

\subsubsection{Sequence Completion}
Table~\ref{tab:sequence-distributions} shows representative prompts of the sequence completion task for each (axis, variant) combination, with the correct and counterfactual labels used for circuit discovery. In sequence completion, each input contains a base sequence of $k$ tokens repeated multiple times with the final repetition truncated (e.g., \emph{``David Mary John Sam Rachel David Mary John Sam Rachel David Mary John Sam''}), and the model is expected to predict the next token (\emph{Rachel}) by matching the current context against earlier occurrences of the same sub-sequence. Sequence completion is the canonical task for studying induction-head circuits, which were originally identified as a basic primitive of in-context learning~\citep{elhage2021mathematical, olsson2022context}.

\begin{table}[htbp]
\centering
\setlength{\tabcolsep}{4pt}
\caption{Input examples for Sequence Completion across the Complexity axis.}
\label{tab:sequence-distributions}
\vspace{0.5em}
\resizebox{\linewidth}{!}{%
\begin{tabular}{p{2.0cm} p{6cm} p{2.0cm} p{3.0cm}}
\toprule
\textbf{Distribution} & \textbf{Prompt} & \textbf{Correct} & \textbf{Counter.} \\
\hline
\multicolumn{4}{c}{\textbf{Complexity}} \\
\hline
2-gram sequence
& \textit{David Mary John Sam Rachel David Mary John Sam Rachel David Mary John Sam}
& Rachel
& David, Mary, John, Sam \\
\midrule
3-gram sequence
& \textit{David Mary Hanna Sam Tom David Mary John Sam Rachel David Mary Hanna Sam}
& Tom
& Mary, Mary, John, Sam, David \\
\midrule
4-gram sequence
& \textit{David Hanna John Sam Tom David Mary John Sam Rachel David Hanna John Sam}
& Tom
& Mary, Mary, John, Sam, David \\
\bottomrule
\end{tabular}}
\end{table}

\subsection{Counterfactual prompt construction}
\label{app:counterfactual-prompts}

Circuit discovery algorithms like EAP-IG, EAP, E-ACT, and \approach require, for every clean
prompt $x_i$, a paired \emph{corrupted} prompt $x_i'$ whose forward-pass activations replace the clean activations along an interpolation path. Across all four tasks, the corrupted prompt is built from the clean prompt by a minimal, task-specific perturbation that breaks
the discriminative cue while preserving token length so that EAP-IG's
position-aligned activation patching is well-defined; \approach{} uses
identical clean/corrupted pairs whether the attribution is computed
per-example (Section~\ref{exp3: our-approach}).
Random sampling inside each rule is seeded (\texttt{random\_seed}\,$=32$)
so that pairs are reproducible.

\textbf{IOI.} For every IOI variant (ABBA, BABA, mixed, filler, letter,
passive), the corrupted prompt is the canonical \textsc{abc} variant of
\citet{wang2023interpretability}: replace the second occurrence of the
subject $S$ in the clean prompt with a name sampled uniformly from the
fixed name vocabulary, excluding the original $S$ and indirect object
$\mathit{IO}$. The 3-person variant departs from the single-token swap:
the trailing list of two non-$\mathit{IO}$ subjects is replaced wholesale
with two newly sampled names disjoint from every name appearing in the
clean prompt. The template body, name vocabulary, and prompt length are
otherwise unchanged.

\textbf{Entity binding.} The corrupted prompt keeps the entity-attribute
list verbatim and replaces only the trailing query identifier with one
drawn uniformly from the corresponding pool excluding identifiers already
used in the prompt. For letter-labelled variants (\textit{comma},
\textit{period}, \textit{colon}, \textit{instruct}, \textit{position}-$k$,
$n$-\textit{comma}) the swap is over single-letter labels; for
color-labelled variants (\textit{color-box-comma}) it is over colour
adjectives; for the country variant it is over country names. The
position-axis variants ($P_1, \ldots, P_8$) hold the entity list and swap
rule fixed and differ only in which position is queried in the clean
prompt.

\textbf{Arithmetic.} For both numerical (\textit{2-operand},
\textit{3-operand}) and verbal phrasings
(\textit{verbal-v1/v2/v3}), the corrupted prompt is an unrelated
expression of the same surface structure and matched token length, drawn
deterministically from the next entry of the (seeded, pre-shuffled) prompt
list. The shuffle decouples operand identities, so the corrupted answer
carries no structural relation to the clean answer; the corrupted prompt's
role is purely to provide off-distribution activations for patching.

\textbf{Sequence completion.} For $k$-gram variants
($k \in \{2, 3, 4\}$), the corrupted prompt replaces the discriminative
tail of the repeated pattern with fresh tokens that are not in the clean
sequence, applied at \emph{every} repetition rather than only the final
one, so that the induction signal is broken throughout. For the 2-gram
pattern $[A\,B\,C\,D\,E\,F]$ the last two tokens $[E\,F]$ are replaced;
for the 3- and 4-gram patterns $[A\,B\,C\,X\,E\,F]$ /
$[A\,B\,X\,C\,E\,F]$, the discriminative interior token $X$ and the final
token $F$ are both replaced. The interleaved $[\ldots Y \ldots]$ patterns
used as foils are left intact.

We provide concrete clean/corrupted pair per task in Table~\ref{tab:counterfactual-prompts}. 

\begin{table}[t!]
\centering
\small
\setlength{\tabcolsep}{4pt}
\caption{One representative clean/corrupted prompt pair per task. The
corrupted prompt is obtained by the task-specific minimal perturbation
described above. The placeholder ``\textsc{rand}'' denotes a name sampled
uniformly from the IOI name vocabulary excluding $\{\mathit{IO}, S\}$;
``\textsc{cf}$_i$'' denotes a token sampled outside the clean sequence.}
\label{tab:counterfactual-prompts}

\begin{tabular}{l p{5.0cm} p{5.0cm}}
\toprule
\textbf{Task} & \textbf{Clean prompt} & \textbf{Corrupted prompt} \\
\midrule
IOI (ABBA) &
\textit{When Mary and John went to the store, John gave a drink to} &
\textit{When Mary and John went to the store, \textsc{rand} gave a drink to} \\
\midrule
IOI (3-person) &
\textit{When Mary, John, and Sam went to the store, John and Mary gave a drink to} &
\textit{When Mary, John, and Sam went to the store, \textsc{rand}$_1$ and \textsc{rand}$_2$ gave a drink to} \\
\midrule
Entity binding &
\textit{The medicine is in box D, the rose is in box C. Box D contains the} &
\textit{The medicine is in box D, the rose is in box C. Box \textsc{rand}-letter contains the} \\
\midrule
Arithmetic (2-op) &
\textit{10 + 50 =} &
\textit{27 + 13 =} \\
\midrule
Arithmetic (verbal-v1) &
\textit{What is the addition of 20 and 40? Answer:} &
\textit{What is the addition of 31 and 12? Answer:} \\
\midrule
Sequence (2-gram) &
\textit{David Mary John Sam Rachel David Mary John Sam Rachel David Mary John Sam} &
\textit{David Mary John Sam \textsc{cf}$_1$ David Mary John Sam \textsc{cf}$_1$ David Mary John Sam} \\
\midrule
Sequence (3-gram) &
\textit{David Mary Hanna Sam Tom David Mary John Sam Rachel David Mary Hanna Sam} &
\textit{David Mary \textsc{cf}$_1$ Sam \textsc{cf}$_2$ David Mary John Sam Rachel David Mary \textsc{cf}$_1$ Sam} \\
\bottomrule
\end{tabular}
\end{table}
\section{Additional results: cross-task mixing across models and attribution methods}
\label{app:mixing}

This appendix extends the cross-task mixing experiment of Section~\ref{subsec: cross-task-mixing} to every model and attribution method for which we have data. Section~\ref{app:mixing-setup} states the design and the coverage of the experiment, Section~\ref{app:mixing-summary} reports the underlying numbers for all cells in a single table, and Sections~\ref{app:mixing-eapig}--\ref{app:mixing-eact} present one figure per (method, model) cell, grouped by attribution method. The main-text figure (Figure~\ref{fig:exp2}) is the EAP-IG $\times$ Llama-3.1-8B-Instruct cell of this grid.

\subsection{Setup and coverage}
\label{app:mixing-setup}

Unless otherwise noted, we follow these conventions throughout this appendix:

\begin{itemize}
    \item \textbf{Task pairs.} We take the four tasks of Section~\ref{subsec:tasks}, fixing one variant of each (arithmetic: 2-operand addition; entity binding: 2-comma; IOI: ABBA and BABA pooled; sequence completion: 2-gram symbolic), and form all six unordered pairs. Each pair is one panel of a figure.
    \item \textbf{Discovery datasets.} For a pair $(A, B)$ we build discovery datasets by replacing a fraction of the examples of one task with examples of the other, holding the total number of examples fixed at $600$. The figures show three of these datasets---pure $A$, a $50/50$ mixture, and pure $B$---sliced out of a finer sweep in steps of $0.1$; only the three-point design was run for \textsc{E-Act} (see below). The full sweep is shown in Section~\ref{app:mixing-sweep}.
    \item \textbf{Evaluation.} Each discovered circuit is evaluated on the held-out test set of pure $A$ and of pure $B$ separately ($200$ examples each), never on the mixture. The two bars of each group are these two scores.
    \item \textbf{Circuit size and metric.} We report faithfulness at a circuit size of 5\% of edges, using the normalized score of Section~\ref{subsec:faithfulness}. This score is not bounded to $[0,1]$: it exceeds $1$ when the circuit produces a larger logit difference than the full model, and falls below $0$ when the circuit does worse than the counterfactual baseline. Several bars in these figures are negative, and this is the reason.
    \item \textbf{Coverage.} We report a (model, task) cell only where the model solves the task often enough for faithfulness to be meaningful, following the accuracy criterion of Section~\ref{app:hd-setup}. Under this rule GPT-2 contributes only the IOI--SC pair (it scores $0.0\%$ on every arithmetic variant and $10.6$--$54.6\%$ on entity binding, Table~\ref{tab:accuracy}) and Qwen2.5-7B-Instruct contributes the three pairs that exclude arithmetic ($28.6\%$ on 2-operand addition). Llama-3.1-8B-Instruct contributes all six pairs. Arithmetic is additionally affected by tokenization: only Llama represents the operands and results used here as single tokens (Section~\ref{app:dataset}), so on the other two models the target label is a sub-word fragment.
    \item \textbf{Methods.} EAP-IG~\citep{hanna2024have} is the default; EAP~\citep{syed2024attribution} and \textsc{E-Act}~\citep{conmy2023acdc} are method-robustness checks. \textsc{E-Act} requires a forward pass per edge and is run only on GPT-2, on the IOI--SC pair, and at $200$ rather than $600$ examples per circuit; its absolute values are therefore not directly comparable with the gradient-based methods, though the qualitative pattern is.
\end{itemize}

This gives six cells (Table~\ref{tab:mixing-coverage}) covering $20$ (method, model, pair) combinations in total.

\begin{table}[htbp]
\centering
\small
\caption{Coverage of the cross-task mixing experiment. A dash marks a (method, model) cell that was not run; within a cell, the listed pairs are those passing the accuracy criterion of Section~\ref{app:hd-setup} (see the Coverage item above).}
\label{tab:mixing-coverage}
\vspace{0.5em}
\begin{tabular}{llcl}
\toprule
\textbf{Method} & \textbf{Model} & \textbf{Pairs} & \textbf{Figure} \\
\midrule
\multirow{3}{*}{EAP-IG}
  & GPT-2          & 1 & Figure~\ref{fig:mixing-eapig-gpt2} \\
  & Qwen-7B-Inst.  & 3 & Figure~\ref{fig:mixing-eapig-qwen} \\
  & Llama-8B-Inst. & 6 & Figure~\ref{fig:exp2} (main text) \\
\midrule
\multirow{3}{*}{EAP}
  & GPT-2          & --- & --- \\
  & Qwen-7B-Inst.  & 3 & Figure~\ref{fig:mixing-eap-qwen} \\
  & Llama-8B-Inst. & 6 & Figure~\ref{fig:mixing-eap-llama} \\
\midrule
\multirow{3}{*}{\textsc{E-Act}}
  & GPT-2          & 1 & Figure~\ref{fig:mixing-eact-gpt2} \\
  & Qwen-7B-Inst.  & --- & --- \\
  & Llama-8B-Inst. & --- & --- \\
\bottomrule
\end{tabular}
\end{table}


\subsection{Faithfulness of pure and mixed circuits across all cells}
\label{app:mixing-summary}

Table~\ref{tab:mixing-summary} reports the numbers behind every bar in the figures of this appendix and of Figure~\ref{fig:exp2}. For a pair $(A, B)$ it gives the faithfulness of the pure-$A$ circuit, the pure-$B$ circuit and the $50/50$ circuit, each evaluated on $A$ and on $B$.

\begin{table}[htbp]
\centering
\small
\caption{Faithfulness at circuit size $0.05$ for the pure and $50/50$ circuits of every task pair, model and attribution method. Columns give the evaluation task; rows give the discovery dataset. Values are the normalized faithfulness of Section~\ref{subsec:faithfulness} and are not bounded to $[0,1]$.}
\label{tab:mixing-summary}
\vspace{0.5em}
\begin{tabular}{lll cc cc cc}
\toprule
& & & \multicolumn{2}{c}{\textbf{Pure $A$ circuit}} & \multicolumn{2}{c}{\textbf{Pure $B$ circuit}} & \multicolumn{2}{c}{\textbf{$50/50$ circuit}} \\
\cmidrule(lr){4-5} \cmidrule(lr){6-7} \cmidrule(lr){8-9}
\textbf{Method} & \textbf{Model} & \textbf{Pair ($A$--$B$)} & on $A$ & on $B$ & on $A$ & on $B$ & on $A$ & on $B$ \\
\midrule
EAP-IG & GPT-2 & IOI--SC & 0.63 & 0.24 & -0.47 & 0.60 & 0.63 & 0.63 \\
\midrule
\multirow{3}{*}{EAP-IG} & \multirow{3}{*}{Qwen-7B-Inst.}
  & EB--IOI    & 0.80 & 0.18 & 0.47 & 0.94 & 0.82 & 0.89 \\
& & EB--SC     & 0.80 & 0.26 & -0.02 & 0.62 & 1.25 & 0.50 \\
& & IOI--SC    & 0.94 & 0.47 & 0.40 & 0.62 & 0.88 & 0.46 \\
\midrule
\multirow{6}{*}{EAP-IG} & \multirow{6}{*}{Llama-8B-Inst.}
  & Arith--EB  & 0.92 & -0.17 & 0.12 & 0.60 & 0.92 & 0.70 \\
& & Arith--IOI & 0.92 & 0.04 & 0.15 & 1.06 & 0.92 & 0.84 \\
& & Arith--SC  & 0.92 & 0.34 & 0.55 & 0.74 & 0.90 & 0.73 \\
& & EB--IOI    & 0.60 & 0.50 & 0.29 & 1.06 & 0.75 & 0.86 \\
& & EB--SC     & 0.60 & 0.02 & -0.12 & 0.74 & 0.79 & 0.81 \\
& & IOI--SC    & 1.06 & 0.50 & 0.27 & 0.74 & 0.94 & 0.78 \\
\midrule
\multirow{3}{*}{EAP} & \multirow{3}{*}{Qwen-7B-Inst.}
  & EB--IOI    & 0.25 & 0.01 & 0.10 & 0.53 & 0.62 & 0.53 \\
& & EB--SC     & 0.25 & 0.14 & -0.07 & 0.52 & 0.64 & 0.44 \\
& & IOI--SC    & 0.53 & 0.43 & 0.36 & 0.52 & 0.45 & 0.46 \\
\midrule
\multirow{6}{*}{EAP} & \multirow{6}{*}{Llama-8B-Inst.}
  & Arith--EB  & 0.92 & -0.13 & 0.02 & 0.49 & 0.71 & 0.08 \\
& & Arith--IOI & 0.92 & -0.01 & 0.22 & 0.51 & 0.92 & 0.47 \\
& & Arith--SC  & 0.92 & 0.25 & 0.28 & 0.68 & 0.87 & 0.67 \\
& & EB--IOI    & 0.49 & 0.49 & 0.29 & 0.51 & 0.31 & 0.53 \\
& & EB--SC     & 0.49 & -0.10 & -0.25 & 0.68 & 0.42 & 0.60 \\
& & IOI--SC    & 0.51 & 0.62 & 0.19 & 0.68 & 0.50 & 0.73 \\
\midrule
\textsc{E-Act} & GPT-2 & IOI--SC & 0.82 & 0.14 & -0.15 & 0.56 & 0.84 & 0.46 \\
\bottomrule
\end{tabular}
\end{table}

\subsection{EAP-IG}
\label{app:mixing-eapig}

The EAP-IG cells are Figure~\ref{fig:exp2} (Llama-3.1-8B-Instruct, in the main text), Figure~\ref{fig:mixing-eapig-qwen} (Qwen2.5-7B-Instruct) and Figure~\ref{fig:mixing-eapig-gpt2} (GPT-2). The main-text pattern holds in all ten pairs: no $50/50$ circuit falls below $0.46$ on either of its two tasks, while $13$ of the $20$ pure-circuit off-task scores fall below $0.30$.

On Qwen2.5-7B-Instruct the clearest pair is EB--SC. The pure sequence-completion circuit scores $-0.02$ on entity binding, and the pure entity-binding circuit scores $0.26$ on sequence completion; the $50/50$ circuit reaches $1.25$ on entity binding and $0.50$ on sequence completion at the same 5\% budget. On GPT-2 the single available pair, IOI--SC, is the starkest case in the whole grid: the pure sequence-completion circuit is at $-0.47$ on IOI, worse than ablating the model to the counterfactual baseline, while the $50/50$ circuit scores $0.63$ on both tasks. A reader given only the $50/50$ circuit and its faithfulness score would have no way to tell that it covers two tasks whose own circuits are mutually useless.

\begin{figure}[htbp]
    \centering
    \includegraphics[width=\textwidth]{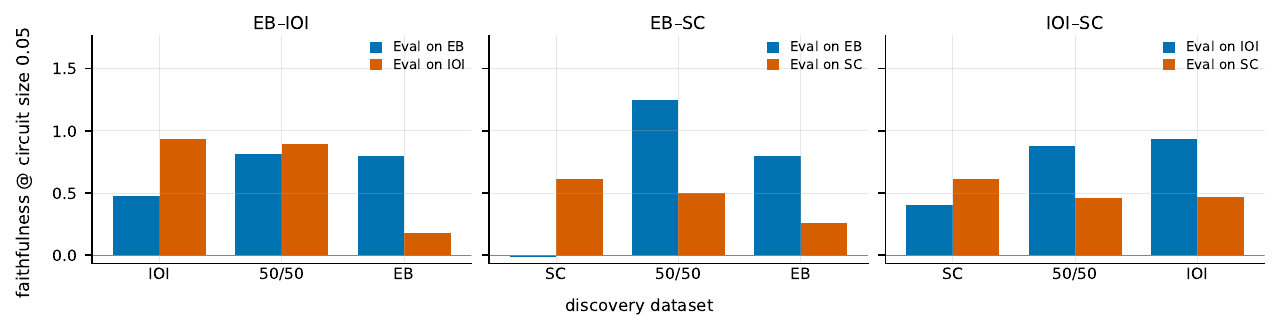}
    \caption{Cross-task mixing with EAP-IG on Qwen2.5-7B-Instruct. One panel per task pair; within a panel the $x$-axis is the discovery dataset (pure, $50/50$, pure) and the two bars are faithfulness at circuit size $0.05$ on each of the two tasks. Arithmetic pairs are omitted because Qwen2.5 scores $28.6\%$ on 2-operand addition (Table~\ref{tab:accuracy}).}
    \label{fig:mixing-eapig-qwen}
\end{figure}

\begin{figure}[htbp]
    \centering
    \includegraphics[width=0.5\textwidth]{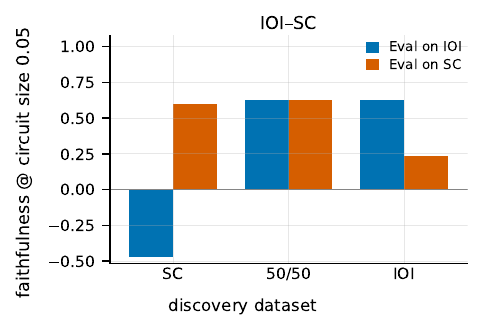}
    \caption{Cross-task mixing with EAP-IG on GPT-2. IOI--SC is the only pair GPT-2 supports: it scores $0.0\%$ on arithmetic and below $55\%$ on entity binding (Table~\ref{tab:accuracy}). The pure sequence-completion circuit is \emph{negative} on IOI, yet the $50/50$ circuit reaches $0.63$ on both tasks.}
    \label{fig:mixing-eapig-gpt2}
\end{figure}

\subsection{EAP}
\label{app:mixing-eap}

Replacing EAP-IG with EAP reproduces the pattern in seven of the nine pairs, at uniformly lower absolute faithfulness---EAP is the cruder approximation of the two, and its pure circuits recover less of the model's behaviour in the first place (for instance $0.25$ in-distribution on Qwen entity binding, against $0.80$ for EAP-IG).

Two pairs, both on Llama-3.1-8B-Instruct, do not show the effect, and it is worth being explicit about why. In EB--IOI the two pure circuits already transfer to each other ($0.49$ and $0.29$), so the pair is not separated under EAP and there is no mixing to observe; the $50/50$ circuit scores $0.31$ on entity binding, below the pure entity-binding circuit itself. In Arith--EB the $50/50$ circuit reaches $0.71$ on arithmetic but only $0.08$ on entity binding: here the mixed circuit does effectively choose one task, which is the opposite of the main-text result. We report both rather than restricting to the pairs that agree. The effect we describe is a possibility that discovery admits and faithfulness does not flag, not an outcome that occurs in every mixture.

\begin{figure}[htbp]
    \centering
    \includegraphics[width=\textwidth]{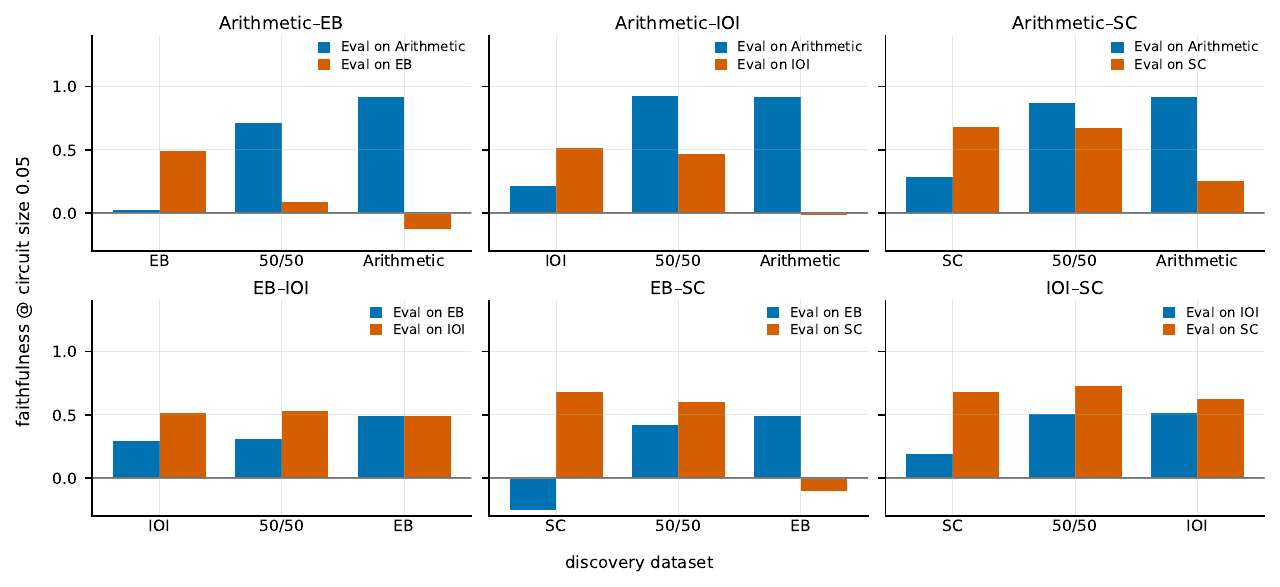}
    \caption{Cross-task mixing with EAP on Llama-3.1-8B-Instruct, the same six pairs as Figure~\ref{fig:exp2}. The pattern reproduces on four pairs. It does not on EB--IOI, where the two pure circuits already transfer to each other, or on Arith--EB, where the $50/50$ circuit is faithful on arithmetic ($0.71$) but not on entity binding ($0.08$).}
    \label{fig:mixing-eap-llama}
\end{figure}

\begin{figure}[htbp]
    \centering
    \includegraphics[width=\textwidth]{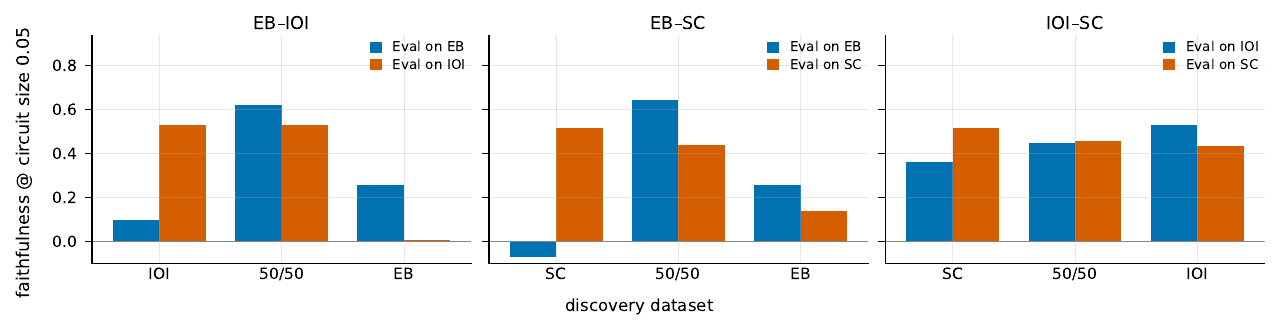}
    \caption{Cross-task mixing with EAP on Qwen2.5-7B-Instruct. Absolute faithfulness is lower than under EAP-IG (Figure~\ref{fig:mixing-eapig-qwen}) throughout, but the ordering is unchanged: the $50/50$ circuit is at least as faithful on each task as that task's own pure circuit is on the other.}
    \label{fig:mixing-eap-qwen}
\end{figure}

\subsection{\textsc{E-Act}}
\label{app:mixing-eact}

\textsc{E-Act} ablates each edge directly rather than approximating the effect with gradients, so it shares no attribution code path with EAP or EAP-IG. It is the only genuinely independent check in this grid, and also the most expensive: it requires a forward pass per edge, which is tractable on GPT-2's $32{,}491$ edges but not on the roughly $10^6$ edges of the larger models. We therefore run it on GPT-2 only, on IOI--SC, and at $200$ examples per circuit instead of $600$.

The result matches the EAP-IG cell on the same pair (Figure~\ref{fig:mixing-eapig-gpt2}). The pure sequence-completion circuit scores $-0.15$ on IOI and the pure IOI circuit scores $0.14$ on sequence completion, while the $50/50$ circuit reaches $0.84$ on IOI and $0.46$ on sequence completion. Because edge overlap and faithfulness both depend on sample size, the absolute values should not be compared across methods; what carries over is the ordering, and it is the same under direct ablation as under both gradient-based methods.

\begin{figure}[htbp]
    \centering
    \includegraphics[width=0.5\textwidth]{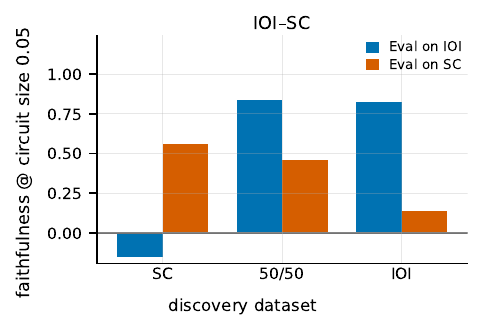}
    \caption{Cross-task mixing with \textsc{E-Act} on GPT-2. Direct edge ablation reproduces the pattern found with both gradient-based methods: each pure circuit is faithful only on its own task, and the $50/50$ circuit is faithful on both.}
    \label{fig:mixing-eact-gpt2}
\end{figure}

\subsection{Faithfulness across the full mixing sweep}
\label{app:mixing-sweep}

The figures above show three points---pure $A$, $50/50$, pure $B$---of a sweep that was run at every mixing proportion from $0.0$ to $1.0$ in steps of $0.1$. This subsection shows the whole sweep, which answers a question the three-point view cannot: whether the effect requires a balanced mixture, or appears as soon as a task is present at all.

These figures use a different vertical axis from the ones above, and the two should not be read against each other. Here each curve is that task's faithfulness at circuit size $0.05$ \emph{divided by the faithfulness of the circuit discovered on that task alone}, so the dashed line at $1.0$ marks parity with a dedicated circuit rather than parity with the full model. A point at $1.0$ means the mixed circuit explains that task as well as a circuit built for it and nothing else. The horizontal axis is the fraction of the first-named task in the discovery dataset, so at $x=0$ the first task is absent and at $x=1$ the second one is.

\begin{figure}[htbp]
    \centering
    \includegraphics[width=\textwidth]{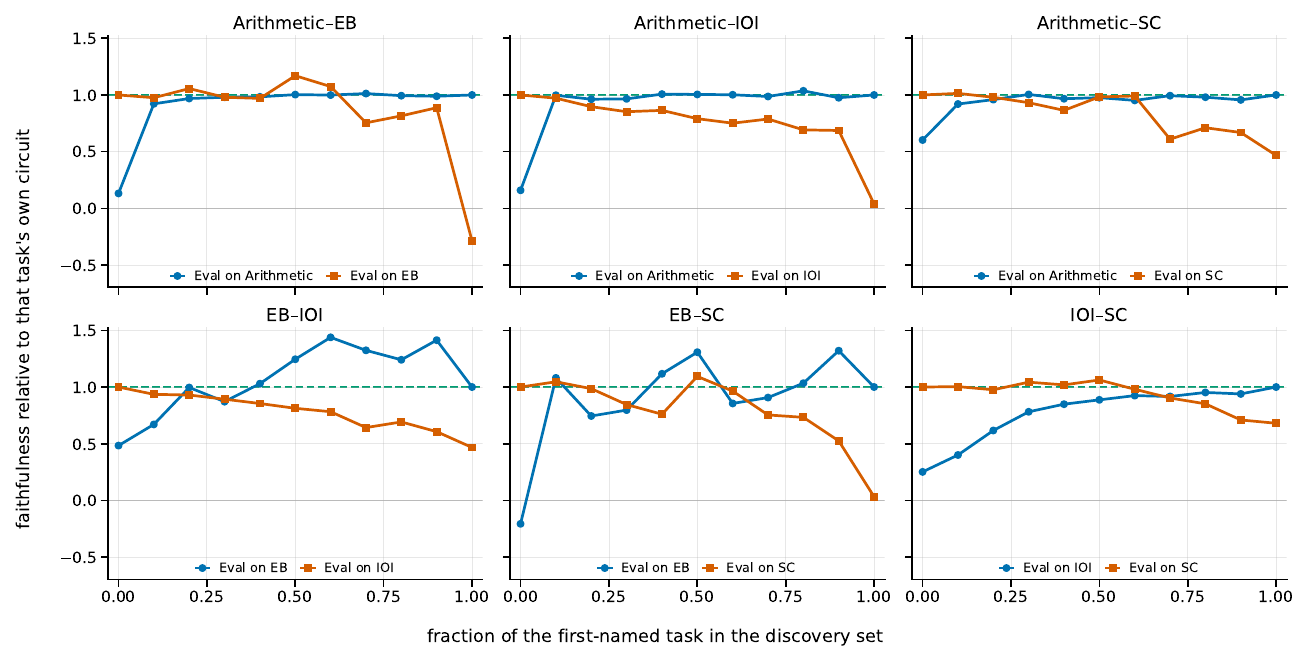}
    \caption{Full mixing sweep with EAP-IG on Llama-3.1-8B-Instruct, one panel per task pair. Each curve is a task's faithfulness at circuit size $0.05$ relative to the circuit discovered on that task alone; the dashed line at $1.0$ is parity with such a dedicated circuit. The $x$-axis is the fraction of the first-named task in the discovery dataset. Both curves are near parity across the interior while each endpoint collapses on the other task.}
    \label{fig:sweep-eapig-llama}
\end{figure}

\begin{figure}[htbp]
    \centering
    \includegraphics[width=\textwidth]{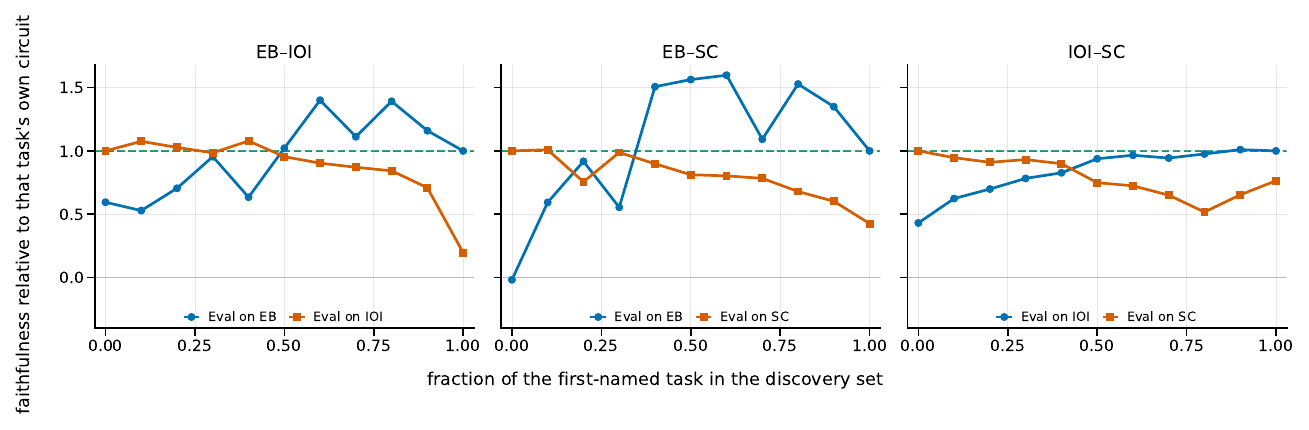}
    \caption{Full mixing sweep with EAP-IG on Qwen2.5-7B-Instruct. Axes as in Figure~\ref{fig:sweep-eapig-llama}.}
    \label{fig:sweep-eapig-qwen}
\end{figure}

\begin{figure}[htbp]
    \centering
    \includegraphics[width=0.5\textwidth]{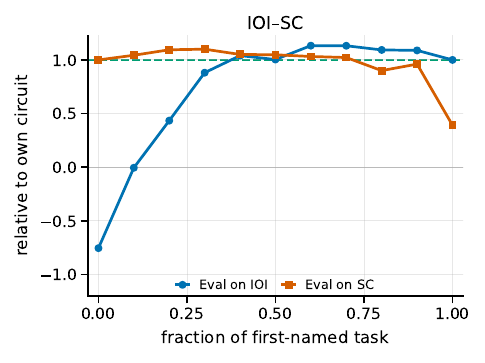}
    \caption{Full mixing sweep with EAP-IG on GPT-2, on the single pair GPT-2 supports. Axes as in Figure~\ref{fig:sweep-eapig-llama}. IOI falls to $-0.75$ of its own circuit under a pure sequence-completion discovery set, recovers to $0.88$ once it makes up $30\%$ of the data and reaches parity at $40\%$.}
    \label{fig:sweep-eapig-gpt2}
\end{figure}

\begin{figure}[htbp]
    \centering
    \includegraphics[width=\textwidth]{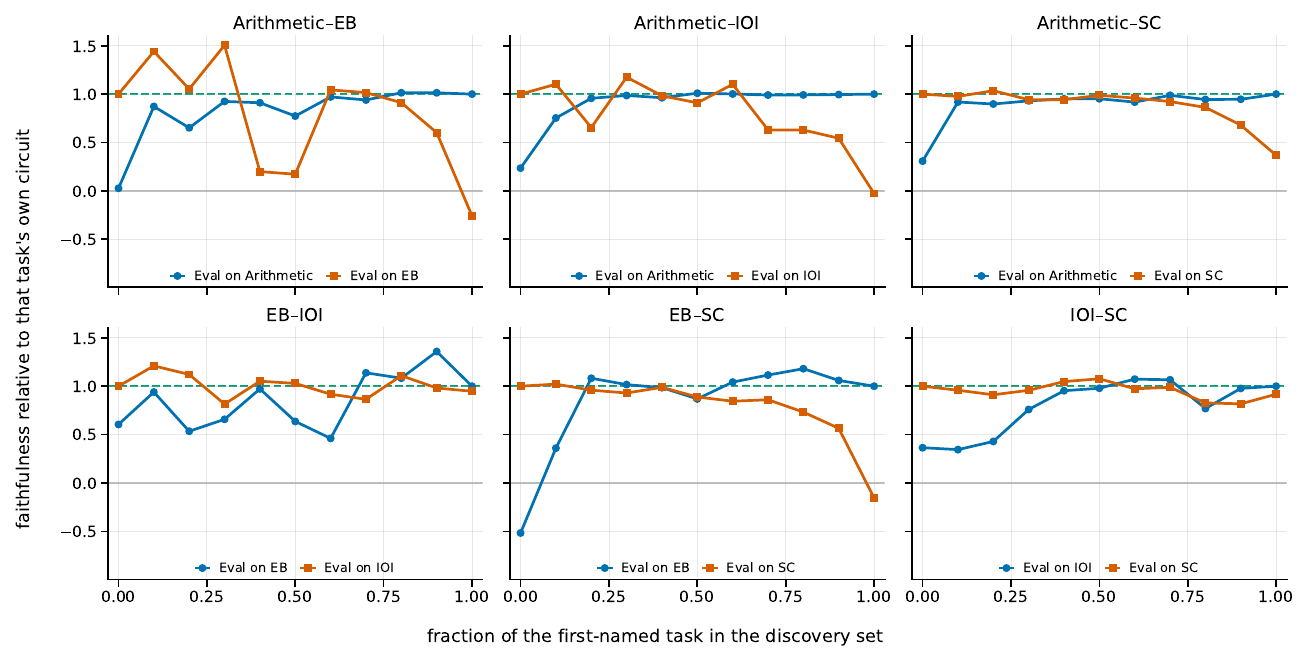}
    \caption{Full mixing sweep with EAP on Llama-3.1-8B-Instruct. Axes as in Figure~\ref{fig:sweep-eapig-llama}. The interior-above-endpoints pattern is clear on the three arithmetic pairs and on EB--SC. It is marginal on IOI--SC and absent on EB--IOI, whose interior clears its endpoints by only $0.02$---the same pair that Figure~\ref{fig:mixing-eap-llama} flagged as unseparated.}
    \label{fig:sweep-eap-llama}
\end{figure}

\begin{figure}[htbp]
    \centering
    \includegraphics[width=\textwidth]{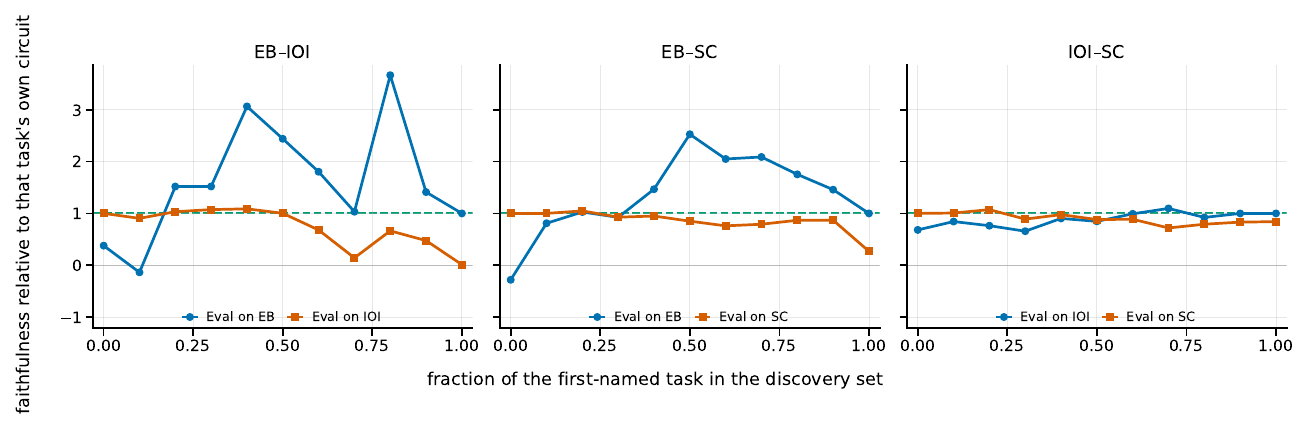}
    \caption{Full mixing sweep with EAP on Qwen2.5-7B-Instruct. Axes as in Figure~\ref{fig:sweep-eapig-llama}. Note the vertical scale: the pure entity-binding circuit scores only $0.25$ in raw terms, so dividing by it inflates the EB--IOI and EB--SC curves. Their shape is meaningful, their height is not.}
    \label{fig:sweep-eap-qwen}
\end{figure}


\subsection{Edge composition of the mixed circuits}
\label{app:mixing-composition}

Section~\ref{subsec: cross-task-mixing} shows that a circuit discovered on a mixed dataset can retain high faithfulness on both constituent tasks. Here, we examine whether this behavioral mixing is also reflected in the circuit's edge composition. For each task pair $(A,B)$ and mixture, we take the top-$5\%$ edges of the mixed circuit and partition them according to their membership in the circuits discovered on $A$ and $B$ separately: edges appearing in both task-specific circuits, in only $A$'s circuit, in only $B$'s circuit, or in neither. This decomposition goes beyond measuring pairwise circuit overlap by asking directly how much of the mixed circuit can be traced to either task-specific circuit.

Table~\ref{tab:mixing-composition} reports this decomposition for the $50/50$ mixture across all method, model, and task-pair combinations. Across all mixing ratios and settings, $88$--$99\%$ of the mixed circuit's edges belong to at least one of the two task-specific circuits, with edges exclusive to each task represented throughout. Across the $172$ (task pair, mixture, setting) combinations, the smaller of the two task-exclusive shares is at least $2.8\%$. Thus, the mixed circuits are composed predominantly of edges found in the constituent task-specific circuits rather than edges unique to the mixed dataset.

\begin{table}[htbp]
\centering
\small
\caption{Composition of the circuit discovered on a $50/50$ mixture, as a
percentage of its top-$5\%$ edges. Columns give membership in the two
task-specific circuits; the four shares are disjoint and sum to $100$.}
\label{tab:mixing-composition}
\vspace{0.5em}
\begin{tabular}{lll cccc}
\toprule
& & & \multicolumn{4}{c}{\textbf{Share of the mixed circuit's edges (\%)}} \\
\cmidrule(lr){4-7}
\textbf{Method} & \textbf{Model} & \textbf{Pair ($A$--$B$)} & both & $A$ only & $B$ only & neither \\
\midrule
EAP-IG & GPT-2 & IOI--SC & 27 & 39 & 32 & 2 \\
\midrule
\multirow{3}{*}{EAP-IG} & \multirow{3}{*}{Qwen-7B-Inst.}
  & EB--IOI    & 37 & 24 & 32 & 6 \\
& & EB--SC     & 24 & 34 & 38 & 5 \\
& & IOI--SC    & 28 & 35 & 32 & 5 \\
\midrule
\multirow{6}{*}{EAP-IG} & \multirow{6}{*}{Llama-8B-Inst.}
  & Arith--EB  & 28 & 15 & 55 & 2 \\
& & Arith--IOI & 27 & 23 & 48 & 2 \\
& & Arith--SC  & 30 & 25 & 34 & 11 \\
& & EB--IOI    & 41 & 37 & 21 & 2 \\
& & EB--SC     & 32 & 46 & 18 & 3 \\
& & IOI--SC    & 34 & 36 & 24 & 6 \\
\midrule
\multirow{3}{*}{EAP} & \multirow{3}{*}{Qwen-7B-Inst.}
  & EB--IOI    & 38 & 25 & 26 & 11 \\
& & EB--SC     & 25 & 19 & 49 & 7 \\
& & IOI--SC    & 30 & 19 & 43 & 8 \\
\midrule
\multirow{6}{*}{EAP} & \multirow{6}{*}{Llama-8B-Inst.}
  & Arith--EB  & 27 & 24 & 46 & 3 \\
& & Arith--IOI & 26 & 38 & 30 & 5 \\
& & Arith--SC  & 30 & 21 & 41 & 9 \\
& & EB--IOI    & 44 & 37 & 14 & 5 \\
& & EB--SC     & 32 & 32 & 32 & 5 \\
& & IOI--SC    & 34 & 17 & 39 & 9 \\
\midrule
\textsc{E-Act} & GPT-2 & IOI--SC & 26 & 31 & 38 & 4 \\
\bottomrule
\end{tabular}
\end{table}


\section{Additional results: cross-variant transfer within a task}
\label{app:exp1}

This appendix presents the full set of cross-variant faithfulness and edge-overlap results supporting Section~\ref{subsec: single-task-mixing} of the main paper. Section~\ref{app:hd-setup} states the conventions used throughout. Section~\ref{app:variant-pairs} explains how the 136 variant pairs of Table~\ref{tab:transfer} are formed, Section~\ref{app:faithfulness-floor} how we decide that a transfer fails, and Section~\ref{app:twoway} lists the seven pairs that fail in both directions. Section~\ref{app:hd-overlap-summary} summarises edge overlap across all (task, model, axis) cells. Section~\ref{app:crossfaith-howto} prints every cross-variant faithfulness value we measured, one table per (task, model, attribution method); every number quoted in Section~\ref{subsec: single-task-mixing} can be checked against these tables. Section~\ref{app:exp1-curves} shows how in-distribution and cross-variant faithfulness evolve with circuit size.

\subsection{Setup}
\label{app:hd-setup}

Unless otherwise noted, we follow these conventions throughout this appendix:

\begin{itemize}
    \item \textbf{Hypothesis-driven circuit discovery methods.} EAP-IG~\citep{hanna2024have} is the default circuit-discovery method; EAP~\citep{syed2024attribution} and E-Act~\citep{conmy2023acdc} results are reported as method-robustness checks. E-Act requires a forward pass per edge and is run only on GPT-2 Small, on IOI.
    \item \textbf{Circuit size and metric.} Faithfulness values and edge overlaps are reported at a circuit size of 5\% of edges, using the normalised faithfulness score of Section~\ref{subsec:faithfulness}, unless explicitly mentioned. The score is not bounded to $[0,1]$ (see Section~\ref{app:crossfaith-howto}).
    \item \textbf{In-distribution variant.} For the faithfulness curves in Section~\ref{app:exp1-curves}, the simplest or canonical variant of each axis is treated as in-distribution: 2-person for IOI complexity and domain, ABBA for IOI syntax, 2-comma for entity binding complexity, delimiter and domain, P1 for entity binding position, 2-operand for arithmetic complexity and domain, verbal v1 for arithmetic syntax, and 2-gram for sequence completion. The cross-variant tables in Section~\ref{app:crossfaith-howto} report all ordered pairs, so no variant is privileged there.
    \item \textbf{Coverage and exclusions.} A (task, model) cell is evaluated only when the model meets the accuracy criterion of Section~\ref{sec:experiment_setup} on the task (Table~\ref{tab:accuracy}). This excludes entity binding and arithmetic on GPT-2 Small and arithmetic on Qwen2.5-7B-Instruct; arithmetic is therefore evaluated only on Llama-3.1-8B-Instruct, whose tokenizer represents operands and answers as single tokens (Section~\ref{app:dataset}). Sequence completion is varied only along the complexity axis.
\end{itemize}

\subsection{Construction of the variant pairs}
\label{app:variant-pairs}

Section~\ref{subsec: single-task-mixing} compares circuits across pairs of variants of the same task. A pair is formed only between two variants that differ along a single axis, so that a transfer failure can be attributed to that one change. Within an axis we take every unordered pair of its variants: an axis with $m$ variants contributes $\binom{m}{2}$ pairs. The base variant of each task (2-person IOI, i.e.\ ABBA and BABA name orders pooled; 2-comma entity binding; 2-operand arithmetic; 2-gram sequence completion) anchors every axis of that task, so it appears in several pairs. The IOI syntax axis pools its two surface changes (name order and voice): we take all three unordered pairs among ABBA, BABA and passive. The entity-binding syntax axis is split into two sub-axes, queried position (P1--P8) and delimiter (2-comma vs.\ 2-period), and pairs are formed within each sub-axis. Table~\ref{tab:variant-pairs} lists the axes, their variants and the resulting pair counts.

A pair is evaluated on a model only when the model meets the accuracy criterion of Section~\ref{app:hd-setup} on the task. GPT-2 therefore contributes only IOI and sequence completion, and arithmetic is evaluated only on Llama-3.1-8B-Instruct. This gives 137 pairs. One further pair is untestable: the entity-binding domain pair (2-comma against 2-colour-box) on Qwen2.5-7B-Instruct has a bootstrap noise floor of roughly 122 percentage points, because the colour-box circuit is unstable on its own test set (in-distribution faithfulness $0.57$, cross-variant $2.49$; Table~\ref{tab:crossfaith-entity-binding-qwen2-5-7b-instruct-eap-ig-inputs}), so no shortfall could exceed the floor. We exclude it, leaving the 136 pairs of Table~\ref{tab:transfer}. Every remaining floor is below 22 percentage points (Table~\ref{tab:floor-summary}).

\begin{table}[t]
\centering\small
\caption{Variant pairs by task, axis and model. Each axis contributes all unordered pairs of its variants; a dash marks a task the model does not meet the accuracy criterion on. The Qwen2.5-7B-Instruct entity-binding domain pair is excluded as untestable (see text).}
\label{tab:variant-pairs}
\begin{tabular}{llcccc}
\toprule
Task & Axis (variants) & Pairs & GPT-2 & Llama-3.1 & Qwen2.5 \\
\midrule
\multirow{3}{*}{IOI}
  & Complexity (2-person, 3-person, filler) & 3 & 3 & 3 & 3 \\
  & Syntax (ABBA, BABA, passive)            & 3 & 3 & 3 & 3 \\
  & Domain (2-person, letter)               & 1 & 1 & 1 & 1 \\
\midrule
\multirow{4}{*}{Entity binding}
  & Complexity (2- to 8-comma)              & 21 & -- & 21 & 21 \\
  & Position (P1--P8)                       & 28 & -- & 28 & 28 \\
  & Delimiter (2-comma, 2-period)           & 1  & -- & 1  & 1 \\
  & Domain (2-comma, 2-colour-box)          & 1  & -- & 1  & 0 \\
\midrule
\multirow{3}{*}{Arithmetic}
  & Complexity (2-, 3-operand)              & 1 & -- & 1 & -- \\
  & Domain (2-operand, verbal v1)           & 1 & -- & 1 & -- \\
  & Syntax (verbal v1, v2, v3)              & 3 & -- & 3 & -- \\
\midrule
Sequence completion
  & Complexity (2-, 3-, 4-gram)             & 3 & 3 & 3 & 3 \\
\midrule
\multicolumn{2}{l}{Total} & & 10 & 66 & 60 \\
\bottomrule
\end{tabular}
\end{table}

\subsection{Estimating variation in faithfulness}
\label{app:faithfulness-floor}

Our cross-variant analysis asks whether a circuit $C^i$ discovered on variant $D^i_\tau$
retains its faithfulness when evaluated on another variant $D^j_\tau$. We compare its
cross-variant faithfulness, $f(C^i,D^j_\tau)$, against the in-distribution faithfulness of
the circuit discovered directly on the target variant, $f(C^j,D^j_\tau)$. However, even
when the underlying circuit and data distribution are unchanged, faithfulness estimated on
a finite test set can vary. We therefore estimate this variation for each variant and use
it to determine whether a cross-variant shortfall is larger than expected from estimation
variability alone.

\paragraph{Estimating the noise floor}
For each variant $D^j_\tau$, we independently bootstrap-resample its in-distribution test
set twice and compute the faithfulness of $C^j$ on each resample. We draw $2{,}000$ such
pairs of resamples and define the variant's \emph{noise floor}, $\epsilon_j$, as the $95$th
percentile of the absolute difference between the two resulting faithfulness estimates.
Thus $\epsilon_j$ measures how far apart two estimates of the same circuit's
in-distribution faithfulness can lie because of resampling alone. Test sets hold $200$
prompts per variant, and floors are estimated separately for each variant at each circuit
size; the main text uses size $0.05$ throughout. In this subsection and in
Table~\ref{tab:s42-twoway}, floors and shortfalls are given in percentage points of
faithfulness, i.e.\ $100\times$ the normalised score used in the main text and in the
tables of Section~\ref{app:crossfaith-howto}.

\paragraph{Magnitude of the floor}
Table~\ref{tab:floor-summary} reports the floors for the variants entering the
analysis. Across them the median floor is $5.6$ points. It is smallest where the task is
easy to fit, at $1.3$ to $3.1$ points on arithmetic with Llama-3.1-8B-Instruct, and largest on
entity binding, reaching $21.9$ points on Qwen2.5-7B-Instruct. Floors vary more between tasks than
between models: all three sequence-completion cells fall between $2.8$ and $7.5$ points,
while the two entity-binding cells start above where most IOI cells end.

One variant has no usable estimate. The $2$-colour-box variant on Qwen2.5-7B-Instruct has an
empty-circuit score almost equal to that of the full model, so the denominator normalising
faithfulness (Section~\ref{subsec:faithfulness}) nearly vanishes and the floor is unstable
at roughly $122$ points. The only pair involving this variant, the entity-binding domain
pair on Qwen2.5-7B-Instruct, is therefore excluded from the analysis
(Section~\ref{app:variant-pairs}).

\begin{table}[t]
\centering\small
\caption{Noise floors at circuit size $0.05$, in percentage points of faithfulness, for the
variants entering the cross-variant analysis. Each floor is the $95$th percentile of
the absolute difference between two independent bootstrap resamples of that variant's
$200$-prompt test set, over $2{,}000$ draws.}
\label{tab:floor-summary}
\begin{tabular}{llrcc}
\toprule
Task & Model & Variants & Range & Median \\
\midrule
IOI                 & GPT-2                 &  7 & $3.1$--$9.6$  & $5.9$ \\
IOI                 & Llama-3.1-8B-Instruct &  6 & $3.0$--$4.7$  & $3.7$ \\
IOI                 & Qwen2.5-7B-Instruct   &  6 & $3.4$--$7.7$  & $6.1$ \\
Entity binding      & Llama-3.1-8B-Instruct &  3 & $9.3$--$12.9$ & $12.1$ \\
Entity binding      & Qwen2.5-7B-Instruct   & 13 & $5.5$--$21.9$ & $7.6$ \\
Arithmetic          & Llama-3.1-8B-Instruct &  5 & $1.3$--$3.1$  & $1.7$ \\
Sequence completion & GPT-2                 &  3 & $3.1$--$5.5$  & $3.7$ \\
Sequence completion & Llama-3.1-8B-Instruct &  3 & $2.8$--$4.7$  & $2.8$ \\
Sequence completion & Qwen2.5-7B-Instruct   &  3 & $6.0$--$7.5$  & $7.4$ \\
\midrule
\multicolumn{2}{l}{All} & 49 & $1.3$--$21.9$ & $5.6$ \\
\bottomrule
\end{tabular}
\end{table}

\paragraph{Deciding when a transfer fails}
For transfer from $D^i_\tau$ to $D^j_\tau$, we measure the faithfulness shortfall relative
to the circuit discovered on the target variant:
\[
    \Delta_{i\rightarrow j} = f(C^j,D^j_\tau)-f(C^i,D^j_\tau).
\]
Both terms are evaluated on the same target test set $D^j_\tau$, so the comparison uses the
same examples and the same faithfulness normalisation; a variant whose empty-circuit score
sits far from the full model's cannot by itself produce an apparent shortfall. We consider
the transfer $i\rightarrow j$ to fail only when this shortfall exceeds the larger of the two
variants' floors,
\[
    \Delta_{i\rightarrow j} > \max(\epsilon_i,\epsilon_j).
\]

\paragraph{Three transfer outcomes}
This criterion is directional. Consequently, a pair $(D^i_\tau,D^j_\tau)$ can exhibit
transfer in both directions, in only one direction, or in neither direction.
Section~\ref{subsec: single-task-mixing} uses these three outcomes to evaluate evidence for
mechanistic diversity, reserving the distinct-mechanism reading for pairs where neither
direction transfers.

\subsection{Variant pairs that fail to transfer in both directions}
\label{app:twoway}

Section~\ref{subsec: single-task-mixing} treats failure in both transfer directions as evidence that two variants may rely on distinct mechanisms. Table~\ref{tab:s42-twoway} reports all seven variant pairs that satisfy this criterion.

For each pair, we report both transfer directions. For a transfer $i \rightarrow j$, $f_{jj} = f(C^j, D^j)$ is the in-distribution faithfulness of the circuit discovered on the target variant $j$, and $f_{ij} = f(C^i, D^j)$ is the faithfulness of the circuit discovered on variant $i$ when transferred to the same test set. Their difference, $\Delta = f_{jj} - f_{ij}$, is the faithfulness shortfall of Section~\ref{app:faithfulness-floor}. A transfer fails when $\Delta$ exceeds the pair's noise floor, defined as the larger of the two variants' bootstrap estimates; a pair is included here only when this occurs in both directions. Faithfulness is reported in percent at circuit size $0.05$, while $\Delta$ and the noise floor are in percentage points. We additionally report the Jaccard similarity $J$ between the two circuits' top-5\% edge sets.

The seven pairs are concentrated in IOI and entity binding. Four occur in IOI: the name-to-letter domain change fails in both directions on all three models, and the fourth pair is a complexity change between 3-person and filler prompts on Llama-3.1-8B-Instruct. The other three pairs are between queried positions in entity binding, all involving the last position P8: P3 and P4 against P8 on Qwen2.5-7B-Instruct, and P1 against P8 on Llama-3.1-8B-Instruct. Arithmetic and sequence completion contain no pairs that fail to transfer in both directions. Despite these transfer failures, the circuits retain substantial structural overlap, with $J$ ranging from $0.30$ to $0.48$. For reference, circuits discovered from disjoint halves of the same variant have $J = 0.580$, whereas two random 5\% edge sets have $J = 0.026$.

\begin{table}[t]
\centering\small
\caption{Variant pairs that fail to transfer in both directions. For each transfer $i \rightarrow j$, $f_{jj}$ is the target variant's in-distribution faithfulness, $f_{ij}$ is the transferred circuit's faithfulness on the same test set, and $\Delta = f_{jj} - f_{ij}$ is the shortfall; all three are in percent. Both directions exceed the pair's bootstrap noise floor (percentage points). $J$ is the Jaccard similarity between the two circuits' top-5\% edges.}
\label{tab:s42-twoway}
\begin{tabular}{lrrrrc}
\toprule
Transfer & $f_{jj}$ & $f_{ij}$ & $\Delta$ & Floor & $J$ \\
\midrule
\multicolumn{6}{l}{\emph{Entity binding, Qwen2.5-7B-Instruct}} \\
\quad P3 $\rightarrow$ P8 & 93  & 39 & 53.8 & \multirow{2}{*}{5.8}  & \multirow{2}{*}{0.46} \\
\quad P8 $\rightarrow$ P3 & 108 & 96 & 12.1 &                       & \\
\quad P4 $\rightarrow$ P8 & 93  & 63 & 30.1 & \multirow{2}{*}{5.3}  & \multirow{2}{*}{0.48} \\
\quad P8 $\rightarrow$ P4 & 98  & 92 & 5.9  &                       & \\
\midrule
\multicolumn{6}{l}{\emph{Entity binding, Llama-3.1-8B-Instruct}} \\
\quad P1 $\rightarrow$ P8 & 91 & 52 & 39.0 & \multirow{2}{*}{12.2} & \multirow{2}{*}{0.45} \\
\quad P8 $\rightarrow$ P1 & 97 & 79 & 18.1 &                       & \\
\midrule
\multicolumn{6}{l}{\emph{IOI, GPT-2}} \\
\quad 2-person $\rightarrow$ letter & 83 & 56 & 27.0 & \multirow{2}{*}{4.0} & \multirow{2}{*}{0.30} \\
\quad letter $\rightarrow$ 2-person & 73 & 60 & 12.7 &                      & \\
\midrule
\multicolumn{6}{l}{\emph{IOI, Llama-3.1-8B-Instruct}} \\
\quad 3-person $\rightarrow$ filler & 98 & 91 & 6.4  & \multirow{2}{*}{4.7} & \multirow{2}{*}{0.47} \\
\quad filler $\rightarrow$ 3-person & 98 & 73 & 24.2 &                      & \\
\quad 2-person $\rightarrow$ letter & 92 & 78 & 14.2 & \multirow{2}{*}{3.2} & \multirow{2}{*}{0.38} \\
\quad letter $\rightarrow$ 2-person & 90 & 79 & 10.1 &                      & \\
\midrule
\multicolumn{6}{l}{\emph{IOI, Qwen2.5-7B-Instruct}} \\
\quad 2-person $\rightarrow$ letter & 84 & 76 & 7.8  & \multirow{2}{*}{5.8} & \multirow{2}{*}{0.39} \\
\quad letter $\rightarrow$ 2-person & 93 & 66 & 26.9 &                      & \\
\bottomrule
\end{tabular}
\end{table}

\subsection{Edge overlap across tasks, models, and axes}
\label{app:hd-overlap-summary}

\begin{table}[t!]
\centering
\small
\caption{Pairwise top-5\% Jaccard edge overlap between circuits discovered on different variants within each (task, model, axis) cell. Reported as mean (min, max) over all unordered variant pairs within the axis. For cells with exactly two variants, mean equals min and max. Dashes mark (task, model, axis) combinations that were not evaluated (Section~\ref{app:hd-setup}).}
\label{tab:hd-jaccard-summary}
\vspace{0.5em}
\begin{tabular}{llccc}
\toprule
\textbf{Task} & \textbf{Model} & \textbf{Complexity} & \textbf{Syntax} & \textbf{Domain} \\
\midrule
\multirow{3}{*}{IOI}
  & GPT-2          & 0.54 (0.52, 0.57) & 0.57 (0.47, 0.79) & 0.30 (0.30, 0.30) \\
  & Qwen-7B-Inst.  & 0.42 (0.40, 0.43) & 0.53 (0.46, 0.81) & 0.39 (0.39, 0.39) \\
  & Llama-8B-Inst. & 0.50 (0.47, 0.51) & 0.60 (0.53, 0.86) & 0.38 (0.38, 0.38) \\
\midrule
\multirow{2}{*}{Entity binding}
  & Qwen-7B-Inst.  & 0.61 (0.42, 0.83) & 0.55 (0.41, 0.76) & 0.46 (0.46, 0.46) \\
  & Llama-8B-Inst. & 0.66 (0.49, 0.84) & 0.60 (0.45, 0.81) & 0.52 (0.52, 0.52) \\
\midrule
Arithmetic
  & Llama-8B-Inst. & 0.36 (0.36, 0.36) & 0.53 (0.51, 0.56) & 0.35 (0.35, 0.35) \\
\midrule
\multirow{3}{*}{Sequence completion}
  & GPT-2          & 0.61 (0.54, 0.74) & --- & --- \\
  & Qwen-7B-Inst.  & 0.53 (0.49, 0.59) & --- & --- \\
  & Llama-8B-Inst. & 0.50 (0.40, 0.68) & --- & --- \\
\bottomrule
\end{tabular}
\end{table}

Table~\ref{tab:hd-jaccard-summary} reports the pairwise top-5\% Jaccard overlap between circuits discovered on different variants within each (task, model, axis) cell, summarised as mean (min, max) over all unordered variant pairs within the axis. Across every populated cell, the mean overlap falls within $[0.30, 0.66]$, indicating that circuits discovered on different variants of the same task share at most a moderate fraction of their attribution-ranked edges. The within-cell spread is modest in most cases---across the 14 cells with three or more variants, the median gap between the cell mean and its minimum is $0.07$, so the mean is a reasonable summary. The exception is the entity-binding complexity and syntax axes, where the seven or eight variants produce gaps as large as $0.19$ between the cell mean and its most distant variant pair.

\subsection{Cross-variant faithfulness, per pair}
\label{app:crossfaith-howto}

Every cross-variant faithfulness value we measured is printed in
Tables~\ref{tab:crossfaith-ioi-gpt2-eap-ig-inputs}--\ref{tab:crossfaith-sequence-completion-llama-3-1-8b-instruct-eap},
one table per (task, model, attribution method). The main text quotes individual
entries from these tables, and the summary in
Table~\ref{tab:transfer} is computed from them, so every number in
Section~\ref{subsec: single-task-mixing} can be checked against the grids below.

\paragraph{How to read these tables.} Entry $(i, j)$ is $f(C^i, D^j)$: the
faithfulness at $5\%$ circuit size of the circuit discovered on variant $i$
when evaluated on variant $j$'s test data. Rows are the discovery variant,
columns the evaluation variant, and the bold diagonal is in-distribution
faithfulness. The two quantities the analysis uses both follow from the grid:
the loss a circuit takes outside its own dataset is its row's diagonal minus
the off-diagonal entry, $f(C^i, D^i) - f(C^i, D^j)$, while the penalty for
using a foreign circuit on a given dataset is that column's diagonal minus the
entry, $f(C^j, D^j) - f(C^i, D^j)$. Variant names follow Appendix~\ref{app:dataset}; ``2-person'' is the IOI base variant with ABBA and BABA name orders pooled. Where a cell was stored as separate per-axis matrices, the table shows one block per axis and a pair appearing in more than one block is taken from the widest; the in-distribution value of a base variant can therefore differ slightly between blocks. A dash marks a pair that was not measured.

\paragraph{Faithfulness is not bounded to $[0,1]$.} It is the normalised score of $f(C,G;m)=\frac{m(C)-m(\varnothing)}{m(G)-m(\varnothing)}$, whose denominator $m(G) - m(\varnothing)$ can
be small, so $125$ of the $817$ entries below fall outside the unit interval
(range $-0.85$ to $2.49$). A value above $1$ means the circuit scores higher
than the full model on that data; a negative value means it scores below the
fully ablated model. We print these as measured rather than clipping them,
since a clipped table would misrepresent roughly an eighth of the data.
Cross-variant drops should be read against the per-variant task accuracies in
Table~\ref{tab:accuracy}: where a model is much less accurate on a variant, a
drop may reflect difficulty rather than a change of mechanism.

\paragraph{Coverage.} Section~\ref{subsec: single-task-mixing} did not evaluate every
combination. Entity binding and arithmetic have no GPT-2 results, and
arithmetic has no Qwen2.5-7B-Instruct results, because these cells fail the
accuracy criterion; sequence completion was varied only along the complexity
axis; and \textsc{E-Act} was run only on GPT-2 for IOI, being too expensive
otherwise. Most cells were stored as separate per-axis matrices rather than one
all-variant matrix, so pairs spanning two different axes are unavailable for
those cells: entity binding, for instance, has $151$ of the $289$ ordered pairs
its $17$ variants could form. Each caption states what its cell covers. A
machine-readable copy of every entry at all ten circuit sizes accompanies the
code release.

\paragraph{\textsc{EAP} circuits are weak in some cells.} The tables make a
limitation visible that the faithfulness curves do not. At $5\%$ of edges,
\textsc{EAP-IG} recovers the model's behaviour in every cell we measure --- all
$69$ of its in-distribution values exceed $0.3$, with a median of $0.90$ --- and
\textsc{E-Act} likewise (median $0.94$). \textsc{EAP} does not: $17$ of its $59$
in-distribution values fall below $0.3$ and four are at or below zero, meaning
the circuit scores no better than the fully ablated model on the very data it
was discovered from. Two cells are affected throughout: on IOI with GPT-2 six of
seven in-distribution values are below $0.3$ (minimum $-0.85$), and on entity
binding with Qwen2.5-7B-Instruct all seven are (median $0.13$). Cross-variant
differences computed from those cells are differences between circuits that do
not explain their own data at this budget, and should not be read as evidence
about mechanism; we report them for completeness rather than as support for any
claim.

\begin{table}[htbp]
\centering
\footnotesize
\setlength{\tabcolsep}{4pt}
\caption{Indirect object identification on GPT-2, circuits discovered with \textsc{EAP-IG}.}
\label{tab:crossfaith-ioi-gpt2-eap-ig-inputs}
\vspace{0.4em}

\begin{tabular}{lccccccc}
\toprule
\textbf{discovered on} & \multicolumn{7}{c}{\textbf{evaluated on}} \\
\cmidrule(l){2-8}
 & ABBA & BABA & 2-person & filler & letter & passive & 3-person \\
\midrule
ABBA & \textbf{0.59} & 0.52 & 0.53 & 0.58 & 0.43 & 0.49 & 0.35 \\
BABA & 0.55 & \textbf{0.85} & 0.85 & 0.89 & 0.77 & 0.85 & 0.68 \\
2-person & 0.40 & 0.76 & \textbf{0.73} & 0.61 & 0.56 & 0.55 & 0.42 \\
filler & 0.36 & 0.56 & 0.55 & \textbf{0.56} & 0.53 & 0.51 & 0.37 \\
letter & 0.46 & 0.62 & 0.60 & 0.56 & \textbf{0.83} & 0.59 & 0.67 \\
passive & 0.35 & 0.72 & 0.71 & 0.37 & 0.58 & \textbf{0.52} & 0.36 \\
3-person & 0.61 & 0.68 & 0.68 & 0.72 & 0.61 & 0.68 & \textbf{0.56} \\
\bottomrule
\end{tabular}

\end{table}

\begin{table}[htbp]
\centering
\footnotesize
\setlength{\tabcolsep}{4pt}
\caption{Indirect object identification on GPT-2, circuits discovered with \textsc{EAP}.}
\label{tab:crossfaith-ioi-gpt2-eap}
\vspace{0.4em}

\textbf{syntax axis}\par\vspace{2pt}
\begin{tabular}{lcccc}
\toprule
\textbf{discovered on} & \multicolumn{4}{c}{\textbf{evaluated on}} \\
\cmidrule(l){2-5}
 & ABBA & BABA & 2-person & passive \\
\midrule
ABBA & \textbf{-0.12} & 0.20 & 0.20 & -0.08 \\
BABA & -0.25 & \textbf{0.12} & 0.10 & -0.22 \\
2-person & -0.31 & 0.07 & \textbf{0.07} & -0.28 \\
passive & -0.47 & -0.02 & -0.02 & \textbf{-0.52} \\
\bottomrule
\end{tabular}

\vspace{10pt}

\textbf{complexity axis}\par\vspace{2pt}
\begin{tabular}{lccc}
\toprule
\textbf{discovered on} & \multicolumn{3}{c}{\textbf{evaluated on}} \\
\cmidrule(l){2-4}
 & 2-person & filler & 3-person \\
\midrule
2-person & \textbf{0.07} & -0.41 & -0.62 \\
filler & 0.33 & \textbf{0.18} & -0.18 \\
3-person & -0.03 & -0.37 & \textbf{-0.85} \\
\bottomrule
\end{tabular}

\vspace{10pt}

\textbf{domain axis}\par\vspace{2pt}
\begin{tabular}{lcc}
\toprule
\textbf{discovered on} & \multicolumn{2}{c}{\textbf{evaluated on}} \\
\cmidrule(l){2-3}
 & 2-person & letter \\
\midrule
2-person & \textbf{0.07} & 0.29 \\
letter & 0.42 & \textbf{0.40} \\
\bottomrule
\end{tabular}

\end{table}

\begin{table}[htbp]
\centering
\footnotesize
\setlength{\tabcolsep}{4pt}
\caption{Indirect object identification on GPT-2, circuits discovered with \textsc{E-Act}.}
\label{tab:crossfaith-ioi-gpt2-eact}
\vspace{0.4em}

\textbf{syntax axis}\par\vspace{2pt}
\begin{tabular}{lcccc}
\toprule
\textbf{discovered on} & \multicolumn{4}{c}{\textbf{evaluated on}} \\
\cmidrule(l){2-5}
 & ABBA & BABA & 2-person & passive \\
\midrule
ABBA & \textbf{0.87} & 0.54 & 0.53 & 0.60 \\
BABA & 0.93 & \textbf{0.94} & 0.93 & 0.97 \\
2-person & 0.57 & 0.95 & \textbf{0.94} & 0.83 \\
passive & 0.61 & 0.98 & 0.98 & \textbf{1.02} \\
\bottomrule
\end{tabular}

\vspace{10pt}

\textbf{complexity axis}\par\vspace{2pt}
\begin{tabular}{lcc}
\toprule
\textbf{discovered on} & \multicolumn{2}{c}{\textbf{evaluated on}} \\
\cmidrule(l){2-3}
 & 2-person & 3-person \\
\midrule
2-person & \textbf{0.94} & 0.77 \\
3-person & 0.95 & \textbf{1.06} \\
\bottomrule
\end{tabular}

\vspace{10pt}

\textbf{domain axis}\par\vspace{2pt}
\begin{tabular}{lcc}
\toprule
\textbf{discovered on} & \multicolumn{2}{c}{\textbf{evaluated on}} \\
\cmidrule(l){2-3}
 & 2-person & letter \\
\midrule
2-person & \textbf{0.94} & 0.62 \\
letter & 0.65 & \textbf{0.90} \\
\bottomrule
\end{tabular}

\vspace{10pt}

\textbf{complexity axis (2-person vs.\ filler)}\par\vspace{2pt}
\begin{tabular}{lcc}
\toprule
\textbf{discovered on} & \multicolumn{2}{c}{\textbf{evaluated on}} \\
\cmidrule(l){2-3}
 & 2-person & filler \\
\midrule
2-person & \textbf{0.94} & 0.95 \\
filler & 0.94 & \textbf{0.92} \\
\bottomrule
\end{tabular}

\end{table}

\begin{table}[htbp]
\centering
\footnotesize
\setlength{\tabcolsep}{4pt}
\caption{Indirect object identification on Qwen2.5-7B-Instruct, circuits discovered with \textsc{EAP-IG}.}
\label{tab:crossfaith-ioi-qwen2-5-7b-instruct-eap-ig-inputs}
\vspace{0.4em}

\begin{tabular}{lccccccc}
\toprule
\textbf{discovered on} & \multicolumn{7}{c}{\textbf{evaluated on}} \\
\cmidrule(l){2-8}
 & ABBA & BABA & 2-person & filler & letter & passive & 3-person \\
\midrule
ABBA & \textbf{0.96} & 0.86 & 0.87 & 0.83 & 0.80 & 0.93 & 0.67 \\
BABA & 0.95 & \textbf{0.90} & 0.87 & 0.78 & 0.69 & 0.91 & 0.59 \\
2-person & 1.09 & 0.93 & \textbf{0.93} & 0.85 & 0.76 & 1.02 & 0.63 \\
filler & 0.99 & 0.85 & 0.88 & \textbf{0.84} & 0.72 & 0.90 & 0.56 \\
letter & 0.79 & 0.66 & 0.66 & 0.61 & \textbf{0.84} & 0.68 & 0.58 \\
passive & 1.10 & 1.02 & 0.99 & 0.86 & 0.80 & \textbf{1.06} & 0.63 \\
3-person & 1.08 & 0.97 & 0.95 & 0.87 & 0.80 & 1.03 & \textbf{0.77} \\
\bottomrule
\end{tabular}

\end{table}

\begin{table}[htbp]
\centering
\footnotesize
\setlength{\tabcolsep}{4pt}
\caption{Indirect object identification on Qwen2.5-7B-Instruct, circuits discovered with \textsc{EAP}.}
\label{tab:crossfaith-ioi-qwen2-5-7b-instruct-eap}
\vspace{0.4em}

\textbf{syntax axis}\par\vspace{2pt}
\begin{tabular}{lcccc}
\toprule
\textbf{discovered on} & \multicolumn{4}{c}{\textbf{evaluated on}} \\
\cmidrule(l){2-5}
 & ABBA & BABA & 2-person & passive \\
\midrule
ABBA & \textbf{0.51} & 0.48 & 0.45 & 0.49 \\
BABA & 0.55 & \textbf{0.49} & 0.51 & 0.55 \\
2-person & 0.42 & 0.44 & \textbf{0.46} & 0.42 \\
passive & 0.58 & 0.48 & 0.47 & \textbf{0.59} \\
\bottomrule
\end{tabular}

\vspace{10pt}

\textbf{complexity axis}\par\vspace{2pt}
\begin{tabular}{lccc}
\toprule
\textbf{discovered on} & \multicolumn{3}{c}{\textbf{evaluated on}} \\
\cmidrule(l){2-4}
 & 2-person & filler & 3-person \\
\midrule
2-person & \textbf{0.46} & 0.33 & 0.26 \\
filler & 0.38 & \textbf{0.37} & 0.21 \\
3-person & 0.26 & 0.21 & \textbf{0.17} \\
\bottomrule
\end{tabular}

\vspace{10pt}

\textbf{domain axis}\par\vspace{2pt}
\begin{tabular}{lcc}
\toprule
\textbf{discovered on} & \multicolumn{2}{c}{\textbf{evaluated on}} \\
\cmidrule(l){2-3}
 & 2-person & letter \\
\midrule
2-person & \textbf{0.46} & 0.50 \\
letter & 0.24 & \textbf{0.58} \\
\bottomrule
\end{tabular}

\end{table}

\begin{table}[htbp]
\centering
\footnotesize
\setlength{\tabcolsep}{4pt}
\caption{Indirect object identification on Llama-3.1-8B-Instruct, circuits discovered with \textsc{EAP-IG}.}
\label{tab:crossfaith-ioi-llama-3-1-8b-instruct-eap-ig-inputs}
\vspace{0.4em}

\textbf{syntax axis}\par\vspace{2pt}
\begin{tabular}{lcccc}
\toprule
\textbf{discovered on} & \multicolumn{4}{c}{\textbf{evaluated on}} \\
\cmidrule(l){2-5}
 & ABBA & BABA & 2-person & passive \\
\midrule
ABBA & \textbf{0.85} & 0.85 & 0.86 & 0.85 \\
BABA & 1.01 & \textbf{0.99} & 1.02 & 1.02 \\
2-person & 0.88 & 0.87 & \textbf{0.90} & 0.90 \\
passive & 0.94 & 0.89 & 0.92 & \textbf{0.88} \\
\bottomrule
\end{tabular}

\vspace{10pt}

\textbf{complexity axis}\par\vspace{2pt}
\begin{tabular}{lccc}
\toprule
\textbf{discovered on} & \multicolumn{3}{c}{\textbf{evaluated on}} \\
\cmidrule(l){2-4}
 & 2-person & filler & 3-person \\
\midrule
2-person & \textbf{0.90} & 0.83 & 0.57 \\
filler & 0.92 & \textbf{0.98} & 0.73 \\
3-person & 0.98 & 0.91 & \textbf{0.98} \\
\bottomrule
\end{tabular}

\vspace{10pt}

\textbf{domain axis}\par\vspace{2pt}
\begin{tabular}{lcc}
\toprule
\textbf{discovered on} & \multicolumn{2}{c}{\textbf{evaluated on}} \\
\cmidrule(l){2-3}
 & 2-person & letter \\
\midrule
2-person & \textbf{0.90} & 0.78 \\
letter & 0.79 & \textbf{0.92} \\
\bottomrule
\end{tabular}

\end{table}

\begin{table}[htbp]
\centering
\footnotesize
\setlength{\tabcolsep}{4pt}
\caption{Indirect object identification on Llama-3.1-8B-Instruct, circuits discovered with \textsc{EAP}.}
\label{tab:crossfaith-ioi-llama-3-1-8b-instruct-eap}
\vspace{0.4em}

\textbf{syntax axis}\par\vspace{2pt}
\begin{tabular}{lcccc}
\toprule
\textbf{discovered on} & \multicolumn{4}{c}{\textbf{evaluated on}} \\
\cmidrule(l){2-5}
 & ABBA & BABA & 2-person & passive \\
\midrule
ABBA & \textbf{0.60} & 0.70 & 0.72 & 0.63 \\
BABA & 0.58 & \textbf{0.66} & 0.64 & 0.57 \\
2-person & 0.47 & 0.60 & \textbf{0.62} & 0.51 \\
passive & 0.55 & 0.69 & 0.70 & \textbf{0.57} \\
\bottomrule
\end{tabular}

\vspace{10pt}

\textbf{complexity axis}\par\vspace{2pt}
\begin{tabular}{lccc}
\toprule
\textbf{discovered on} & \multicolumn{3}{c}{\textbf{evaluated on}} \\
\cmidrule(l){2-4}
 & 2-person & filler & 3-person \\
\midrule
2-person & \textbf{0.62} & 0.48 & 0.25 \\
filler & 0.01 & \textbf{0.02} & 0.01 \\
3-person & 0.03 & 0.02 & \textbf{0.02} \\
\bottomrule
\end{tabular}

\vspace{10pt}

\textbf{domain axis}\par\vspace{2pt}
\begin{tabular}{lcc}
\toprule
\textbf{discovered on} & \multicolumn{2}{c}{\textbf{evaluated on}} \\
\cmidrule(l){2-3}
 & 2-person & letter \\
\midrule
2-person & \textbf{0.62} & 0.34 \\
letter & 0.61 & \textbf{0.65} \\
\bottomrule
\end{tabular}

\end{table}

\begin{table}[htbp]
\centering
\footnotesize
\setlength{\tabcolsep}{4pt}
\caption{Entity binding on Qwen2.5-7B-Instruct, circuits discovered with \textsc{EAP-IG}.}
\label{tab:crossfaith-entity-binding-qwen2-5-7b-instruct-eap-ig-inputs}
\vspace{0.4em}

\textbf{syntax axis}\par\vspace{2pt}
\resizebox{\linewidth}{!}{%
\begin{tabular}{lcccccccccc}
\toprule
\textbf{discovered on} & \multicolumn{10}{c}{\textbf{evaluated on}} \\
\cmidrule(l){2-11}
 & 2-comma & 2-period & P1 & P2 & P3 & P4 & P5 & P6 & P7 & P8 \\
\midrule
2-comma & \textbf{1.10} & 1.07 & 0.77 & 0.77 & 0.64 & 0.59 & 0.59 & 0.69 & 0.63 & 0.60 \\
2-period & 0.53 & \textbf{0.45} & 0.42 & 0.63 & 0.48 & 0.41 & 0.35 & 0.37 & 0.29 & 0.18 \\
P1 & 0.94 & 0.80 & \textbf{0.80} & 0.89 & 0.78 & 0.75 & 0.70 & 0.73 & 0.68 & 0.27 \\
P2 & 1.15 & 1.24 & 0.94 & \textbf{1.11} & 1.02 & 0.98 & 0.89 & 0.92 & 0.83 & 0.33 \\
P3 & 1.66 & 1.70 & 1.10 & 1.12 & \textbf{1.08} & 1.02 & 1.05 & 1.00 & 0.83 & 0.39 \\
P4 & 1.23 & 1.39 & 1.35 & 1.18 & 1.03 & \textbf{0.98} & 0.94 & 0.94 & 0.80 & 0.63 \\
P5 & 0.98 & 1.19 & 0.64 & 0.89 & 0.77 & 0.78 & \textbf{0.75} & 0.74 & 0.60 & 0.29 \\
P6 & 1.59 & 1.73 & 1.25 & 1.13 & 1.05 & 1.04 & 0.94 & \textbf{0.95} & 0.79 & 0.51 \\
P7 & 1.37 & 1.60 & 1.28 & 1.14 & 1.03 & 0.99 & 1.00 & 0.94 & \textbf{0.91} & 0.82 \\
P8 & 1.45 & 1.50 & 1.22 & 1.06 & 0.96 & 0.92 & 0.92 & 0.95 & 0.86 & \textbf{0.93} \\
\bottomrule
\end{tabular}
}

\vspace{10pt}

\textbf{complexity axis}\par\vspace{2pt}
\begin{tabular}{lccccccc}
\toprule
\textbf{discovered on} & \multicolumn{7}{c}{\textbf{evaluated on}} \\
\cmidrule(l){2-8}
 & 2-comma & 3-comma & 4-comma & 5-comma & 6-comma & 7-comma & 8-comma \\
\midrule
2-comma & \textbf{1.11} & 0.82 & 0.68 & 0.63 & 0.67 & 0.63 & 0.68 \\
3-comma & 0.82 & \textbf{0.41} & 0.47 & 0.47 & 0.45 & 0.40 & 0.44 \\
4-comma & 1.17 & 0.90 & \textbf{0.73} & 0.77 & 0.79 & 0.76 & 0.83 \\
5-comma & 1.42 & 1.15 & 1.02 & \textbf{0.98} & 1.04 & 0.93 & 1.00 \\
6-comma & 1.23 & 1.03 & 0.95 & 0.88 & \textbf{0.90} & 0.87 & 0.89 \\
7-comma & 1.52 & 1.26 & 1.00 & 0.93 & 0.96 & \textbf{0.94} & 0.95 \\
8-comma & 1.98 & 1.55 & 1.14 & 1.04 & 1.06 & 1.02 & \textbf{1.06} \\
\bottomrule
\end{tabular}

\vspace{10pt}

\textbf{domain axis}\par\vspace{2pt}
\begin{tabular}{lcc}
\toprule
\textbf{discovered on} & \multicolumn{2}{c}{\textbf{evaluated on}} \\
\cmidrule(l){2-3}
 & 2-comma & 2-colour-box \\
\midrule
2-comma & \textbf{1.11} & 2.49 \\
2-colour-box & 0.54 & \textbf{0.57} \\
\bottomrule
\end{tabular}

\end{table}

\begin{table}[htbp]
\centering
\footnotesize
\setlength{\tabcolsep}{4pt}
\caption{Entity binding on Qwen2.5-7B-Instruct, circuits discovered with \textsc{EAP}. Only the complexity axis was measured for this cell.}
\label{tab:crossfaith-entity-binding-qwen2-5-7b-instruct-eap}
\vspace{0.4em}

\begin{tabular}{lccccccc}
\toprule
\textbf{discovered on} & \multicolumn{7}{c}{\textbf{evaluated on}} \\
\cmidrule(l){2-8}
 & 2-comma & 3-comma & 4-comma & 5-comma & 6-comma & 7-comma & 8-comma \\
\midrule
2-comma & \textbf{0.11} & 0.16 & 0.18 & 0.22 & 0.18 & 0.18 & 0.15 \\
3-comma & 0.19 & \textbf{0.10} & 0.11 & 0.15 & 0.13 & 0.10 & 0.13 \\
4-comma & 0.11 & 0.16 & \textbf{0.12} & 0.15 & 0.12 & 0.09 & 0.10 \\
5-comma & 0.21 & 0.12 & 0.11 & \textbf{0.13} & 0.11 & 0.10 & 0.10 \\
6-comma & 0.24 & 0.16 & 0.18 & 0.18 & \textbf{0.17} & 0.15 & 0.15 \\
7-comma & 0.25 & 0.17 & 0.23 & 0.24 & 0.18 & \textbf{0.15} & 0.18 \\
8-comma & 0.67 & 0.44 & 0.32 & 0.30 & 0.30 & 0.26 & \textbf{0.25} \\
\bottomrule
\end{tabular}

\end{table}

\begin{table}[htbp]
\centering
\footnotesize
\setlength{\tabcolsep}{4pt}
\caption{Entity binding on Llama-3.1-8B-Instruct, circuits discovered with \textsc{EAP-IG}.}
\label{tab:crossfaith-entity-binding-llama-3-1-8b-instruct-eap-ig-inputs}
\vspace{0.4em}

\textbf{syntax axis}\par\vspace{2pt}
\resizebox{\linewidth}{!}{%
\begin{tabular}{lcccccccccc}
\toprule
\textbf{discovered on} & \multicolumn{10}{c}{\textbf{evaluated on}} \\
\cmidrule(l){2-11}
 & 2-comma & 2-period & P1 & P2 & P3 & P4 & P5 & P6 & P7 & P8 \\
\midrule
2-comma & \textbf{0.65} & 0.79 & 1.05 & 0.73 & 0.75 & 0.79 & 0.76 & 0.80 & 0.79 & 0.78 \\
2-period & 0.47 & \textbf{0.70} & 0.51 & 0.58 & 0.60 & 0.61 & 0.61 & 0.63 & 0.66 & 0.80 \\
P1 & 0.58 & 0.64 & \textbf{0.97} & 0.96 & 0.88 & 0.90 & 0.88 & 0.85 & 0.76 & 0.52 \\
P2 & 0.79 & 0.88 & 0.87 & \textbf{1.00} & 0.96 & 0.97 & 0.94 & 0.83 & 0.80 & 0.70 \\
P3 & 0.72 & 0.81 & 0.59 & 1.12 & \textbf{0.93} & 0.96 & 0.94 & 0.92 & 0.87 & 0.85 \\
P4 & 0.66 & 0.77 & 0.78 & 1.15 & 0.95 & \textbf{0.96} & 0.95 & 0.85 & 0.83 & 0.64 \\
P5 & 0.83 & 1.04 & 0.81 & 1.04 & 0.90 & 0.93 & \textbf{0.91} & 0.87 & 0.87 & 0.80 \\
P6 & 0.71 & 0.82 & 0.72 & 1.19 & 1.03 & 1.06 & 1.02 & \textbf{0.97} & 0.90 & 0.84 \\
P7 & 0.71 & 0.88 & 1.17 & 1.10 & 0.97 & 1.00 & 1.00 & 0.98 & \textbf{1.00} & 0.95 \\
P8 & 0.67 & 0.83 & 0.79 & 1.02 & 0.95 & 0.99 & 0.98 & 0.98 & 0.96 & \textbf{0.91} \\
\bottomrule
\end{tabular}
}

\vspace{10pt}

\textbf{complexity axis}\par\vspace{2pt}
\begin{tabular}{lccccccc}
\toprule
\textbf{discovered on} & \multicolumn{7}{c}{\textbf{evaluated on}} \\
\cmidrule(l){2-8}
 & 2-comma & 3-comma & 4-comma & 5-comma & 6-comma & 7-comma & 8-comma \\
\midrule
2-comma & \textbf{0.65} & 0.77 & 0.74 & 0.76 & 0.80 & 0.83 & 0.79 \\
3-comma & 0.54 & \textbf{0.78} & 0.85 & 0.86 & 0.95 & 0.92 & 0.94 \\
4-comma & 0.75 & 0.95 & \textbf{0.97} & 1.02 & 1.02 & 1.05 & 0.97 \\
5-comma & 0.80 & 0.93 & 1.02 & \textbf{1.03} & 1.05 & 1.07 & 1.01 \\
6-comma & 0.79 & 0.85 & 0.84 & 0.87 & \textbf{0.82} & 0.86 & 0.81 \\
7-comma & 1.07 & 0.94 & 0.95 & 0.99 & 0.99 & \textbf{1.00} & 0.99 \\
8-comma & 0.87 & 0.86 & 0.89 & 0.95 & 0.95 & 0.94 & \textbf{0.92} \\
\bottomrule
\end{tabular}

\vspace{10pt}

\textbf{domain axis}\par\vspace{2pt}
\begin{tabular}{lcc}
\toprule
\textbf{discovered on} & \multicolumn{2}{c}{\textbf{evaluated on}} \\
\cmidrule(l){2-3}
 & 2-comma & 2-colour-box \\
\midrule
2-comma & \textbf{0.65} & 0.73 \\
2-colour-box & 0.68 & \textbf{0.83} \\
\bottomrule
\end{tabular}

\end{table}

\begin{table}[htbp]
\centering
\footnotesize
\setlength{\tabcolsep}{4pt}
\caption{Entity binding on Llama-3.1-8B-Instruct, circuits discovered with \textsc{EAP}.}
\label{tab:crossfaith-entity-binding-llama-3-1-8b-instruct-eap}
\vspace{0.4em}

\textbf{syntax axis}\par\vspace{2pt}
\resizebox{\linewidth}{!}{%
\begin{tabular}{lcccccccccc}
\toprule
\textbf{discovered on} & \multicolumn{10}{c}{\textbf{evaluated on}} \\
\cmidrule(l){2-11}
 & 2-comma & 2-period & P1 & P2 & P3 & P4 & P5 & P6 & P7 & P8 \\
\midrule
2-comma & \textbf{0.00} & 0.04 & 0.07 & 0.00 & -0.00 & -0.01 & -0.01 & 0.01 & -0.00 & 0.02 \\
2-period & 0.64 & \textbf{0.78} & 1.23 & 1.41 & 1.14 & 1.10 & 1.01 & 0.96 & 0.88 & 0.81 \\
P1 & 0.44 & 0.64 & \textbf{0.76} & 0.54 & 0.39 & 0.44 & 0.41 & 0.39 & 0.32 & 0.25 \\
P2 & 0.54 & 0.71 & 0.75 & \textbf{0.77} & 0.78 & 0.83 & 0.76 & 0.68 & 0.55 & 0.18 \\
P3 & 0.31 & 0.38 & 0.56 & 0.57 & \textbf{0.56} & 0.59 & 0.59 & 0.59 & 0.54 & 0.29 \\
P4 & 0.41 & 0.60 & 0.36 & 0.68 & 0.68 & \textbf{0.66} & 0.67 & 0.66 & 0.64 & 0.46 \\
P5 & 0.66 & 0.74 & 0.45 & 0.59 & 0.65 & 0.72 & \textbf{0.75} & 0.74 & 0.71 & 0.57 \\
P6 & 0.59 & 0.66 & 0.42 & 0.59 & 0.55 & 0.69 & 0.65 & \textbf{0.67} & 0.61 & 0.63 \\
P7 & 0.68 & 0.83 & 0.49 & 0.43 & 0.46 & 0.64 & 0.69 & 0.74 & \textbf{0.65} & 0.64 \\
P8 & 0.50 & 0.59 & 0.78 & 0.96 & 0.76 & 0.86 & 0.89 & 0.89 & 0.83 & \textbf{0.66} \\
\bottomrule
\end{tabular}
}

\vspace{10pt}

\textbf{complexity axis}\par\vspace{2pt}
\begin{tabular}{lccccccc}
\toprule
\textbf{discovered on} & \multicolumn{7}{c}{\textbf{evaluated on}} \\
\cmidrule(l){2-8}
 & 2-comma & 3-comma & 4-comma & 5-comma & 6-comma & 7-comma & 8-comma \\
\midrule
2-comma & \textbf{0.00} & 0.06 & 0.02 & 0.02 & 0.03 & 0.00 & 0.01 \\
3-comma & 0.51 & \textbf{0.57} & 0.68 & 0.66 & 0.66 & 0.63 & 0.62 \\
4-comma & 0.51 & 0.65 & \textbf{0.79} & 0.82 & 0.86 & 0.87 & 0.85 \\
5-comma & 0.98 & 0.91 & 1.01 & \textbf{1.02} & 1.05 & 1.09 & 1.03 \\
6-comma & 0.56 & 0.64 & 0.71 & 0.78 & \textbf{0.83} & 0.88 & 0.88 \\
7-comma & 0.85 & 0.79 & 0.84 & 0.90 & 0.97 & \textbf{0.96} & 0.96 \\
8-comma & 0.83 & 0.74 & 0.83 & 0.85 & 0.85 & 0.86 & \textbf{0.88} \\
\bottomrule
\end{tabular}

\vspace{10pt}

\textbf{domain axis}\par\vspace{2pt}
\begin{tabular}{lcc}
\toprule
\textbf{discovered on} & \multicolumn{2}{c}{\textbf{evaluated on}} \\
\cmidrule(l){2-3}
 & 2-comma & 2-colour-box \\
\midrule
2-comma & \textbf{0.00} & 0.02 \\
2-colour-box & 0.48 & \textbf{0.58} \\
\bottomrule
\end{tabular}

\end{table}

\begin{table}[htbp]
\centering
\footnotesize
\setlength{\tabcolsep}{4pt}
\caption{Arithmetic on Llama-3.1-8B-Instruct, circuits discovered with \textsc{EAP-IG}.}
\label{tab:crossfaith-arithmetic-llama-3-1-8b-instruct-eap-ig-inputs}
\vspace{0.4em}

\textbf{syntax axis}\par\vspace{2pt}
\begin{tabular}{lccc}
\toprule
\textbf{discovered on} & \multicolumn{3}{c}{\textbf{evaluated on}} \\
\cmidrule(l){2-4}
 & verbal-v1 & verbal-v2 & verbal-v3 \\
\midrule
verbal-v1 & \textbf{0.97} & 0.98 & 1.01 \\
verbal-v2 & 0.99 & \textbf{1.00} & 1.03 \\
verbal-v3 & 0.96 & 0.97 & \textbf{1.00} \\
\bottomrule
\end{tabular}

\vspace{10pt}

\textbf{complexity axis}\par\vspace{2pt}
\begin{tabular}{lcc}
\toprule
\textbf{discovered on} & \multicolumn{2}{c}{\textbf{evaluated on}} \\
\cmidrule(l){2-3}
 & 2-operand & 3-operand \\
\midrule
2-operand & \textbf{0.93} & 0.90 \\
3-operand & 0.93 & \textbf{0.90} \\
\bottomrule
\end{tabular}

\vspace{10pt}

\textbf{domain axis}\par\vspace{2pt}
\begin{tabular}{lcc}
\toprule
\textbf{discovered on} & \multicolumn{2}{c}{\textbf{evaluated on}} \\
\cmidrule(l){2-3}
 & 2-operand & verbal-v1 \\
\midrule
2-operand & \textbf{0.93} & 1.00 \\
verbal-v1 & 0.90 & \textbf{0.97} \\
\bottomrule
\end{tabular}

\end{table}

\begin{table}[htbp]
\centering
\footnotesize
\setlength{\tabcolsep}{4pt}
\caption{Arithmetic on Llama-3.1-8B-Instruct, circuits discovered with \textsc{EAP}.}
\label{tab:crossfaith-arithmetic-llama-3-1-8b-instruct-eap}
\vspace{0.4em}

\textbf{syntax axis}\par\vspace{2pt}
\begin{tabular}{lccc}
\toprule
\textbf{discovered on} & \multicolumn{3}{c}{\textbf{evaluated on}} \\
\cmidrule(l){2-4}
 & verbal-v1 & verbal-v2 & verbal-v3 \\
\midrule
verbal-v1 & \textbf{0.96} & 0.97 & 1.01 \\
verbal-v2 & 0.97 & \textbf{0.98} & 0.98 \\
verbal-v3 & 0.97 & 0.98 & \textbf{1.00} \\
\bottomrule
\end{tabular}

\vspace{10pt}

\textbf{complexity axis}\par\vspace{2pt}
\begin{tabular}{lcc}
\toprule
\textbf{discovered on} & \multicolumn{2}{c}{\textbf{evaluated on}} \\
\cmidrule(l){2-3}
 & 2-operand & 3-operand \\
\midrule
2-operand & \textbf{0.89} & 0.83 \\
3-operand & 0.91 & \textbf{0.91} \\
\bottomrule
\end{tabular}

\vspace{10pt}

\textbf{domain axis}\par\vspace{2pt}
\begin{tabular}{lcc}
\toprule
\textbf{discovered on} & \multicolumn{2}{c}{\textbf{evaluated on}} \\
\cmidrule(l){2-3}
 & 2-operand & verbal-v1 \\
\midrule
2-operand & \textbf{0.89} & 0.95 \\
verbal-v1 & 0.92 & \textbf{0.96} \\
\bottomrule
\end{tabular}

\end{table}

\begin{table}[htbp]
\centering
\footnotesize
\setlength{\tabcolsep}{4pt}
\caption{Sequence completion on GPT-2, circuits discovered with \textsc{EAP-IG}.}
\label{tab:crossfaith-sequence-completion-gpt2-eap-ig-inputs}
\vspace{0.4em}

\begin{tabular}{lccc}
\toprule
\textbf{discovered on} & \multicolumn{3}{c}{\textbf{evaluated on}} \\
\cmidrule(l){2-4}
 & 2-gram & 3-gram & 4-gram \\
\midrule
2-gram & \textbf{0.70} & 0.88 & 0.89 \\
3-gram & 0.76 & \textbf{0.82} & 0.86 \\
4-gram & 0.71 & 0.81 & \textbf{0.82} \\
\bottomrule
\end{tabular}

\end{table}

\begin{table}[htbp]
\centering
\footnotesize
\setlength{\tabcolsep}{4pt}
\caption{Sequence completion on GPT-2, circuits discovered with \textsc{EAP}.}
\label{tab:crossfaith-sequence-completion-gpt2-eap}
\vspace{0.4em}

\begin{tabular}{lccc}
\toprule
\textbf{discovered on} & \multicolumn{3}{c}{\textbf{evaluated on}} \\
\cmidrule(l){2-4}
 & 2-gram & 3-gram & 4-gram \\
\midrule
2-gram & \textbf{0.62} & 0.85 & 0.85 \\
3-gram & 0.78 & \textbf{0.83} & 0.87 \\
4-gram & 0.82 & 0.87 & \textbf{0.88} \\
\bottomrule
\end{tabular}

\end{table}

\begin{table}[htbp]
\centering
\footnotesize
\setlength{\tabcolsep}{4pt}
\caption{Sequence completion on Qwen2.5-7B-Instruct, circuits discovered with \textsc{EAP-IG}.}
\label{tab:crossfaith-sequence-completion-qwen2-5-7b-instruct-eap-ig-inputs}
\vspace{0.4em}

\begin{tabular}{lccc}
\toprule
\textbf{discovered on} & \multicolumn{3}{c}{\textbf{evaluated on}} \\
\cmidrule(l){2-4}
 & 2-gram & 3-gram & 4-gram \\
\midrule
2-gram & \textbf{0.57} & 0.73 & 0.76 \\
3-gram & 0.58 & \textbf{0.74} & 0.76 \\
4-gram & 0.59 & 0.74 & \textbf{0.77} \\
\bottomrule
\end{tabular}

\end{table}

\begin{table}[htbp]
\centering
\footnotesize
\setlength{\tabcolsep}{4pt}
\caption{Sequence completion on Qwen2.5-7B-Instruct, circuits discovered with \textsc{EAP}.}
\label{tab:crossfaith-sequence-completion-qwen2-5-7b-instruct-eap}
\vspace{0.4em}

\begin{tabular}{lccc}
\toprule
\textbf{discovered on} & \multicolumn{3}{c}{\textbf{evaluated on}} \\
\cmidrule(l){2-4}
 & 2-gram & 3-gram & 4-gram \\
\midrule
2-gram & \textbf{0.54} & 0.67 & 0.69 \\
3-gram & 0.56 & \textbf{0.73} & 0.76 \\
4-gram & 0.55 & 0.74 & \textbf{0.74} \\
\bottomrule
\end{tabular}

\end{table}

\begin{table}[htbp]
\centering
\footnotesize
\setlength{\tabcolsep}{4pt}
\caption{Sequence completion on Llama-3.1-8B-Instruct, circuits discovered with \textsc{EAP-IG}.}
\label{tab:crossfaith-sequence-completion-llama-3-1-8b-instruct-eap-ig-inputs}
\vspace{0.4em}

\begin{tabular}{lccc}
\toprule
\textbf{discovered on} & \multicolumn{3}{c}{\textbf{evaluated on}} \\
\cmidrule(l){2-4}
 & 2-gram & 3-gram & 4-gram \\
\midrule
2-gram & \textbf{0.70} & 0.84 & 0.89 \\
3-gram & 0.75 & \textbf{0.84} & 0.87 \\
4-gram & 0.78 & 0.85 & \textbf{0.89} \\
\bottomrule
\end{tabular}

\end{table}

\begin{table}[htbp]
\centering
\footnotesize
\setlength{\tabcolsep}{4pt}
\caption{Sequence completion on Llama-3.1-8B-Instruct, circuits discovered with \textsc{EAP}.}
\label{tab:crossfaith-sequence-completion-llama-3-1-8b-instruct-eap}
\vspace{0.4em}

\begin{tabular}{lccc}
\toprule
\textbf{discovered on} & \multicolumn{3}{c}{\textbf{evaluated on}} \\
\cmidrule(l){2-4}
 & 2-gram & 3-gram & 4-gram \\
\midrule
2-gram & \textbf{0.79} & 0.77 & 0.80 \\
3-gram & 0.78 & \textbf{0.85} & 0.88 \\
4-gram & 0.64 & 0.84 & \textbf{0.88} \\
\bottomrule
\end{tabular}

\end{table}
\subsection{Faithfulness across circuit sizes}
\label{app:exp1-curves}

The tables above report faithfulness at a single circuit size ($5\%$ of edges). This section shows how in-distribution and cross-variant faithfulness evolve with circuit size, one figure per (task, axis, model) cell, using EAP-IG. In each figure, panel (a) plots the faithfulness of the in-distribution circuit (Section~\ref{app:hd-setup}) on its own variant (solid) and on every other variant of the axis (dashed) as the circuit size grows from $0.1\%$ to $100\%$ of edges; panels (b) and (c) repeat, as heatmaps, the cross-variant faithfulness matrix at $5\%$ of edges (Section~\ref{app:crossfaith-howto}) and the pairwise Jaccard similarity of the top-5\% edge sets (Section~\ref{app:hd-overlap-summary}) for that cell. The EAP counterparts of every cell are reported numerically in the tables of Section~\ref{app:crossfaith-howto}; we omit their figures because they add no values beyond those tables and because, as noted there, EAP circuits fail to explain their own data in several cells at this budget.

\paragraph{Indirect object identification.}
Figures~\ref{fig:exp1-3panels-ioi-gpt2-complexity}--\ref{fig:exp1-3panels-ioi-llama-domain} cover the complexity axis (2-person against 3-person and filler), the syntax axis (ABBA against BABA, 2-person and passive) and the domain axis (2-person names against letter labels) on GPT-2 Small, Qwen2.5-7B-Instruct and Llama-3.1-8B-Instruct. The dataset variants are described in Appendix~\ref{app:dataset}, with example prompts in Table~\ref{tab:ioi-distributions}. Faithfulness curves separate unevenly across variants at small edge budgets on all three models, but the pattern is not a uniform drop: some variants (e.g., 3-person on GPT-2, letter on Llama-3.1-8B-Instruct) fall well below the in-distribution curve, while others track it closely or even exceed it at certain sizes. The axis that degrades most differs across models, indicating that circuit specialisation is real but not consistently anchored to a single dimension of surface variation.

\begin{figure}[htbp]
    \centering
    \includegraphics[width=\textwidth]{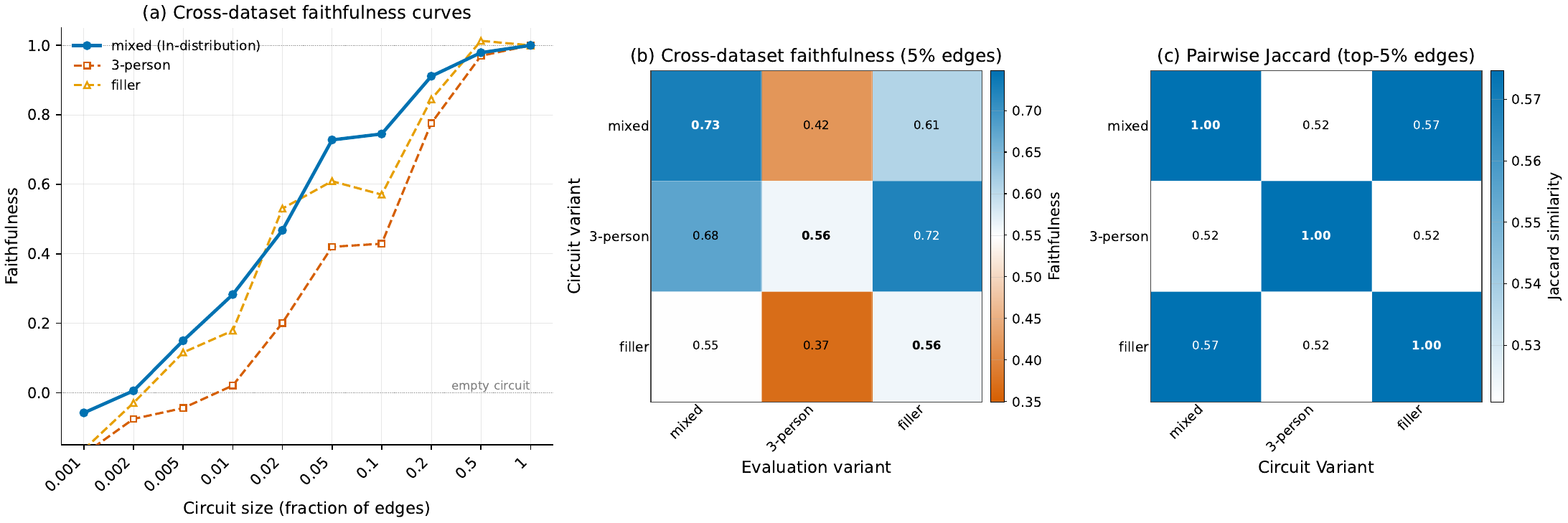}
    \caption{IOI complexity-axis results on GPT-2 Small (EAP-IG).
    \textbf{(a)} Faithfulness curves: the 2-person in-distribution circuit (solid) and OOD variants (dashed) across circuit sizes.
    \textbf{(b)} Cross-variant faithfulness matrix at 5\% of edges: row indicates discovery variant, column indicates evaluation variant.
    \textbf{(c)} Pairwise Jaccard similarity over top-5\% edges between circuits.}
    \label{fig:exp1-3panels-ioi-gpt2-complexity}
\end{figure}

\begin{figure}[htbp]
    \centering
    \includegraphics[width=\textwidth]{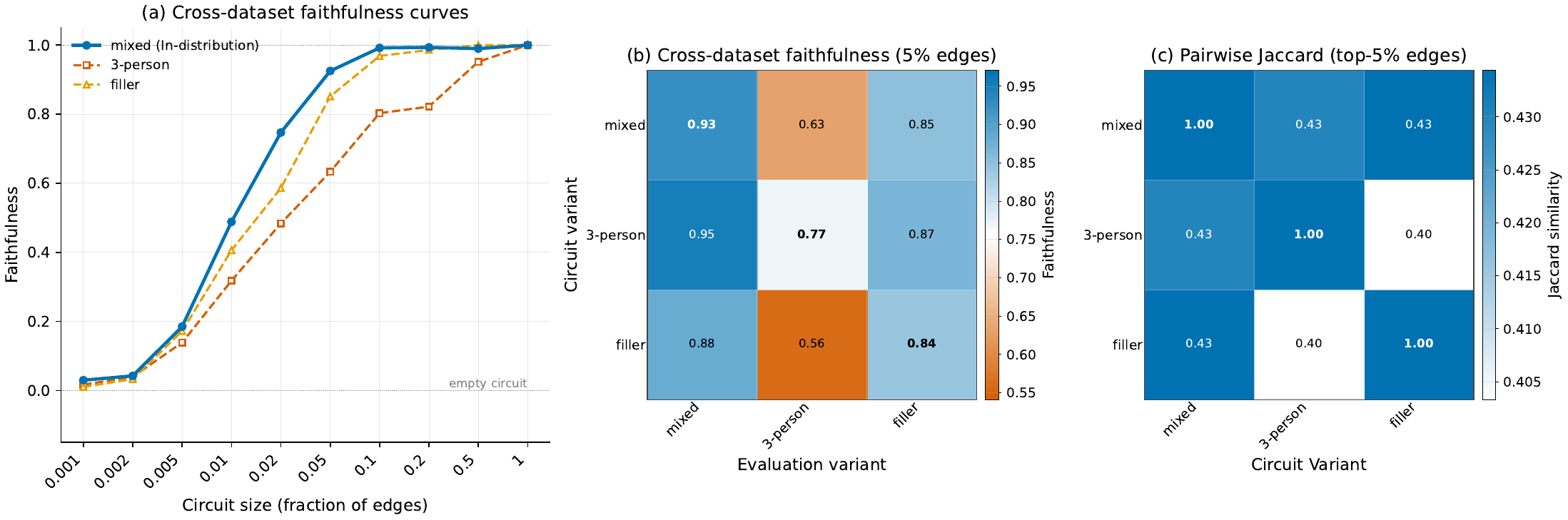}
    \caption{IOI complexity-axis results on Qwen2.5-7B-Instruct (EAP-IG). Panels as in Figure~\ref{fig:exp1-3panels-ioi-gpt2-complexity}.}
    \label{fig:exp1-3panels-ioi-qwen-complexity}
\end{figure}

\begin{figure}[htbp]
    \centering
    \includegraphics[width=\textwidth]{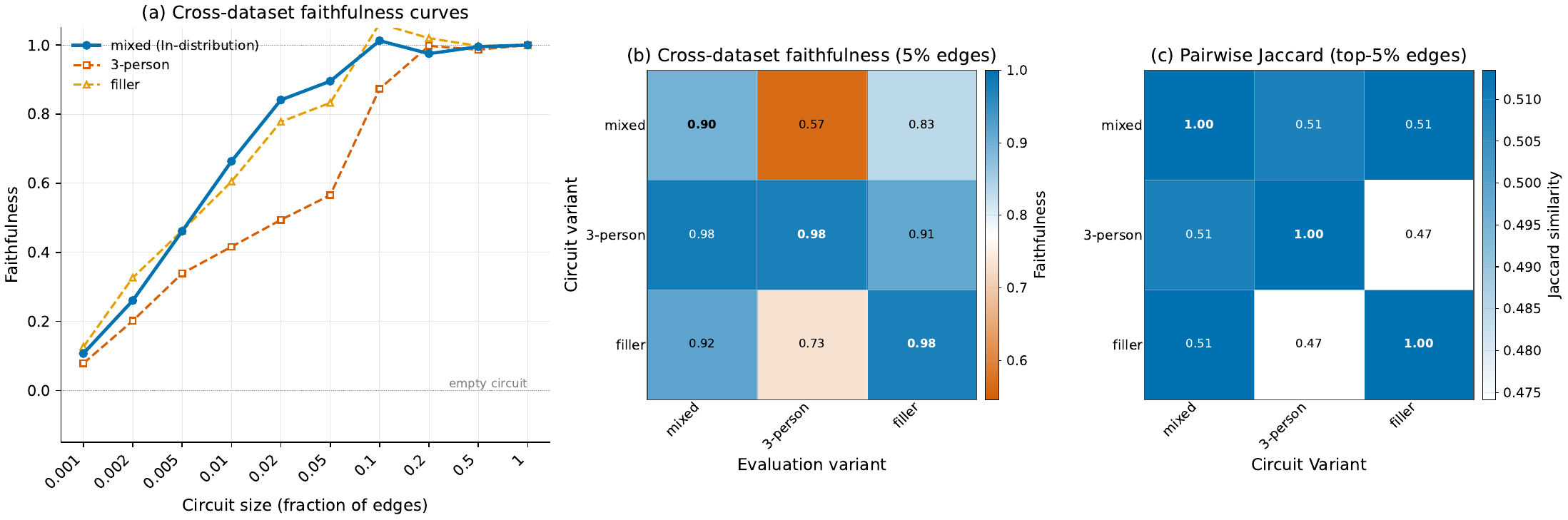}
    \caption{IOI complexity-axis results on Llama-3.1-8B-Instruct (EAP-IG). Panels as in Figure~\ref{fig:exp1-3panels-ioi-gpt2-complexity}.}
    \label{fig:exp1-3panels-ioi-llama-complexity}
\end{figure}

\begin{figure}[htbp]
    \centering
    \includegraphics[width=\textwidth]{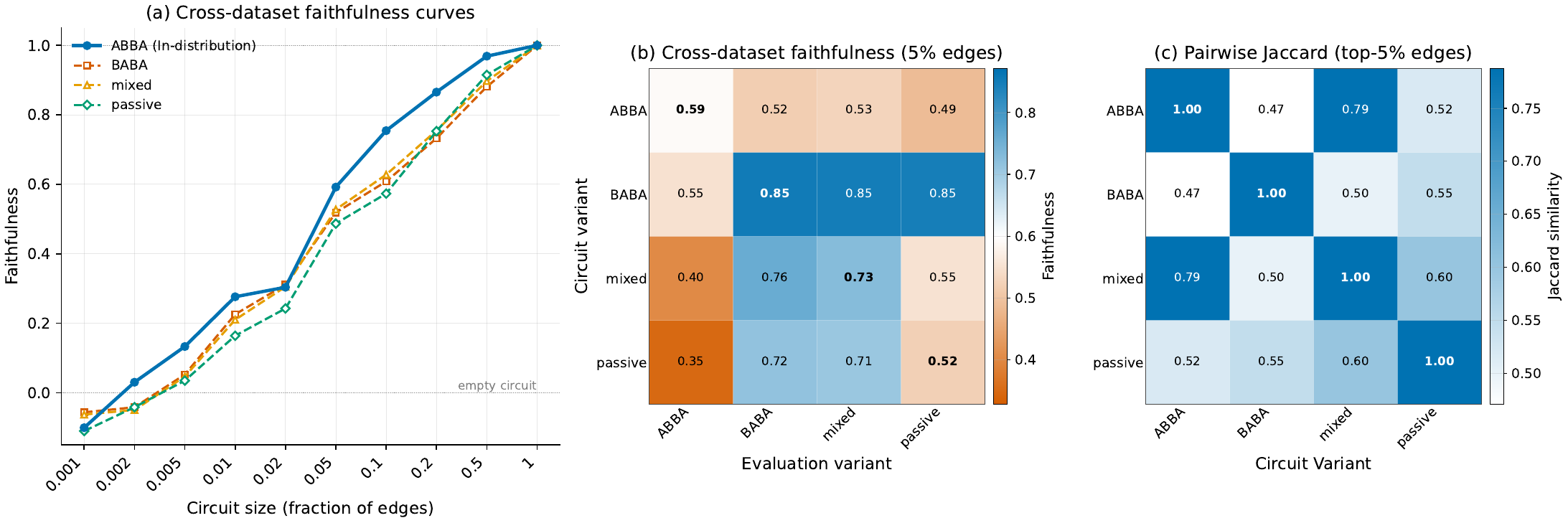}
    \caption{IOI syntax-axis results on GPT-2 Small (EAP-IG). \textbf{(a)} Faithfulness curves: the ABBA in-distribution circuit (solid) and OOD variants (dashed). \textbf{(b)} Cross-variant faithfulness matrix at 5\% of edges. \textbf{(c)} Pairwise Jaccard similarity over top-5\% edges.}
    \label{fig:exp1-3panels-ioi-gpt2-syntax}
\end{figure}

\begin{figure}[htbp]
    \centering
    \includegraphics[width=\textwidth]{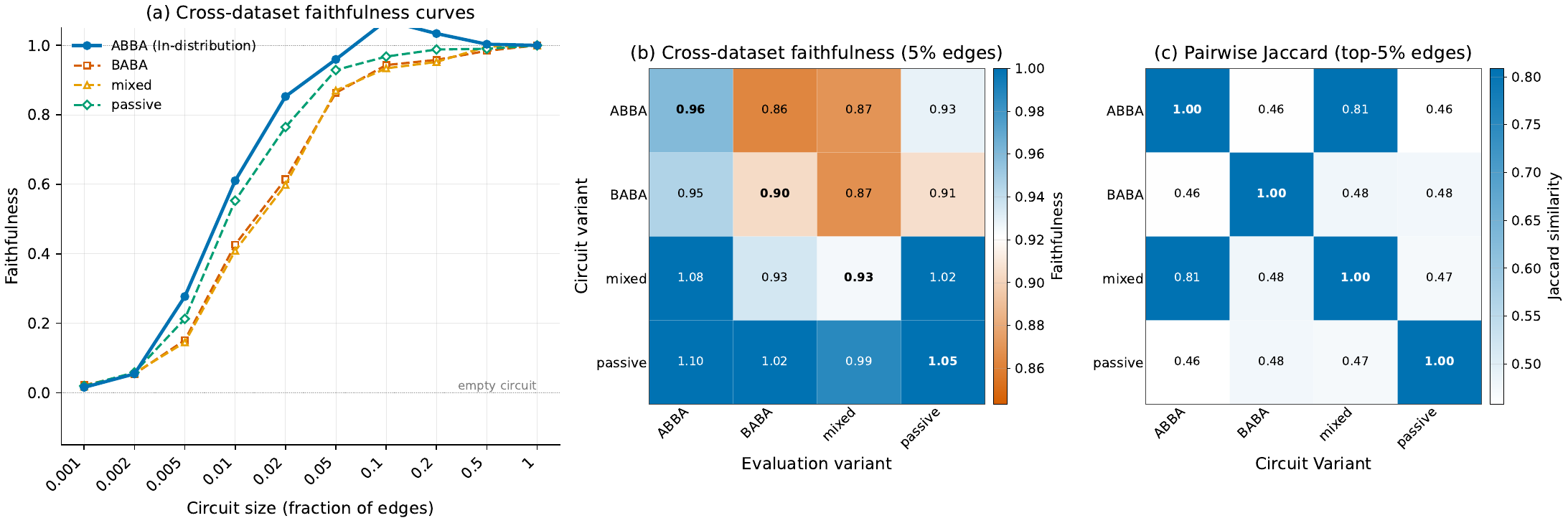}
    \caption{IOI syntax-axis results on Qwen2.5-7B-Instruct (EAP-IG). Panels as in Figure~\ref{fig:exp1-3panels-ioi-gpt2-syntax}.}
    \label{fig:exp1-3panels-ioi-qwen-syntax}
\end{figure}

\begin{figure}[htbp]
    \centering
    \includegraphics[width=\textwidth]{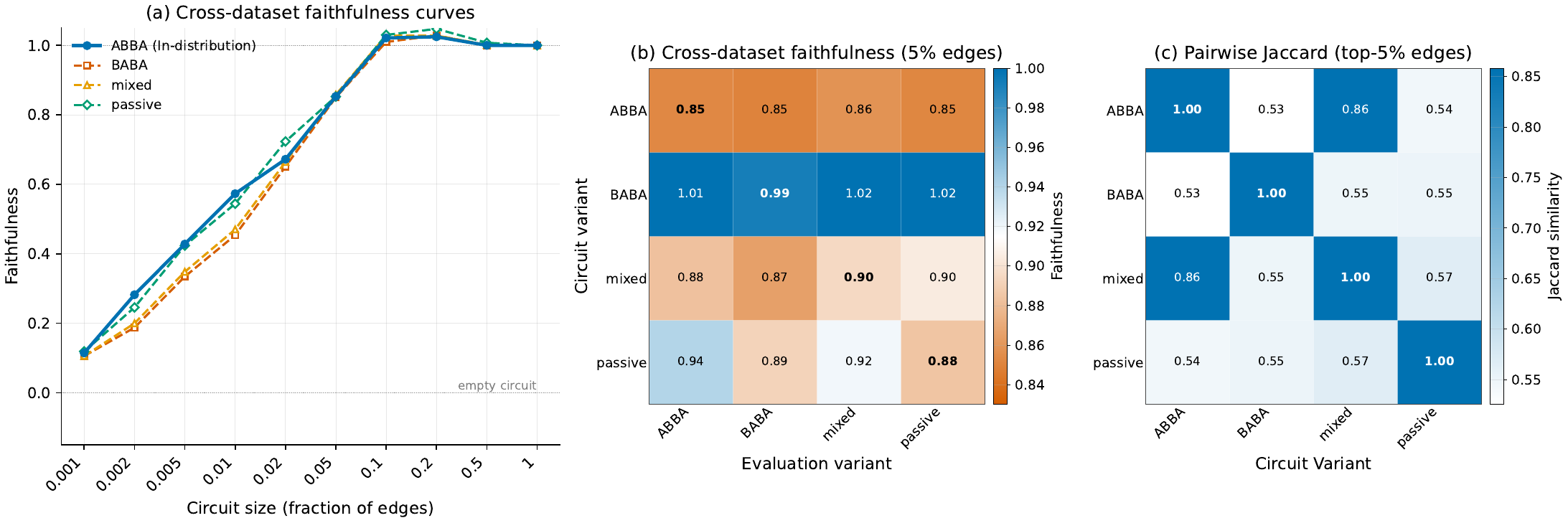}
    \caption{IOI syntax-axis results on Llama-3.1-8B-Instruct (EAP-IG). Panels as in Figure~\ref{fig:exp1-3panels-ioi-gpt2-syntax}.}
    \label{fig:exp1-3panels-ioi-llama-syntax}
\end{figure}

\begin{figure}[htbp]
    \centering
    \includegraphics[width=\textwidth]{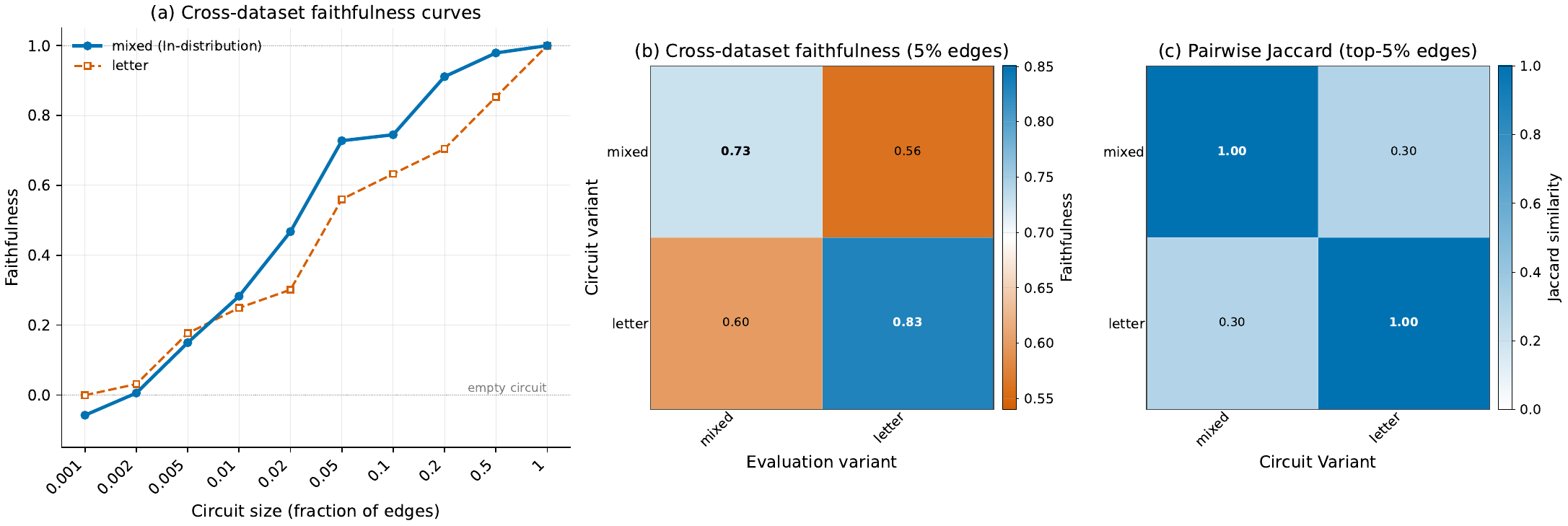}
    \caption{IOI domain-axis results on GPT-2 Small (EAP-IG). \textbf{(a)} Faithfulness curves: the person-name (2-person) in-distribution circuit (solid) and the letter-label OOD variant (dashed). \textbf{(b)} Cross-variant faithfulness matrix at 5\% of edges. \textbf{(c)} Pairwise Jaccard similarity over top-5\% edges.}
    \label{fig:exp1-3panels-ioi-gpt2-domain}
\end{figure}

\begin{figure}[htbp]
    \centering
    \includegraphics[width=\textwidth]{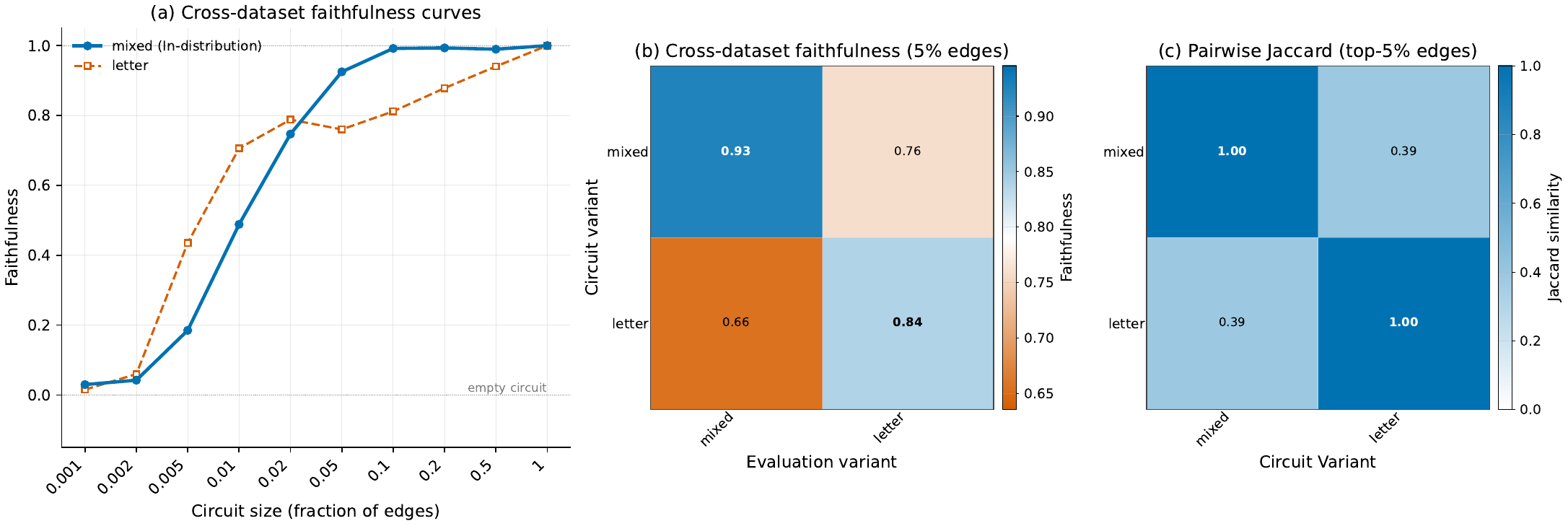}
    \caption{IOI domain-axis results on Qwen2.5-7B-Instruct (EAP-IG). Panels as in Figure~\ref{fig:exp1-3panels-ioi-gpt2-domain}.}
    \label{fig:exp1-3panels-ioi-qwen-domain}
\end{figure}

\begin{figure}[htbp]
    \centering
    \includegraphics[width=\textwidth]{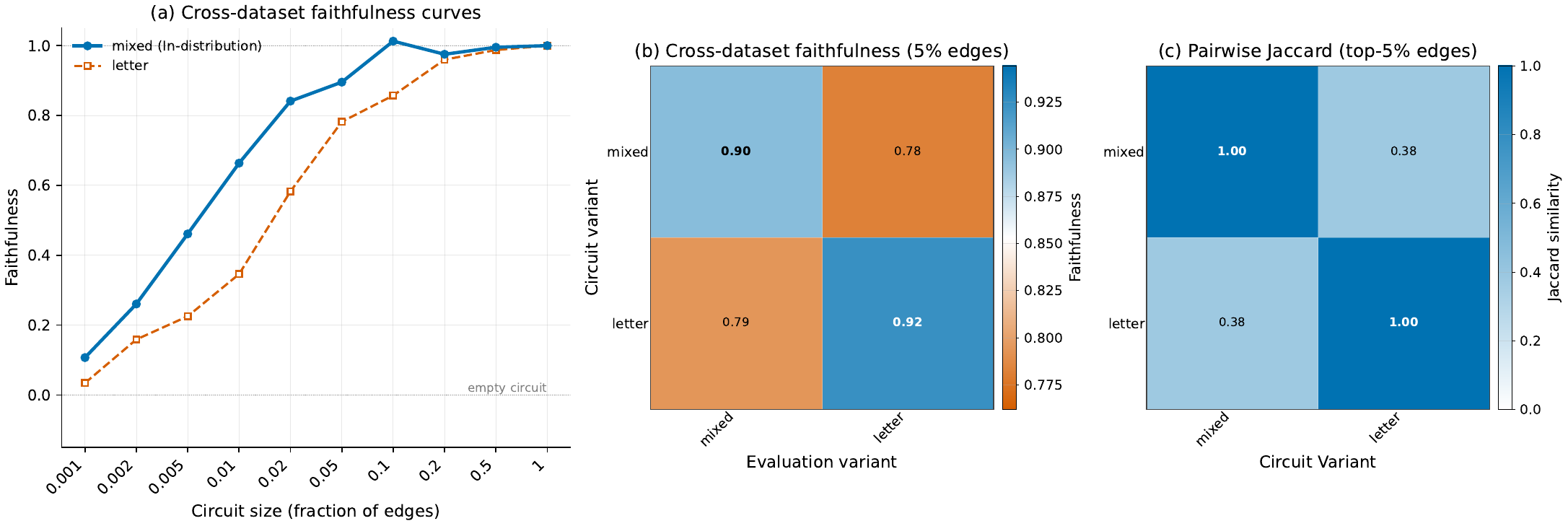}
    \caption{IOI domain-axis results on Llama-3.1-8B-Instruct (EAP-IG). Panels as in Figure~\ref{fig:exp1-3panels-ioi-gpt2-domain}.}
    \label{fig:exp1-3panels-ioi-llama-domain}
\end{figure}

\paragraph{Entity binding.}
Figures~\ref{fig:exp1-eb-qwen-complexity}--\ref{fig:exp1-3panels-eb-llama-domain} cover the complexity axis (2-comma against 3- to 8-comma), the position sub-axis of the syntax axis (P1 against P2--P8, on a fixed 8-pair template), the delimiter sub-axis (2-comma against 2-period) and the domain axis (letter-labelled boxes against colour-labelled boxes) on Qwen2.5-7B-Instruct and Llama-3.1-8B-Instruct; GPT-2 Small is excluded by the accuracy criterion (Table~\ref{tab:accuracy}). The dataset variants are described in Appendix~\ref{app:dataset}, with example prompts in Table~\ref{tab:entity-binding-distributions}. On the position sub-axis the P1 curve degrades monotonically with distance from P1 on Qwen2.5-7B-Instruct, mirroring the smooth band structure of the Jaccard matrix in Figure~\ref{fig:exp1-eb-qwen-combined}(c) and the transfer failures of Table~\ref{tab:s42-twoway}.

\begin{figure}[htbp]
    \centering
    \includegraphics[width=\textwidth]{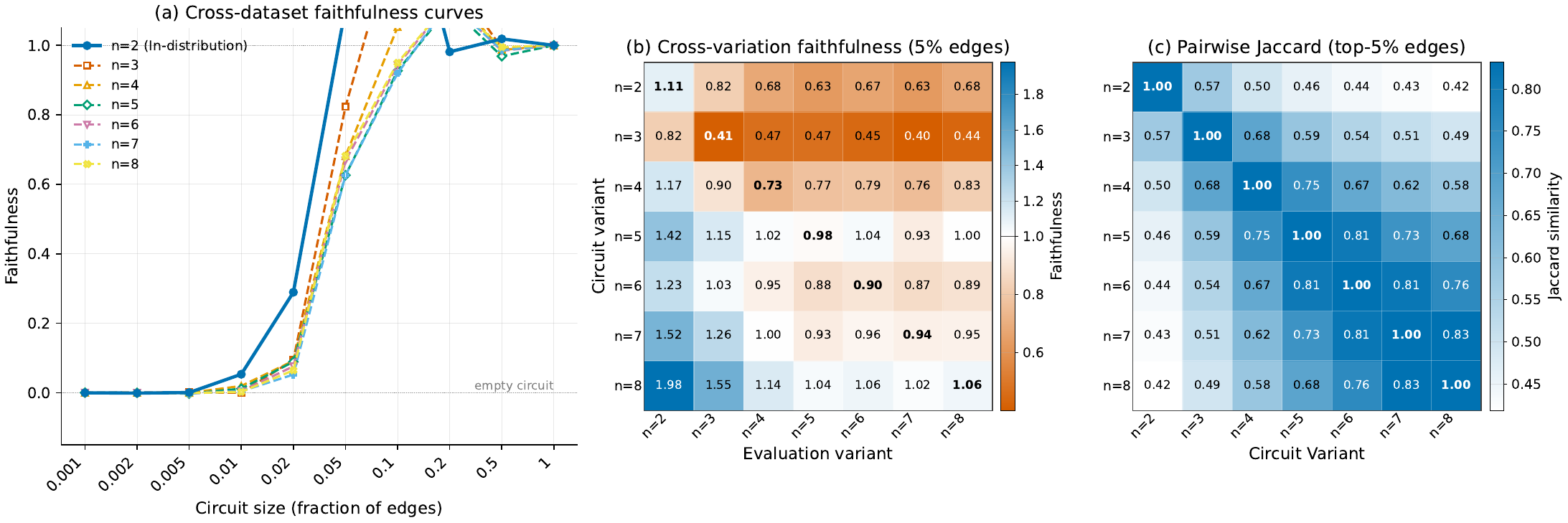}
    \caption{Complexity-variant analysis of entity binding on Qwen2.5-7B-Instruct, three views of the seven complexity-variant circuits ($n \in \{2, \ldots, 8\}$ entity--attribute pairs). \textbf{(a)} Faithfulness curves of the $n{=}2$ circuit evaluated against all seven variants. \textbf{(b)} Cross-variant faithfulness matrix at 5\% edge budget; circuits discovered on larger-$n$ variants over-recover on smaller-$n$ evaluations. \textbf{(c)} Pairwise Jaccard similarity over the top-5\% attribution-ranked edges. Circuits are discovered with EAP-IG.}
    \label{fig:exp1-eb-qwen-complexity}
\end{figure}

\begin{figure}[htbp]
    \centering
    \includegraphics[width=\textwidth]{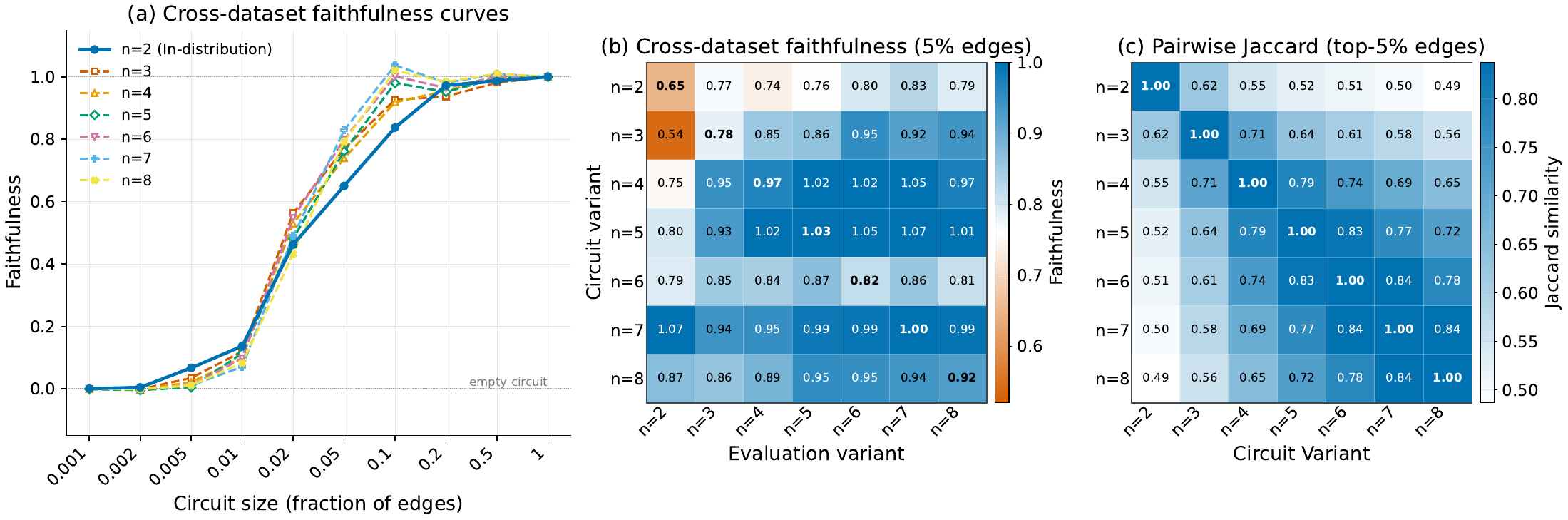}
    \caption{Complexity-variant analysis of entity binding on Llama-3.1-8B-Instruct, same setup as Figure~\ref{fig:exp1-eb-qwen-complexity}. \textbf{(a)} Faithfulness curves of the $n{=}2$ circuit; on Llama, larger-$n$ circuits actually exceed the in-distribution curve at moderate budgets. \textbf{(b)} Cross-variant faithfulness matrix; the $n{=}2$ row is uniformly low, indicating that the $n{=}2$ circuit fails to recover larger-$n$ behavior. \textbf{(c)} Pairwise Jaccard similarity over the top-5\% attribution-ranked edges, exhibiting the same smooth band structure as Qwen.}
    \label{fig:exp1-eb-llama-complexity}
\end{figure}

\begin{figure}[htbp]
    \centering
    \includegraphics[width=\textwidth]{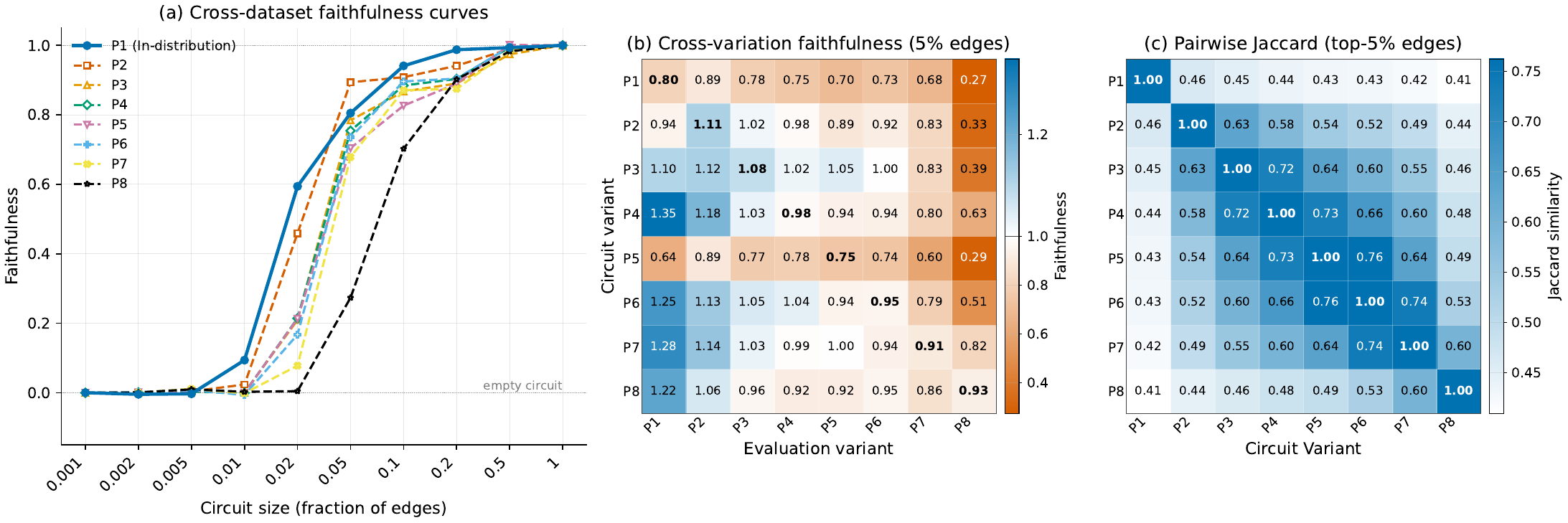}
    \caption{Position-variant analysis of entity binding on Qwen2.5-7B-Instruct, three views of the same eight position-variant circuits (P1--P8). \textbf{(a)} Faithfulness curves of the P1 circuit evaluated against all eight variants; the curve degrades monotonically with distance from P1. \textbf{(b)} Cross-variant faithfulness matrix at 5\% edge budget. \textbf{(c)} Pairwise Jaccard similarity over the top-5\% attribution-ranked edges. Circuits are discovered with EAP-IG.}
    \label{fig:exp1-eb-qwen-combined}
\end{figure}

\begin{figure}[htbp]
    \centering
    \includegraphics[width=\textwidth]{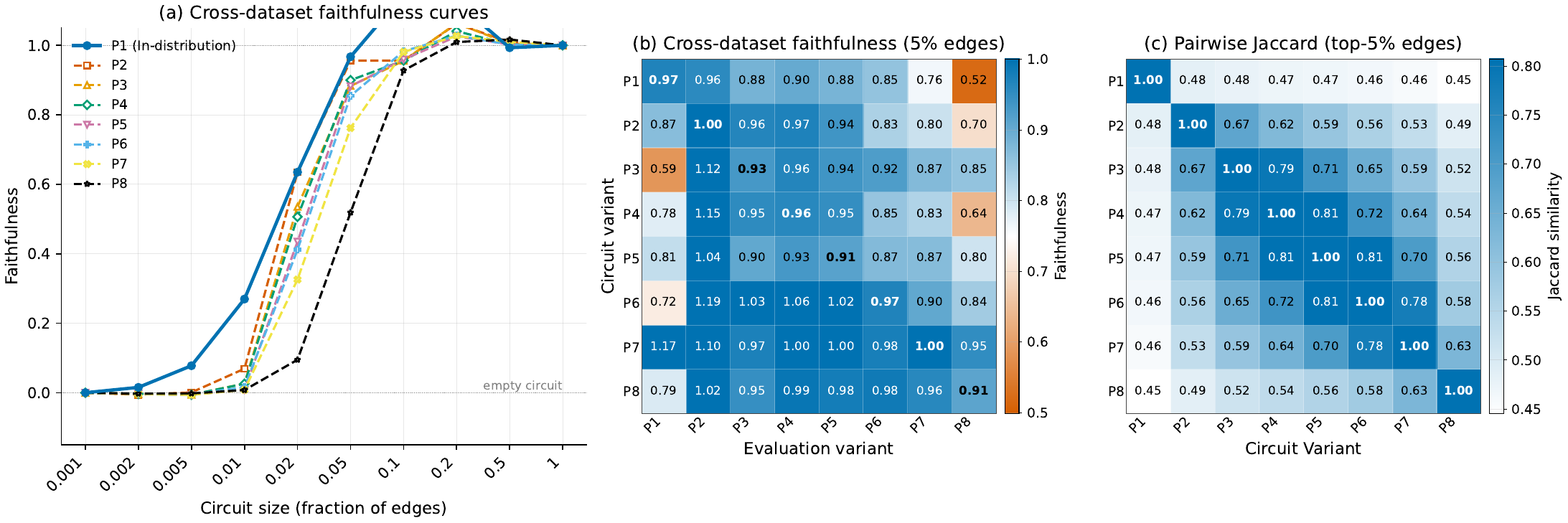}
    \caption{Position-variant analysis of entity binding on Llama-3.1-8B-Instruct (EAP-IG). Panels as in Figure~\ref{fig:exp1-eb-qwen-combined}.}
    \label{fig:exp1-3panels-eb-llama-syntax}
\end{figure}

\begin{figure}[htbp]
    \centering
    \includegraphics[width=\textwidth]{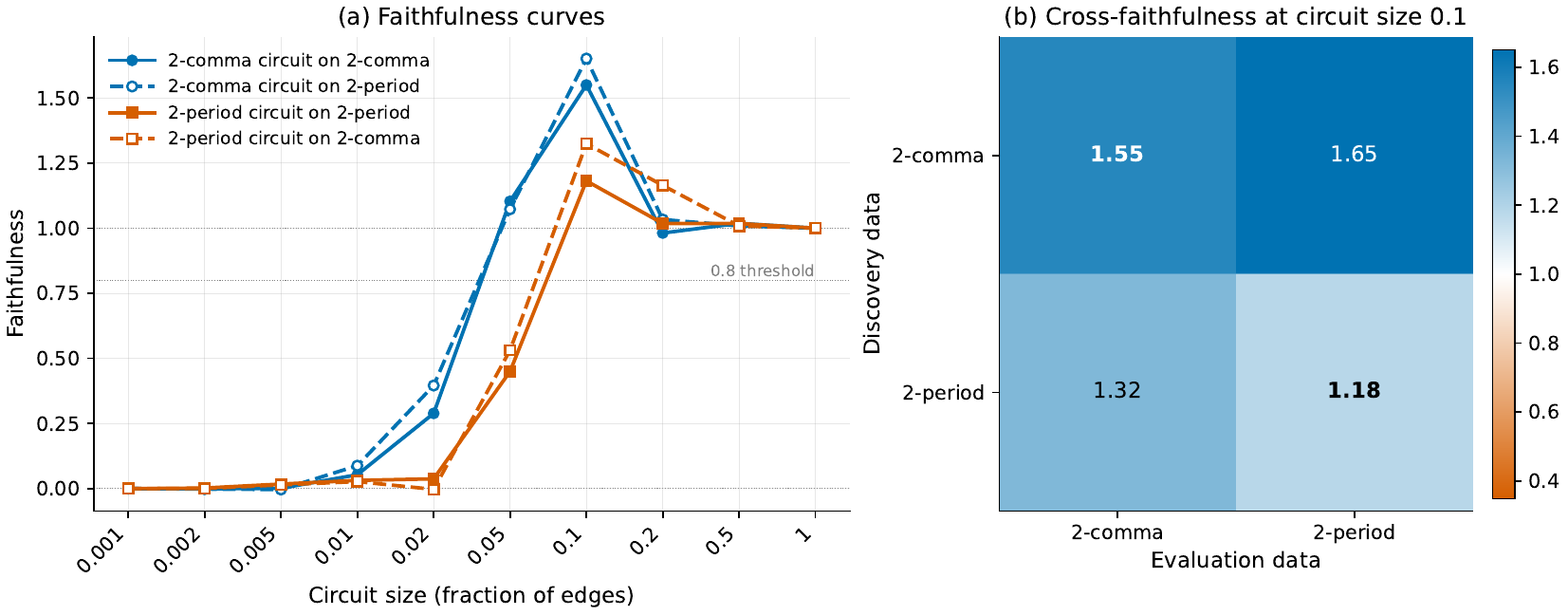}
    \caption{Delimiter-variant analysis of entity binding on Qwen2.5-7B-Instruct, comparing 2-comma and 2-period variants. \textbf{(a)} Faithfulness curves of each circuit evaluated on each distribution. \textbf{(b)} Cross-variant faithfulness matrix, shown here at circuit size 0.1 rather than 0.05; the 0.05 values are in Table~\ref{tab:crossfaith-entity-binding-qwen2-5-7b-instruct-eap-ig-inputs}.}
    \label{fig:exp1-eb-qwen-delimiter}
\end{figure}

\begin{figure}[htbp]
    \centering
    \includegraphics[width=\textwidth]{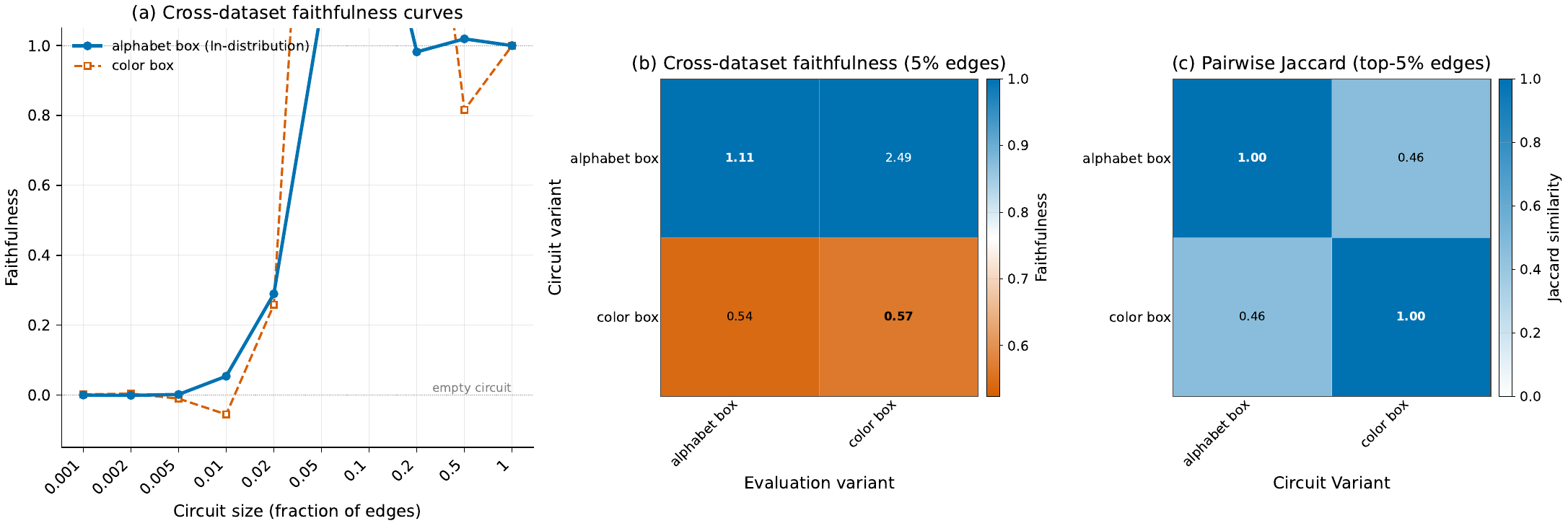}
    \caption{Entity binding domain-axis results on Qwen2.5-7B-Instruct (EAP-IG). \textbf{(a)} Faithfulness curves: the letter-label in-distribution circuit (solid) and the colour-label OOD variant (dashed). \textbf{(b)} Cross-variant faithfulness matrix at 5\% of edges. \textbf{(c)} Pairwise Jaccard similarity over top-5\% edges. This pair is excluded from the transfer analysis because the colour-box circuit's noise floor is unstable (Section~\ref{app:faithfulness-floor}).}
    \label{fig:exp1-3panels-eb-qwen-domain}
\end{figure}

\begin{figure}[htbp]
    \centering
    \includegraphics[width=\textwidth]{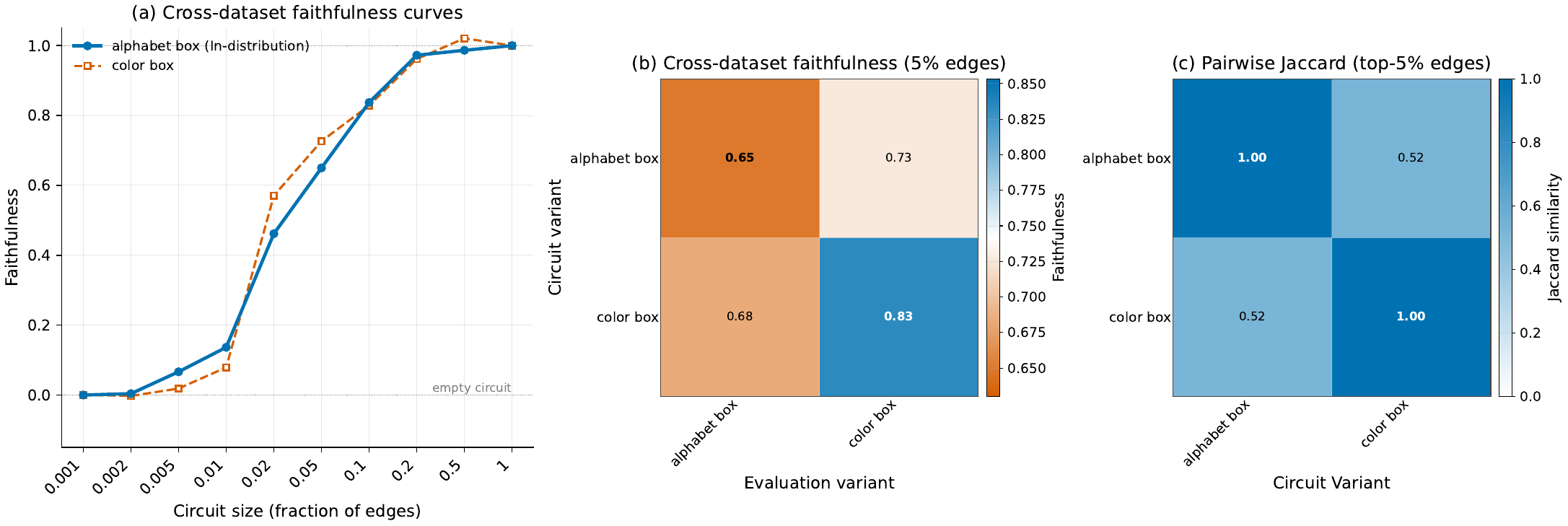}
    \caption{Entity binding domain-axis results on Llama-3.1-8B-Instruct (EAP-IG). Panels as in Figure~\ref{fig:exp1-3panels-eb-qwen-domain}.}
    \label{fig:exp1-3panels-eb-llama-domain}
\end{figure}

\paragraph{Arithmetic addition.}
Figures~\ref{fig:exp1-3panels-arithmetic-llama-complexity}--\ref{fig:exp1-3panels-arithmetic-llama-domain} cover the complexity axis (2-operand against 3-operand), the syntax axis (three natural-language phrasings, verbal v1 against v2 and v3, with operands and answers held constant) and the domain axis (symbolic 2-operand against verbal v1) on Llama-3.1-8B-Instruct, the only model that meets the accuracy criterion on this task (GPT-2 Small and Qwen2.5-7B-Instruct tokenise most two- and three-digit numbers as multiple tokens; Appendix~\ref{app:dataset}). Example prompts are in Table~\ref{tab:arithmetic-distributions}.

\begin{figure}[htbp]
    \centering
    \includegraphics[width=\textwidth]{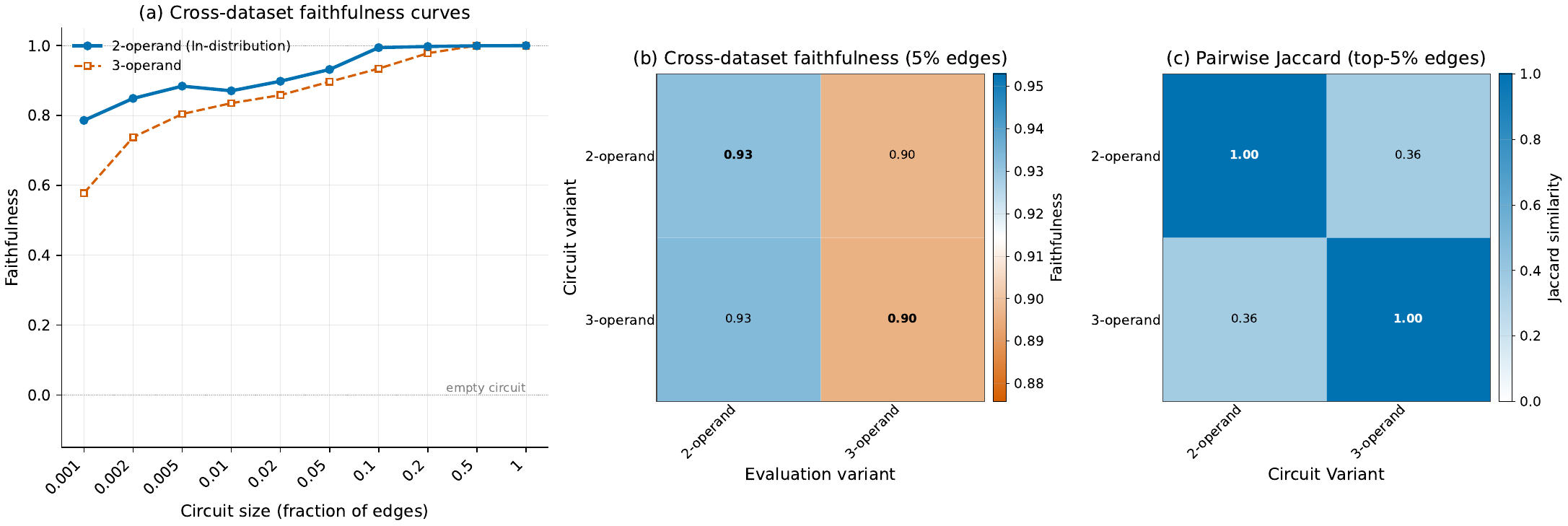}
    \caption{Arithmetic complexity-axis results on Llama-3.1-8B-Instruct (EAP-IG). \textbf{(a)} Faithfulness curves: 2-operand in-distribution circuit (solid) and 3-operand OOD variant (dashed). \textbf{(b)} Cross-variant faithfulness matrix at 5\% of edges. \textbf{(c)} Pairwise Jaccard similarity over top-5\% edges.}
    \label{fig:exp1-3panels-arithmetic-llama-complexity}
\end{figure}

\begin{figure}[htbp]
    \centering
    \includegraphics[width=\textwidth]{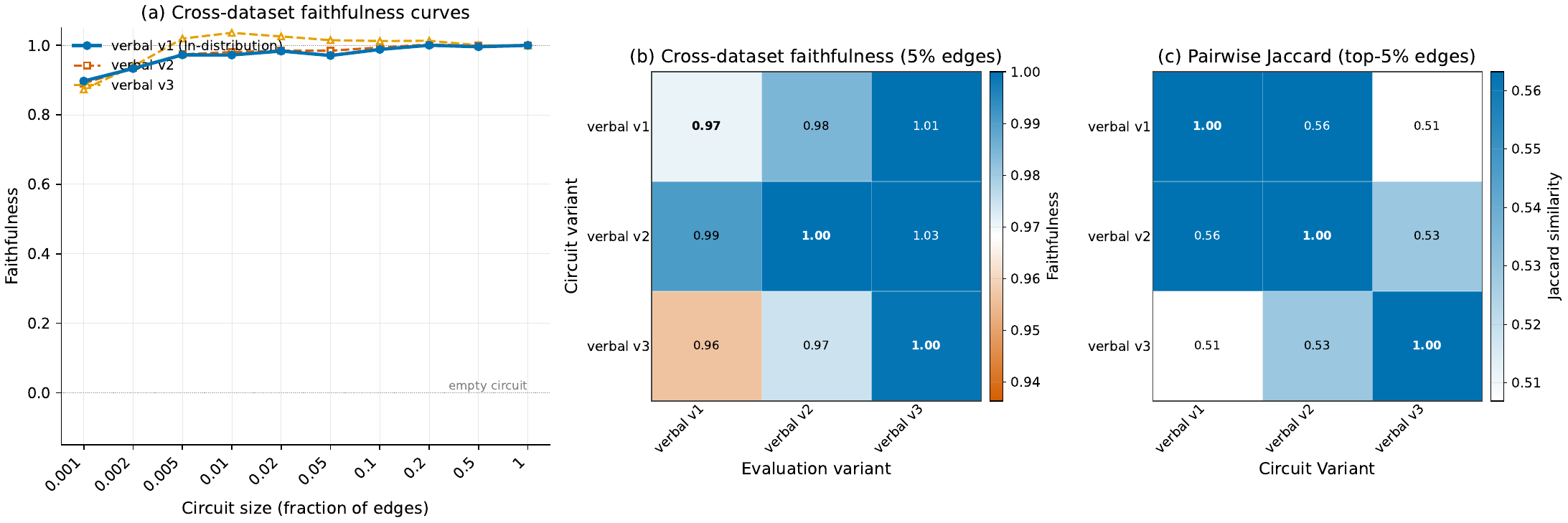}
    \caption{Arithmetic syntax-axis results on Llama-3.1-8B-Instruct (EAP-IG): the verbal v1 in-distribution circuit against verbal v2 and v3. Panels as in Figure~\ref{fig:exp1-3panels-arithmetic-llama-complexity}.}
    \label{fig:exp1-3panels-arithmetic-llama-syntax}
\end{figure}

\begin{figure}[htbp]
    \centering
    \includegraphics[width=\textwidth]{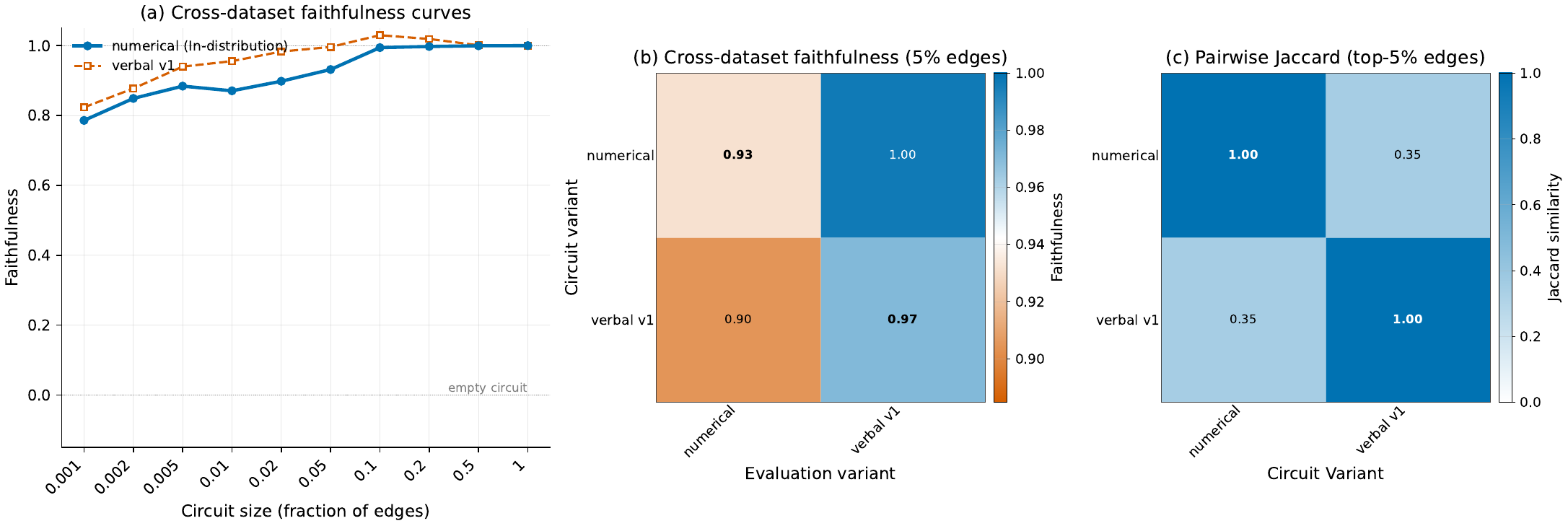}
    \caption{Arithmetic domain-axis results on Llama-3.1-8B-Instruct (EAP-IG): the symbolic 2-operand in-distribution circuit against verbal v1. Panels as in Figure~\ref{fig:exp1-3panels-arithmetic-llama-complexity}.}
    \label{fig:exp1-3panels-arithmetic-llama-domain}
\end{figure}

\paragraph{Sequence completion.}
Figures~\ref{fig:exp1-3panels-seqcomp-gpt2-complexity}--\ref{fig:exp1-3panels-seqcomp-llama-complexity} cover the complexity axis (2-gram against 3-gram and 4-gram) on GPT-2 Small, Qwen2.5-7B-Instruct and Llama-3.1-8B-Instruct. Sequence completion is studied along the complexity axis only: there is no surface-form (syntax) or lexical-domain (domain) variation that preserves the task's repetition structure. Example prompts are in Table~\ref{tab:sequence-distributions}.

\begin{figure}[htbp]
    \centering
    \includegraphics[width=\textwidth]{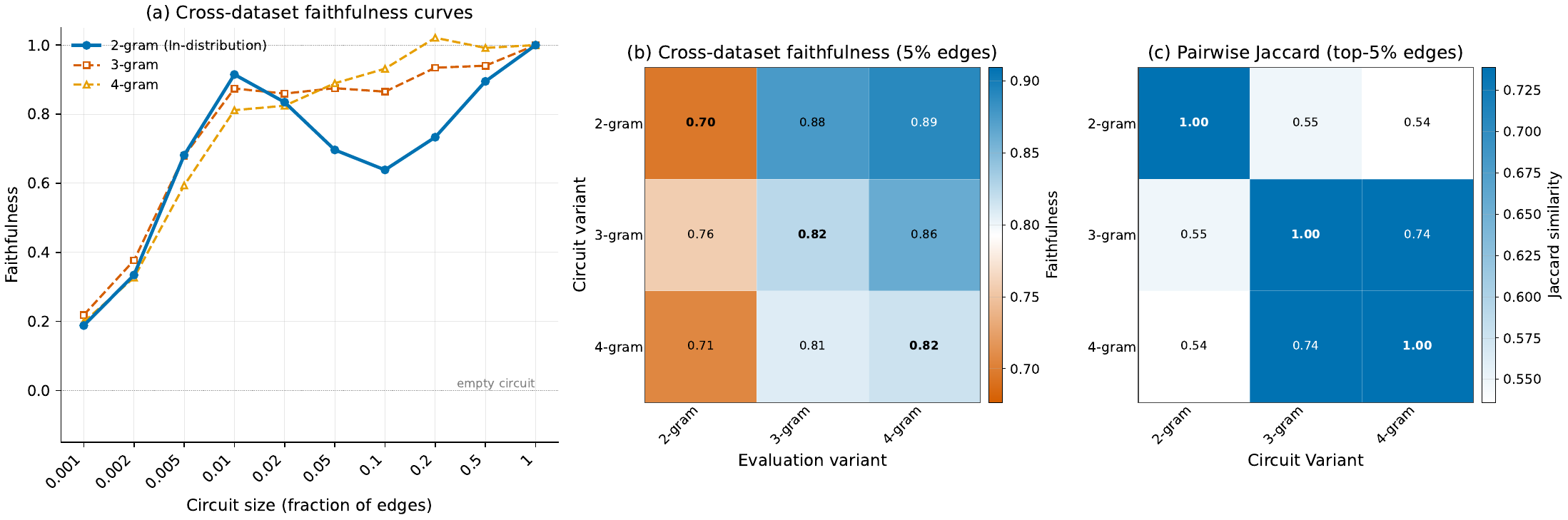}
    \caption{Sequence completion complexity-axis results on GPT-2 Small (EAP-IG). \textbf{(a)} Faithfulness curves: 2-gram in-distribution circuit (solid) and 3-gram, 4-gram OOD variants (dashed). \textbf{(b)} Cross-variant faithfulness matrix at 5\% of edges. \textbf{(c)} Pairwise Jaccard similarity over top-5\% edges.}
    \label{fig:exp1-3panels-seqcomp-gpt2-complexity}
\end{figure}

\begin{figure}[htbp]
    \centering
    \includegraphics[width=\textwidth]{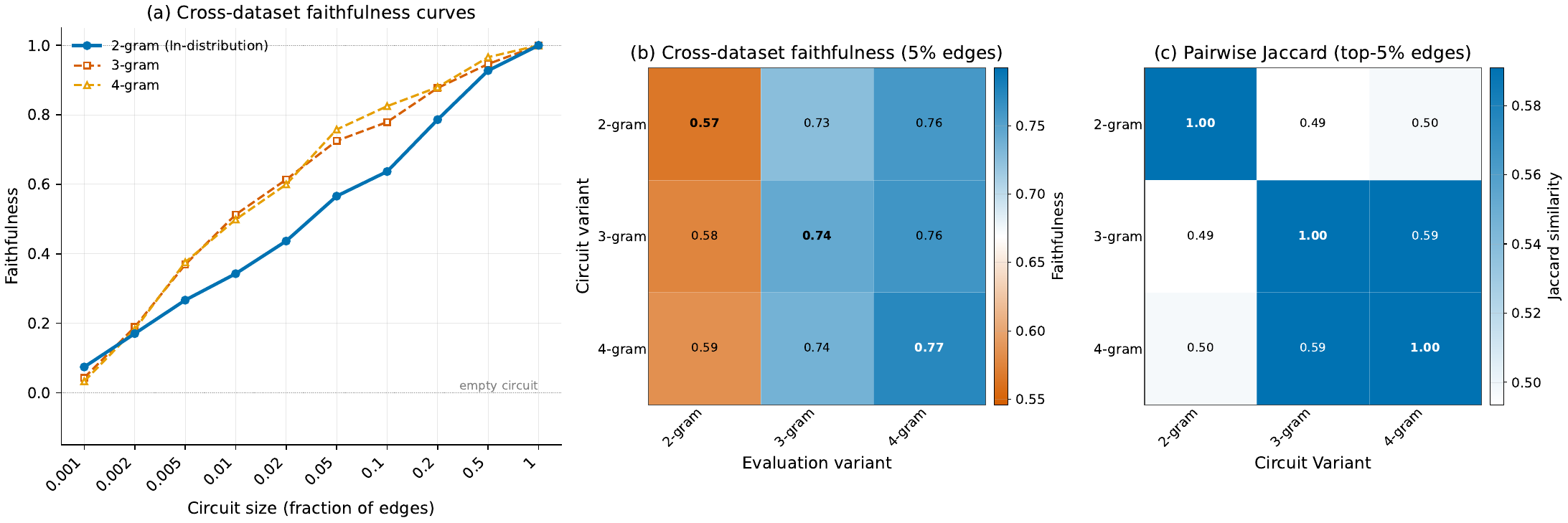}
    \caption{Sequence completion complexity-axis results on Qwen2.5-7B-Instruct (EAP-IG). Panels as in Figure~\ref{fig:exp1-3panels-seqcomp-gpt2-complexity}.}
    \label{fig:exp1-3panels-seqcomp-qwen-complexity}
\end{figure}

\begin{figure}[htbp]
    \centering
    \includegraphics[width=\textwidth]{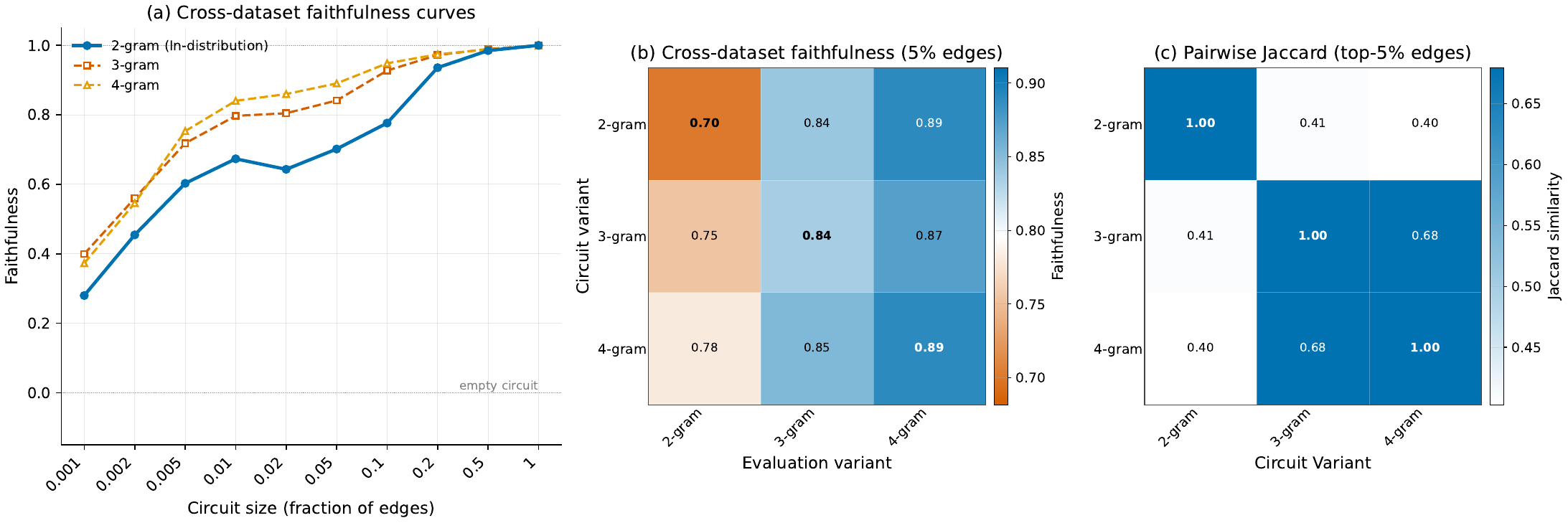}
    \caption{Sequence completion complexity-axis results on Llama-3.1-8B-Instruct (EAP-IG). Panels as in Figure~\ref{fig:exp1-3panels-seqcomp-gpt2-complexity}.}
    \label{fig:exp1-3panels-seqcomp-llama-complexity}
\end{figure}
\section{\approach{} setup and Additional Results}
\label{app:dcd}

\subsection{Algorithm for Instantiation}
\begin{algorithm}[t]
\caption{$\Pi_{\textsc{Cluster}}$: grouping by clustering per-example attributions}
\label{alg:cluster}
\begin{algorithmic}[1]
\Require Model $G=(N,E)$; dataset $D=\{x_1,\ldots,x_n\}$; attribution method $\mathcal{F}$; binarization threshold $\gamma$; reduction to dimension $r$; distance metric $d$; clustering method \textsc{Cluster}
\Ensure A partition $\{D_k\}$ of $D$
 
\For{$i = 1, \ldots, n$}
    \State $\mathbf{s}_i \gets \big(s(e; x_i)\big)_{e \in E}$ \Comment{per-example edge attributions via $\mathcal{F}$}
    \State $\mathbf{b}_i \gets \textsc{Binarize}(\mathbf{s}_i, \gamma)$ \Comment{top-magnitude edges holding mass $\gamma$}
\EndFor
 
\State $\{\tilde{\mathbf{s}}_i\} \gets \textsc{Reduce}(\{\mathbf{b}_i\}, r)$ \Comment{e.g., PCA or SVD to $\mathbb{R}^r$}
\State $K^{*} \gets \textsc{SelectK}(\{\tilde{\mathbf{s}}_i\})$ \Comment{e.g., gap statistic, silhouette, or elbow}
\State $\{D_k\}_{k=1}^{K^{*}} \gets \textsc{Cluster}(\{\tilde{\mathbf{s}}_i\}, K^{*}, d)$
 
\State \Return $\{D_k\}$
\end{algorithmic}
\end{algorithm}

\subsection{Experimental setup}\label{app:dcd-data}
This section provides experimental setup details for validating the effectiveness of \approach{} (Section~\ref{exp3: our-approach}). We specifically provide additional details on the dataset construction and clustering of the dataset into different groups based on how similarly the model processes each instance in the dataset. 
\subsubsection{Dataset Construction}
Table~\ref{tab:mixture-sizes} lists the variants included in the cross-tasks and within-task mixture for each model, along with the number of training, validation, and test examples after the 60/20/20 split. We sample $n = 1{,}000$ examples per model uniformly across the listed variants, so the per-variant sample size scales inversely with the number of variants in the mixture (e.g., $\sim$167 per variant for GPT-2 with 6 variants, $\sim$77 per variant for Llama-3.1-8B-Instruct with 13 variants). Tasks are excluded from a certain when they have low accuracy; see Table~\ref{tab:accuracy} for raw accuracies. For instance, \emph{cross-task mixed dataset} for GPT-2 consists of only IOI and sequence completion because they have near-zero accuracy for entity-binding and arithmetic addition tasks.

\begin{table}[t]
\caption{Discovery mixtures used for \approachC{}. Each mixture holds about $1{,}000$ examples,
split 60/20/20 into discovery, validation and test sets (D/V/T), sampled evenly across the listed
variants. IOI \emph{2-person} is the \textsc{abba} and \textsc{baba} name orders together; P$_i$ is
the queried position in entity binding (EB); Arith.\ is arithmetic and SC sequence completion. Tasks
are omitted for a model when its accuracy on them is too low (Table~\ref{tab:accuracy}).}
\label{tab:mixture-sizes}
\centering
\footnotesize
\setlength{\tabcolsep}{3pt}
\begin{tabularx}{\linewidth}{@{}ll>{\raggedright\arraybackslash}Xr@{}}
\toprule
Mixture & Model & Variants & D/V/T \\
\midrule
Cross-task
 & GPT-2 & IOI: 2-person, letter, 3-person; SC: 2-, 3-, 4-gram & 594/198/204 \\
 & Qwen  & IOI: 2-person, letter, 3-person; EB: 2-comma, colour box, P$_1$; SC: 2-, 3-gram & 598/198/202 \\
 & Llama & IOI: 2-person, letter, 3-person; EB: 2-comma, colour box, P$_1$, P$_8$; Arith.: 2-, 3-operand, verbal v1; SC: 2-, 3-, 4-gram & 585/195/208 \\
\midrule
IOI    & all   & \textsc{abba}, \textsc{baba}, filler, letter, passive, 3-person & 594/198/204 \\
EB     & \shortstack[l]{Qwen,\\Llama} & 2-, 3-, 5-comma; P$_1$, P$_5$, P$_8$; 2-, 8-period; 2-, 8-colour box & 600/200/200 \\
Arith. & Llama & 2-, 3-operand; verbal v1, v2, v3 & 600/200/200 \\
SC     & all   & 2-, 3-, 4-gram & 597/198/204 \\
\bottomrule
\end{tabularx}
\end{table}

\subsubsection{Clustering details}\label{app:clustering}

We describe \approach{}-Cluster in Algorithm~\ref{alg:cluster}. Given $D = \{x_1, \ldots, x_n\}$, we obtain per-example edge attribution vectors $\mathbf{s}_i \in \mathbb{R}^{|E|}$ by applying the circuit discovery method $\mathcal{F}$ at the instance level, then preprocess and cluster them as follows.

\textbf{Binarization.} Per-example attribution vectors $\mathbf{s}_i \in \mathbb{R}^{|E|}$
are sparse and dominated by a small number of high-magnitude edges. We binarize
$\mathbf{s}_i$ by retaining the smallest set of top-magnitude edges whose cumulative
absolute attribution accounts for a fraction $\gamma = 0.99$ of the total per example,
yielding $\mathbf{b}_i \in \{0,1\}^{|E|}$. Because attribution is heavy-tailed, this
typically retains only a small fraction of edges by count while preserving 99\% of
the attribution mass. This makes similarity robust to magnitude differences and
aligns the signal with how circuits are conventionally defined — as sparse edge
subsets. Binarization is optional; results without it are similar.

\textbf{Dimensionality reduction.} We reduce $\mathbf{b}_i$ to $\tilde{\mathbf{s}}_i \in \mathbb{R}^{r}$ via PCA or truncated SVD, fixing $r = 20$ across all settings (pilot runs over $r \in \{5, 10, 15, 20, 30, 40, 50\}$ showed silhouette scores stable in this range). Both methods use \texttt{scikit-learn} defaults with \texttt{random\_state}\,$=42$.

\textbf{Distance metric.} For all clustering algorithms, we use Jaccard distance between the binary vectors $\mathbf{b}_i$, $d(\mathbf{b}_i, \mathbf{b}_j) = 1 - |\mathbf{b}_i \cap \mathbf{b}_j| / |\mathbf{b}_i \cup \mathbf{b}_j|$. We also experimented with cosine and Euclidean distances without the binarization preprocessing step and found that both yielded lower-quality clusters, as per silhouette scores compared to Jaccard distance. So, we based our experiments on the Jaccard distance.

\textbf{Clustering algorithms.} We evaluate three: (1) $K$-means; (2) agglomerative with average linkage, cut at $K$; (3) divisive via recursive $2$-means on the highest-variance cluster until $K$ groups remain. 
We report \textsc{\approach-kmeans-PCA} in the main paper.

\textbf{Selecting $K^{*}$ for cross-task mixture.} We sweep $K \in \{2, 3, \ldots, 20\}$ and select $K^{*}$
via the gap statistic~\citep{tibshirani2001estimating}, drawing $B = 20$ reference
datasets uniformly within the axis-aligned bounding box of $\tilde{\mathbf{s}}$ and
clustering each with $k$-means (\texttt{n\_init}\,$=5$). We compute the pooled
within-cluster dispersion $W_K$ in the reduced ($r=20$) space — the gap statistic
is degenerate in the original $\sim\!10^4$--$10^6$-dimensional binary space — and
apply the one-standard-error rule, picking the smallest $K$ with
$\mathrm{Gap}(K) \geq \mathrm{Gap}(K{+}1) - s_{K+1}$, where
$s_K = \sigma_K \sqrt{1 + 1/B}$ and $\sigma_K$ is the population standard deviation
(\texttt{ddof}\,$=0$) of the $B$ reference $\log W_K^{(b)}$ values. Table~\ref{tab:dcd-k-star} reports $K^{*}$ for every (task, model, variant).

\textbf{Selecting $K$ for the within-task mixtures.} The within-task results in
Section~\ref{subsec:evaluation} set $K$ to the number of variants on each
axis rather than to the gap-statistic $K^{*}$. The within-task mixtures are
smaller and contain only two to eight variants, and on them the gap statistic
over-splits: it returns a median $K^{*}$ of $10$ on axes with two or three
variants, so that examples of a single variant are spread across several
clusters and no cluster circuit can be read as belonging to one variant. We
therefore fix $K$ to the variant count. This is an oracle choice of $K$: it
uses the number of variants on the axis, but not which variant any example
belongs to, and \krandom{} uses the same $K$, so the comparison in
Section~\ref{subsec:evaluation} isolates whether attribution similarity,
rather than the number of groups, recovers the variant boundaries. The
within-task numbers should accordingly be read as what the grouping can
recover when $K$ is right, not as what the pipeline's own rule returns;
selecting $K$ automatically for such mixtures is an open problem for this
instantiation. Cross-task results use the gap-statistic $K^{*}$ throughout.

\begin{table}[h]
\centering
\setlength{\tabcolsep}{4pt}
\caption{Gap-statistic $K^{*}$ per (task, model) for the four \approach{} clustering variants at reduced dimension $r = 20$ and binarization threshold $\gamma = 0.99$.}
\label{tab:dcd-k-star}
\resizebox{\linewidth}{!}{%
\begin{tabular}{llcccc}
\toprule
\textbf{Task} & \textbf{Model} & \textsc{agglom-SVD} & \textsc{divisive-SVD} & \textsc{kmeans-PCA} & \textsc{kmeans-SVD} \\
\midrule
all-tasks            & GPT-2 Small             & 6 & 12 & 7  & 7  \\
all-tasks            & Qwen2.5-7B-Instruct     & 13 & 12 & 7  & 8  \\
all-tasks            & Llama-3.1-8B-Instruct   & 9  & 11 & 11 & 11 \\
\midrule
arithmetic           & Llama-3.1-8B-Instruct   & 2& 3& 2& 4\\
\midrule
entity-binding       & Qwen2.5-7B-Instruct     & 5 & 7 & 6 & 6 \\
entity-binding       & Llama-3.1-8B-Instruct   & 2 & 2 & 2 & 2 \\
\midrule
IOI                  & GPT-2 Small             & 3 & 3 & 3 & 2 \\
IOI                  & Qwen2.5-7B-Instruct     & 7 & 6 & 5 & 3 \\
IOI                  & Llama-3.1-8B-Instruct   & 4 & 4 & 3 & 3 \\
\midrule
SC  & GPT-2 Small             & 3 & 3 & 2 & 2 \\
SC  & Qwen2.5-7B-Instruct     & 2 & 2 & 3 & 3 \\
SC  & Llama-3.1-8B-Instruct   & 2 & 2 & 2 & 2 \\
\bottomrule
\end{tabular}}
\end{table}

\subsection{Best-of-$K$ faithfulness on the cross-task mixture}
\label{app:dcd-best-of-k}
Before interpreting the clusters, we first check whether the $K$ circuits discovered from subsets of the data collectively explain the held-out examples as faithfully as a single circuit discovered by hypothesis-driven methods. We measure \emph{best-of-$K$ faithfulness}: for each test example $x_i$, we take the most faithful circuit, $f^\ast(x_i)=\max_k f(C_k,\{x_i\})$, and average over the test set. We use this only as a coverage sanity check: high best-of-$K$ faithfulness indicates that the circuit set collectively covers the test examples, not that the clustering reflects the correct mechanistic partition. We compare the best-of-$K$ faithfulness of \approachC{} with hypothesis-driven approaches (EAP-IG, EAP, E-Act), \krep{}, adapted from \cite{franco2026finding}, which uses \approachC{}’s clusters but takes each cluster’s circuit to be the per-example circuit of the cluster’s medoid (i.e., the example whose attribution vector is closest to all others in its cluster); and \krandom{}, which discovers $K$ circuits from random partitions of the same discovery set, with $K$ and the partition sizes matched to \approachC{}'s clusters, so that it isolates what best-of-$K$ selection alone contributes. The random-edges curve in the figures is a circuit of the same size whose edges are drawn uniformly at random, included as a floor.

\begin{figure}[t!]
    \centering
    \includegraphics[width=0.95\linewidth]{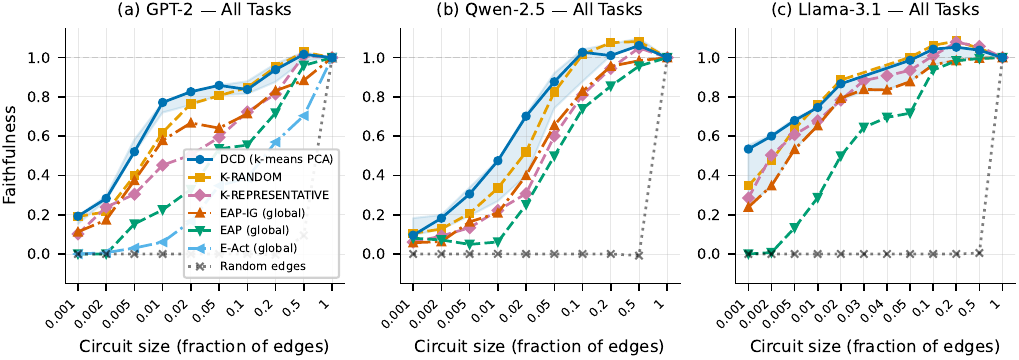}
    \caption{Best-of-$K$ faithfulness against circuit size on the cross-task mixture for
    GPT-2, Qwen2.5-7B-Instruct and Llama-3.1-8B-Instruct. The line shows \approachC{} with
    $k$-means on PCA-reduced attributions; the shaded band spans the four \approachC{} variants.
    \textsc{EAP-IG}, \textsc{EAP} and \textsc{E-Act} return a single circuit for the whole mixture.}
    \label{fig:dcd-faith-all}
\end{figure}

Figure~\ref{fig:dcd-faith-all} reports best-of-$K$ faithfulness against circuit size for the cross-task mixture on all three models and for \textit{within-task mixture datasets} in Figures~\ref{fig:dcd-faith-ioi}--\ref{fig:dcd-faith-sc}, where each dataset combines multiple variants within a single task (IOI, entity binding, arithmetic, or sequence completion) rather than mixing across tasks.

\subsection{Composition of the \approachC{} clusters}
\label{app:cluster-composition}

This appendix shows what every cluster reported in Section~\ref{subsec:evaluation} is made of: for
each cluster, how many of its examples come from each task and variant. \approachC{} never sees these
labels; we add them afterwards only to read the clusters. Colours follow one hue per task (IOI red,
sequence completion blue, entity binding green, arithmetic purple), with shades for variants; on
ordered axes (list length, queried position, $n$-gram length) the shades run from dark to light.

\subsubsection{Cross-task clusters}
\label{app:composition-crosstask}

\begin{figure}[h]
    \centering
    \includegraphics[width=\linewidth]{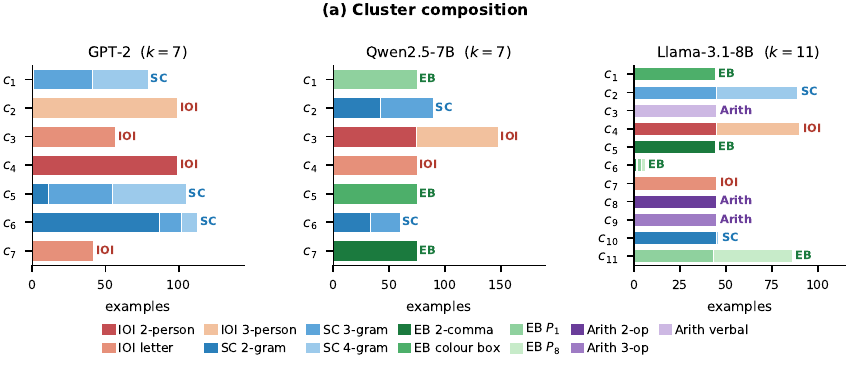}
    \caption{Composition of the cross-task \approachC{} clusters, $k$ chosen by the gap statistic.
    Each bar is one cluster; the label at its end names the task of its examples.}
    \label{fig:app-crosstask-composition}
\end{figure}

On all three models every cluster holds examples of a single task
(Figure~\ref{fig:app-crosstask-composition}). Many clusters go further and hold a single variant.
On GPT-2 the 3-person, letter and 2-person IOI prompts fall into separate clusters, and on Llama
so do the three arithmetic variants, the 2-comma and colour-box entity-binding prompts, and the
2-gram sequence-completion prompts. Other clusters pool variants of one task, most often the
sequence-completion $n$-grams, and on Llama the entity-binding positions $P_1$ and $P_8$ share a
cluster.

\subsubsection{Within-task clusters on GPT-2}
\label{app:composition-gpt2}

\begin{figure}[h]
    \centering
    \includegraphics[width=\linewidth]{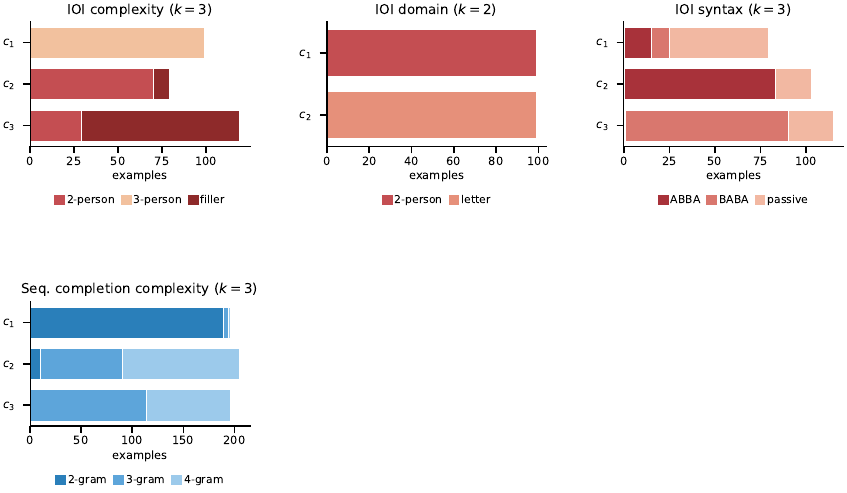}
    \caption{Composition of the within-task \approachC{} clusters on GPT-2, one panel per axis, $k$
    equal to the number of variants.}
    \label{fig:app-within-composition-gpt2}
\end{figure}

On IOI domain the two clusters are exactly the name and letter-label prompts, and on IOI complexity
the 3-person prompts form a cluster of their own ($99$ of $99$) while 2-person and filler prompts
are mostly, but not fully, apart ($70$ and $90$ of $99$ in their main clusters). IOI syntax splits
roughly by name order: most \textsc{abba} prompts share one cluster and most \textsc{baba} prompts
another, with the passive prompts spread across all three. On sequence completion the 2-gram
prompts form a cluster of their own ($189$ of its $196$ examples), while 3- and 4-gram prompts mix
across the other two (Figure~\ref{fig:app-within-composition-gpt2}).

\subsubsection{Within-task clusters on Qwen2.5}
\label{app:composition-qwen}

\begin{figure}[h]
    \centering
    \includegraphics[width=\linewidth]{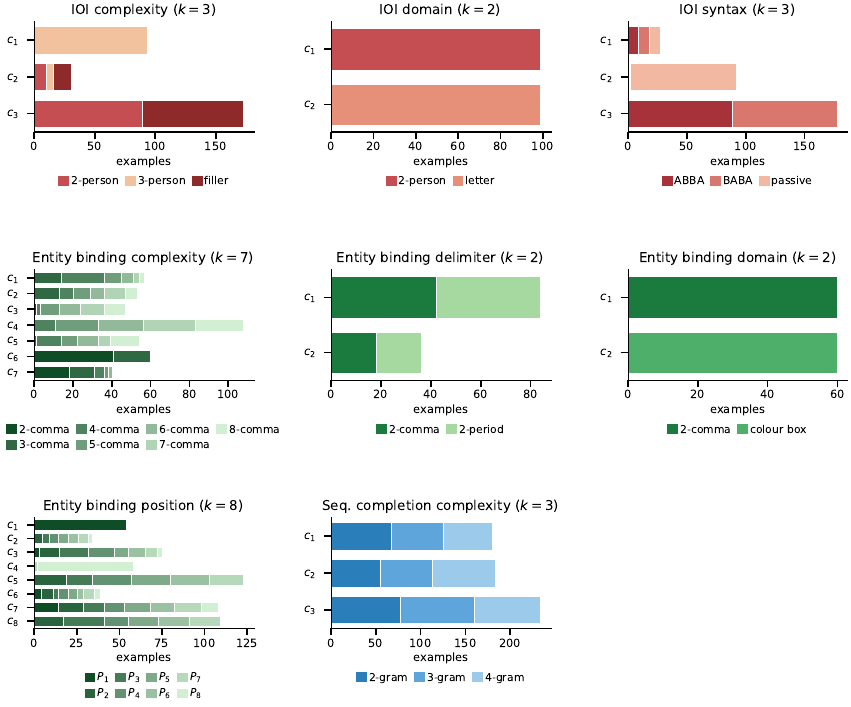}
    \caption{Composition of the within-task \approachC{} clusters on Qwen2.5-7B-Instruct.}
    \label{fig:app-within-composition-qwen}
\end{figure}

The IOI and entity-binding domain axes split exactly by variant. On IOI complexity the 3-person
prompts again form a cluster of their own, while 2-person and filler prompts share one, and on IOI
syntax the passive prompts separate while \textsc{abba} and \textsc{baba} prompts share a cluster.
On entity-binding position the clusters isolate the two end positions, $P_1$ ($54$ examples, all
$P_1$) and $P_8$ ($56$ of $58$), and spread the six middle positions across the rest; on
entity-binding complexity the 2-comma prompts gather in one cluster with part of the 3-comma
prompts, and the longer lists spread over the others. The delimiter axis is not separated at all:
each cluster holds 2-comma and 2-period prompts in equal numbers. On sequence completion every
cluster holds all three $n$-gram lengths in similar proportions
(Figure~\ref{fig:app-within-composition-qwen}).

\subsubsection{Within-task clusters on Llama}
\label{app:composition-llama}

\begin{figure}[h]
    \centering
    \includegraphics[width=\linewidth]{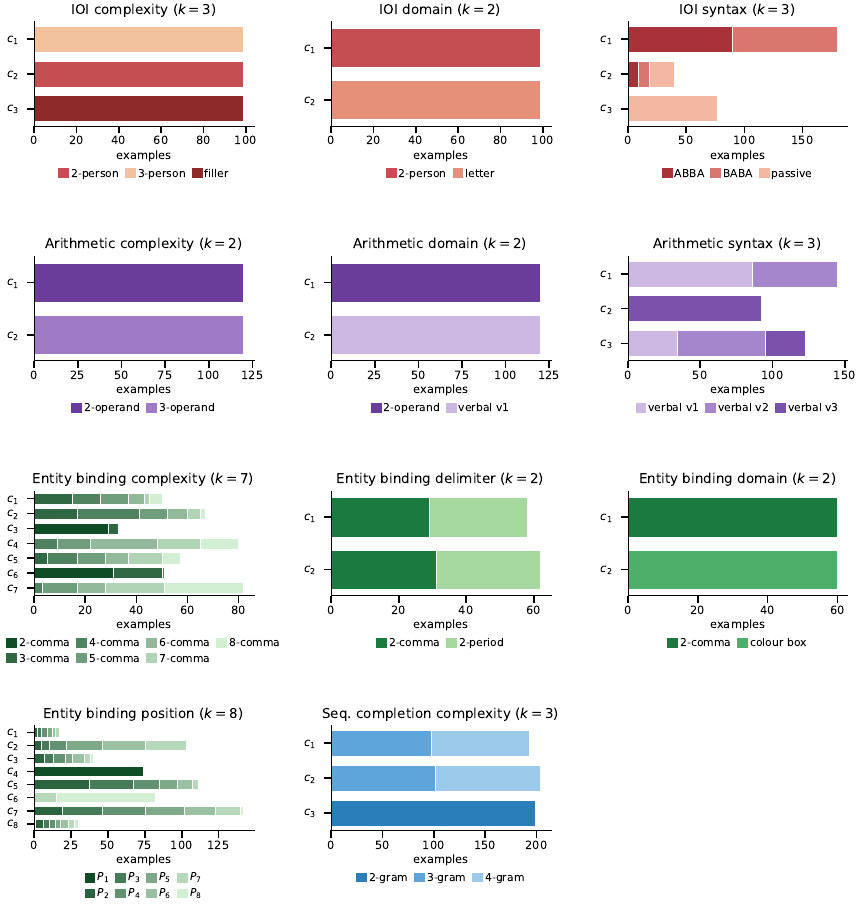}
    \caption{Composition of the within-task \approachC{} clusters on Llama-3.1-8B-Instruct.}
    \label{fig:app-within-composition-llama}
\end{figure}

Llama has the most clusters that hold a single variant. IOI complexity and domain, arithmetic
complexity and domain, and entity-binding domain all split exactly by variant. On arithmetic syntax
the third phrasing (\emph{how much is}) forms a cluster of its own, while the first two share
clusters; on IOI syntax \textsc{abba} and \textsc{baba} prompts share a cluster and the passive
prompts mostly form another; and on sequence completion the 2-gram prompts form a cluster of their
own ($199$ of $200$) while 3- and 4-gram prompts share the other two. The entity-binding axes
behave as on Qwen2.5: the position clusters isolate $P_1$ ($74$ of $74$) and $P_8$ ($67$ of $82$,
the rest $P_7$) and mix the middle positions, the complexity clusters follow list length only
loosely, and the delimiter axis is not separated
(Figure~\ref{fig:app-within-composition-llama}).


\subsection{Clusters on the pairs where neither circuit transfers}
\label{app:twoway-clusters}

Section~\ref{subsec: single-task-mixing} finds seven variant pairs where neither variant's circuit
transfers to the other. Figure~\ref{fig:app-twoway-clusters} shows, for each of these pairs, how the
circuits of the \approachC{} clusters (left) and of random splits of the same sizes (right) score on
the test prompts of the two variants. 

\begin{figure}[h]
    \centering
    \begin{minipage}[b]{0.32\linewidth}\centering
        \includegraphics[width=\linewidth]{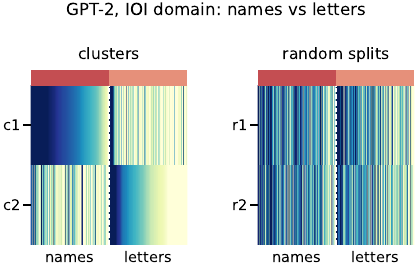}\\(a)
    \end{minipage}\hfill
    \begin{minipage}[b]{0.32\linewidth}\centering
        \includegraphics[width=\linewidth]{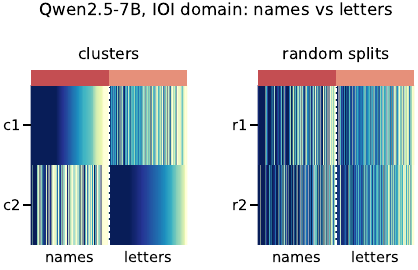}\\(b)
    \end{minipage}\hfill
    \begin{minipage}[b]{0.32\linewidth}\centering
        \includegraphics[width=\linewidth]{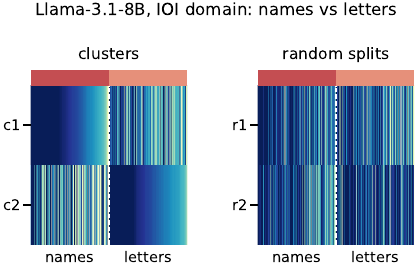}\\(c)
    \end{minipage}\\[6pt]
    \begin{minipage}[b]{0.32\linewidth}\centering
        \includegraphics[width=\linewidth]{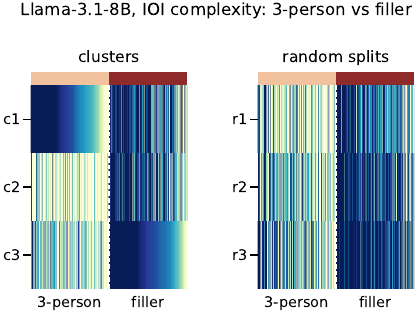}\\(d)
    \end{minipage}\\[6pt]
    \begin{minipage}[b]{0.32\linewidth}\centering
        \includegraphics[width=\linewidth]{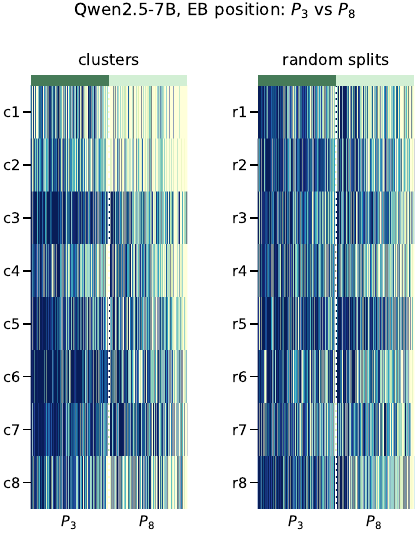}\\(e)
    \end{minipage}\hfill
    \begin{minipage}[b]{0.32\linewidth}\centering
        \includegraphics[width=\linewidth]{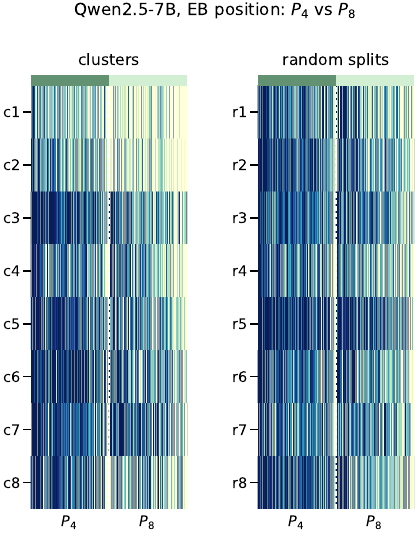}\\(f)
    \end{minipage}\hfill
    \begin{minipage}[b]{0.32\linewidth}\centering
        \includegraphics[width=\linewidth]{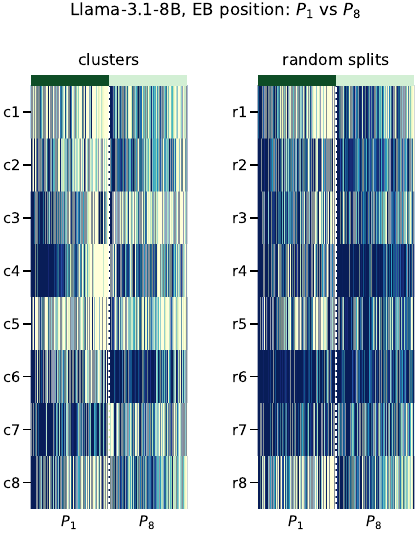}\\(g)
    \end{minipage}
    \caption{Cluster circuits ($c_i$) and random-split circuits ($r_i$) on the seven pairs where
    neither variant's circuit transfers, circuit size $0.05$. (a--c) IOI names against letter labels
    on GPT-2, Qwen2.5 and Llama; (d) IOI 3-person against filler on Llama; (e--g) entity-binding
    positions $P_3$ and $P_4$ against $P_8$ on Qwen2.5, and $P_1$ against $P_8$ on Llama.}
    \label{fig:app-twoway-clusters}
\end{figure}

\begin{figure}[H]
    \centering
    \includegraphics[width=0.8\textwidth]{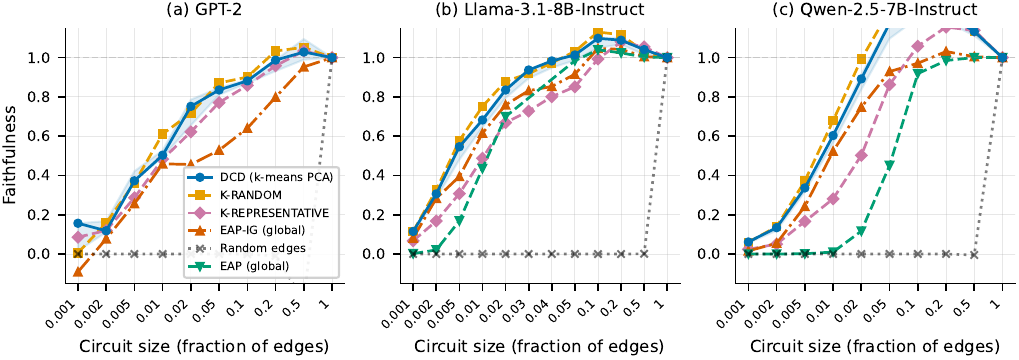}
    \caption{Best-of-$K$ faithfulness vs.\ circuit size on the IOI mixture across GPT-2, Llama-3.1-8B-Instruct, and Qwen-2.5-7B-Instruct. Shaded region: range across DCD variants.}
    \label{fig:dcd-faith-ioi}
\end{figure}

\begin{figure}[H]
    \centering
    \includegraphics[width=0.8\textwidth]{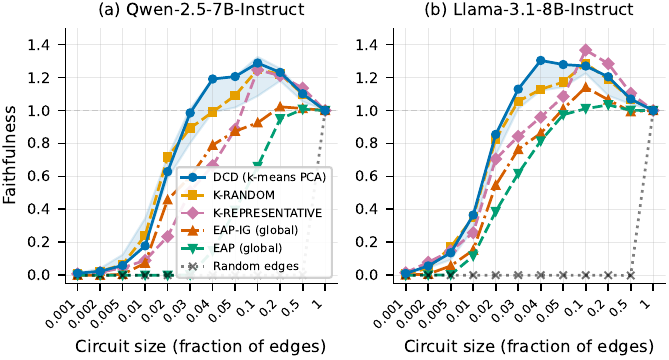}
    \caption{Best-of-$K$ faithfulness vs.\ circuit size on the entity binding mixture across Qwen2.5-7b-instruct and Llama-3.1-8B-Instruct. Shaded region: range across DCD variants.}
    \label{fig:dcd-faith-eb}
\end{figure}

\begin{figure}[H]
    \centering
    \includegraphics[width=0.4\textwidth]{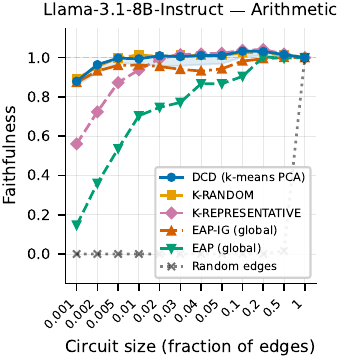}
    \caption{Best-of-$K$ faithfulness vs.\ circuit size on the arithmetic mixture in Llama-3.1-8B-Instruct. Shaded region: range across DCD variants.}
    \label{fig:dcd-faith-add}
\end{figure}

\begin{figure}[H]
    \centering
    \includegraphics[width=0.7\textwidth]{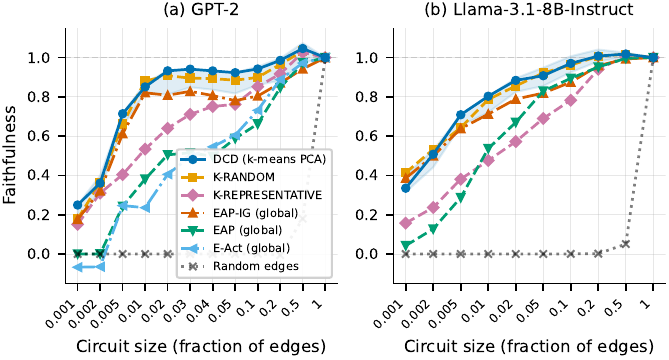}
    \caption{Best-of-$K$ faithfulness vs.\ circuit size on the sequence-completion mixture across Qwen2.5-7b-instruct and Llama-3.1-8B-Instruct. Shaded region: range across DCD variants.}
    \label{fig:dcd-faith-sc}
\end{figure}

\subsection{Additional Results: DCD circuits have coherent and interpretable faithfulness behaviors}
\label{sec:appendix-dcd-coherent}

We provide additional results on Llama-3.1-8B-Instruct and Qwen2.5-7B-Instruct in Figure~\ref{fig:dcd-coherence-llama} and Figure~\ref{fig:dcd-coherence-qwen}, extending the cluster-composition and per-example faithfulness analysis from Section~\ref{subsec:evaluation} to the larger models and to all-task mixed datasets that include entity binding and arithmetic variants.

We observe the same pattern as in GPT-2 across both models. On Llama-3.1-8B-Instruct (Figure~\ref{fig:dcd-coherence-llama}), \textit{DCD} partitions the all-task mixed dataset into eleven clusters, each consisting almost entirely of examples from a single task: $C_4$ and $C_7$ contain only IOI examples, $C_2$ and $C_{10}$ contain only sequence completion, $C_3$, $C_8$, and $C_9$ contain only arithmetic, and $C_1$, $C_5$, $C_6$, and $C_{11}$ contain only entity binding. More finely, \textit{DCD} also separates within-task variants: arithmetic clusters split between symbolic (\texttt{2-op-add}, \texttt{3-op-add}) and verbal phrasing examples, and entity binding clusters distinguish position-0 from position-7 queries and color-box from comma delimiters. On Qwen2.5-7B-Instruct (Figure~\ref{fig:dcd-coherence-qwen}), \textit{DCD} recovers seven clusters that similarly separate IOI, sequence completion, and entity binding into distinct groups, with the entity-binding examples further split by position and surface form across $C_1$, $C_5$, and $C_7$.

Per-example faithfulness mirrors the cluster composition on both models. Each \textit{DCD} circuit achieves high faithfulness on examples from its dominant task and near-zero faithfulness elsewhere, producing the same clear block structure observed on GPT-2 in Figure~\ref{fig:dcd-coherence-gpt2}. For instance, on Llama, the arithmetic circuits ($C_3$, $C_8$, $C_9$) show high faithfulness only on the arithmetic columns of the heatmap, and the IOI circuits ($C_4$, $C_7$) show high faithfulness only on the IOI columns. The same coherence holds on Qwen2.5: each circuit owns a contiguous block of examples corresponding to its cluster's task and variant, with little spillover.

Together, these results show that \textit{DCD}'s ability to discover coherent, mechanism-specific circuits is not specific to GPT-2 or to small task mixtures. Even on larger instruction-tuned models and on mixtures that span four tasks and over a dozen variants, \textit{DCD} recovers circuits whose explanatory scope aligns with how the model organizes its computation rather than with human-defined task labels.

\begin{figure}[H]
    \centering
    \includegraphics[width=\textwidth]{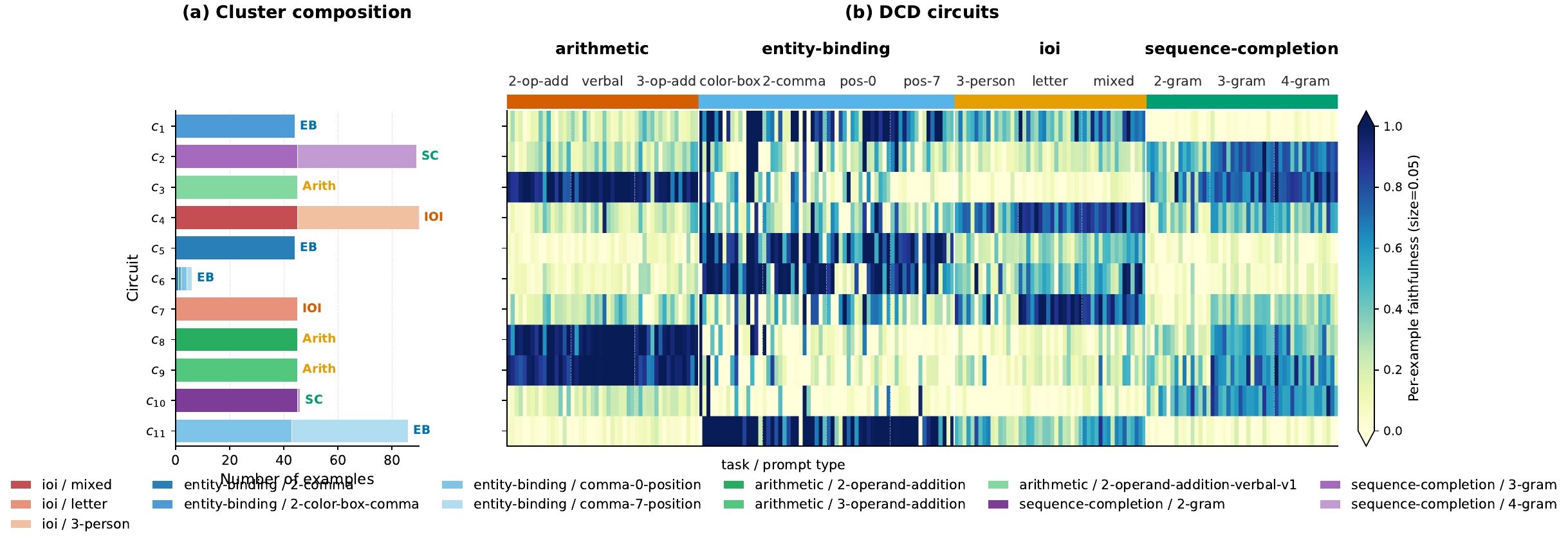}
    \caption{Results on Llama-3.1-8b-Instruct. (a) Each \approach{} cluster consists of examples from only a single task; (b) per-example faithfulness shows clear block structure -- each \approach{} circuit achieves high faithfulness on a subset of examples. Each row across (c1-c11) corresponds to one circuit}
    \label{fig:dcd-coherence-llama}
\end{figure}

\begin{figure}[H]
    \centering
    \includegraphics[width=\textwidth]{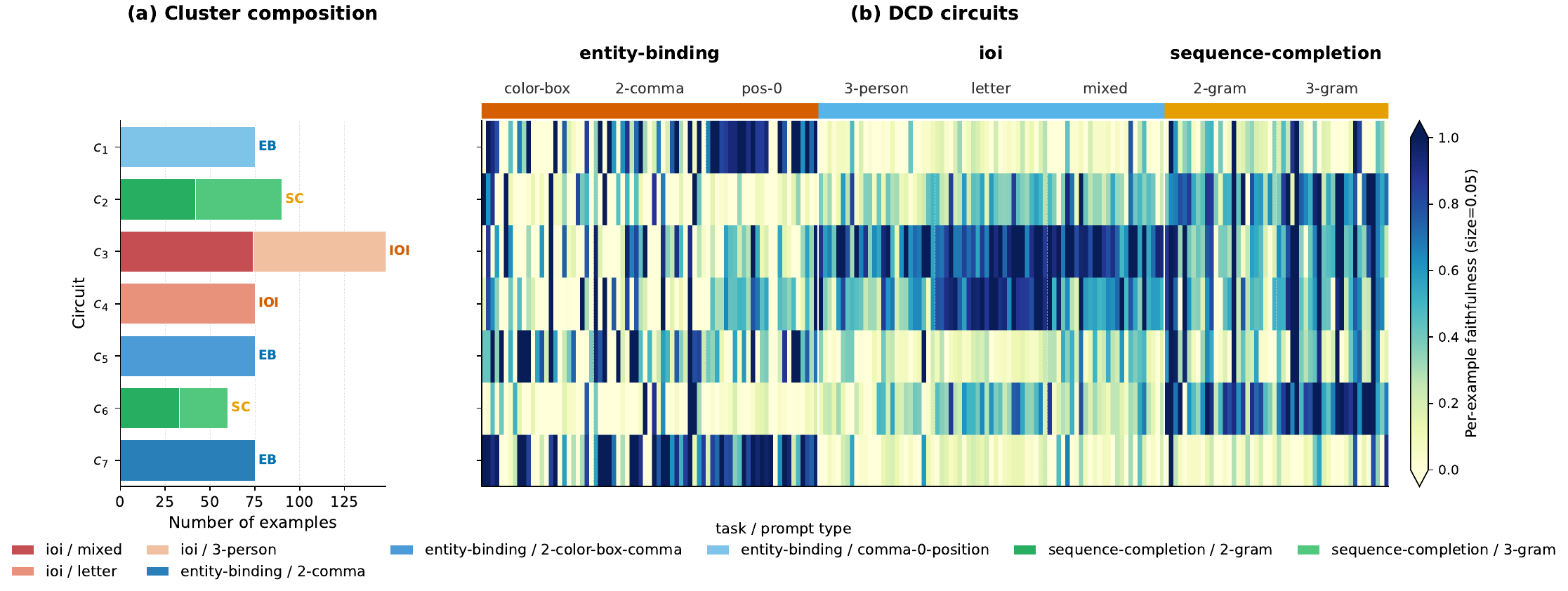}
    \caption{Results on Qwen2.5-7b-Instruct. (a) Each \approach{} cluster consists of examples from only a single task; (b) per-example faithfulness shows clear block structure -- each \approach{} circuit achieves high faithfulness on a subset of examples. Each row across (c1-c7) corresponds to one circuit.}
    \label{fig:dcd-coherence-qwen}
\end{figure}

\subsection{Compute resources}\label{app:compute}

All experiments were run on a single NVIDIA A100 GPU per run: A100-80GB for
Llama-3.1-8B-Instruct, A100-40GB for GPT-2 Small and Qwen2.5-7B-Instruct. No experiment required multi-GPU or multi-node setups. The dominant cost is per-example \textsc{EAP-IG} attribution on Llama-3.1-8B-Instruct, which scales with the number of examples. Similarly, \textsc{E-ACT} is computationally expensive, and we were unable to conduct circuit studies with \textsc{E-ACT} on Llama-3.1-8B-Instruct and Qwen2.5-7B-Instruct even on A100-80GB. The clustering step ($K \in \{2,\ldots,20\}$ sweep with $B = 20$ gap-statistic reference samples) is negligible relative to attribution
on the larger models.

\section{Asset attribution and licenses}\label{app:assets}

This section enumerates the external assets used in this paper and their licenses.

\textbf{Models.} We use GPT-2 Small~\citep{radford2019language} (MIT License),
Qwen2.5-7B and Qwen2.5-7B-Instruct~\citep{qwen2024qwen2} (Apache 2.0), and Llama-3.1-8B-Instruct~\citep{dubey2024llama} (Llama 3.1 Community License). All three are used for non-commercial academic research consistent with their license
terms.

\textbf{Circuit discovery code.} Our implementation of EAP, EAP-IG, and E-ACT builds on the MIB benchmark codebase~\citep{mueller2025mib}
(\url{https://github.com/aaronmueller/mib}, Apache 2.0 license). The original methods are described in~\citet{syed2024attribution} (EAP),
\citet{hanna2024have} (EAP-IG), and \citet{conmy2023acdc} (E-ACT).

\textbf{Dataset templates.} The synthetic prompt templates for the four tasks are
constructed following procedures described in prior circuit-discovery work:
IOI~\citep{wang2023interpretability}, entity binding~\citep{gur2025mixing}, arithmetic~\citep{stolfo2023mechanistic,nikankin2025arithmetic},
and sequence completion~\citep{elhage2021mathematical,olsson2022context}. We
generate entity binding, arithmetic, and sequence completion ourselves; we use IOI using \citet{wang2023interpretability} codebase (https://github.com/redwoodresearch/Easy-Transformer) [MIT license].

\textbf{Software libraries.} We use standard open-source scientific computing
libraries — PyTorch, Hugging Face Transformers, scikit-learn, NumPy, and matplotlib
— under their respective permissive licenses (BSD, Apache 2.0, PSF).

\end{document}